\documentclass[11pt,twoside]{article}
\usepackage{fullpage}
\usepackage{epsf}
\usepackage{fancyhdr}
\usepackage{graphics}
\usepackage{graphicx}
\usepackage{psfrag}
\usepackage{microtype}
\usepackage{subfigure}
\usepackage{algorithmic}
\usepackage{color,xcolor}
\usepackage{caption}
\usepackage{subcaption}
\usepackage{subfigure}
\usepackage{amsmath, wasysym, verbatim}
\usepackage{newunicodechar}

\usepackage[linesnumbered,ruled]{algorithm2e}
\DontPrintSemicolon

\usepackage{color}

\usepackage{amsthm}
\usepackage{amsfonts}
\usepackage{amssymb,bbm}

\usepackage{mathrsfs}

\usepackage{url}
\usepackage[colorlinks,linkcolor=magenta,citecolor=blue, pagebackref=true,backref=true]{hyperref}
\renewcommand*{\backrefalt}[4]{%
    \ifcase #1 \footnotesize{(Not cited.)}%
    \or        \footnotesize{(Cited on page~#2.)}%
    \else      \footnotesize{(Cited on pages~#2.)}%
    \fi}

\usepackage{nicefrac}
\usepackage{comment}

\usepackage{chngpage}

 \usepackage{tabularx}%
\usepackage{enumitem}
\usepackage{booktabs}
\usepackage{caption}

\usepackage{bm,bbm}
\usepackage{mathtools}
\usepackage{cleveref}

\usepackage{eucal}

\theoremstyle{plain}
\newtheorem{theorem}{Theorem}
\newtheorem{lemma}{Lemma}

\newtheorem{proposition}{Proposition}

\usepackage{scalerel}

\newcommand*\R[0]{\mathbb{R}}
\newcommand*\Z[0]{\mathbb{Z}}

\newcommand*\pt[0]{\tilde{p}}

\def\ball{\mathbb{B}}

\newcommand{\cover}{N}

\newcommand{\stepsize}{\eta}

\newcommand{\real}{\ensuremath{\mathbb{R}}}

\newcommand{\density}{p}

\newcommand{\abss}[1]{\left| #1 \right |}

\newcommand{\weightFun}{\ensuremath{w}}
\newcommand{\mydefn}{\ensuremath{:=}}

\newcommand{\defn}{:=}

\newcommand{\matsnorm}[2]{|\!|\!| #1 | \! | \!|_{{#2}}}
\newcommand{\vecnorm}[2]{\| #1\|_{#2}}

\newcommand{\inprod}[2]{\ensuremath{\langle #1 , \, #2 \rangle}}

\newcommand{\kull}[2]{\ensuremath{D_{\text{KL}}(#1\; \| \; #2)}}

\newcommand{\Exs}{\ensuremath{{\mathbb{E}}}}
\newcommand{\Prob}{\ensuremath{{\mathbb{P}}}}

\theoremstyle{plain}

\newlength{\widebarargwidth}
\newlength{\widebarargheight}
\newlength{\widebarargdepth}
\DeclareRobustCommand{\widebar}[1]{%
  \settowidth{\widebarargwidth}{\ensuremath{#1}}%
  \settoheight{\widebarargheight}{\ensuremath{#1}}%
  \settodepth{\widebarargdepth}{\ensuremath{#1}}%
  \addtolength{\widebarargwidth}{-0.3\widebarargheight}%
  \addtolength{\widebarargwidth}{-0.3\widebarargdepth}%
  \makebox[0pt][l]{\hspace{0.3\widebarargheight}%
    \hspace{0.3\widebarargdepth}%
    \addtolength{\widebarargheight}{0.3ex}%
    \rule[\widebarargheight]{0.95\widebarargwidth}{0.1ex}}%
  {#1}}

\makeatletter
\long\def\@makecaption#1#2{
        \vskip 0.8ex
        \setbox\@tempboxa\hbox{\small {\bf #1:} #2}
        \parindent 1.5em  
        \dimen0=\hsize
        \advance\dimen0 by -3em
        \ifdim \wd\@tempboxa >\dimen0
                \hbox to \hsize{
                        \parindent 0em
                        \hfil
                        \parbox{\dimen0}{\def\baselinestretch{0.96}\small
                                {\bf #1.} #2
                                }
                        \hfil}
        \else \hbox to \hsize{\hfil \box\@tempboxa \hfil}
        \fi
        }
\makeatother

\long\def\comment#1{}
\definecolor{battleshipgrey}{rgb}{0.52, 0.52, 0.51}
\definecolor{darkgray}{rgb}{0.66, 0.66, 0.66}
\definecolor{darkgreen}{rgb}{0.0, 0.2, 0.13}
\definecolor{darkspringgreen}{rgb}{0.09, 0.45, 0.27}
\definecolor{dukeblue}{rgb}{0.0, 0.0, 0.61}
\definecolor{olivedrab7}{rgb}{0.24, 0.2, 0.12}
\definecolor{darkblue}{rgb}{0.0, 0.0, 0.55}
\definecolor{darkscarlet}{rgb}{0.34, 0.01, 0.1}
\definecolor{candyapplered}{rgb}{1.0, 0.03, 0.0}
\definecolor{ao(english)}{rgb}{0.0, 0.5, 0.0}
\definecolor{applegreen}{rgb}{0.55, 0.71, 0.0}

\newcommand{\mwlcomment}[1]{{\bf{{\color{orange}{{Wenlong --- #1}}}}}}

\newcommand{\E}{\mathbb E}

\newcommand{\Var}{\mathrm{Var}}

\newcommand{\policyclass}{\Pi}

\newcommand{\policy}{\pi}

\newcommand{\support}{\mathrm{supp}}

\def\pt{\partial}

\newtheorem{assumption}{Assumption}
\newcommand{\simiid}{\stackrel{\mathrm{i.i.d.}}{\sim}}
\newcommand{\numobs}{\ensuremath{n}}

\newcommand{\T}{\mathbb{T}}

\newcommand{\ltwospace}{\mathbb{L}^2}

\newcommand{\totaltime}{T}

\newcommand{\stationary}{\xi}

\newcommand{\fakerefassumelip}[1]{\hyperref[assume:smooth-high-order]{{\color{magenta} {\upshape\textbf (}{\upshape{\textbf{Lip}}#1}{\upshape\textbf )}}} }

\DeclareFontFamily{U}{mathx}{}
\DeclareFontShape{U}{mathx}{m}{n}{<-> mathx10}{}
\DeclareSymbolFont{mathx}{U}{mathx}{m}{n}

\DeclareMathAccent{\widecheck}{0}{mathx}{"71}

\usepackage{mathbbol}

\long\def\comment#1{}

\newcommand{\ltwonorm}[1]{\vecnorm{#1}{\ltwospace}}

\newcommand{\Fclass}{\ensuremath{\mathcal{F}}}

\newcommand{\Rad}{\mathfrak{R}}
\usepackage[framemethod=TikZ]{mdframed}

\newenvironment{narrowpara}
  {\par\addvspace{\smallskipamount}\narrower\noindent\ignorespaces}
  {\par\addvspace{\smallskipamount}}

\newcommand{\LinSpace}{\mathbb{K}}

\newcommand{\Dset}{\mathcal{D}}

\newcommand{\tburn}{t_0}

\newcommand{\ffunc}{\ensuremath{f}}
\newcommand{\ctrlE}{H}
\newcommand{\internalE}{E}
\renewcommand{\density}{\rho}
\newcommand{\temperature}{\tau}
\renewcommand{\stationary}{\mu}

\newcommand{\empsobo}{\widehat{\mathbb{H}}^1}
\newcommand{\empsobonorm}[1]{\vecnorm{#1}{\empsobo}}

\newcommand{\empltwo}{\widehat{\mathbb{L}}^2}
\newcommand{\empltwonorm}[1]{\vecnorm{#1}{\empltwo}}

\newcommand{\initDistr}{\nu_0}
\newcommand{\occupmsr}{\nu_*}

\newcommand{\rhotilde}{\widetilde{\rho}}

\newcommand{\torus}{\mathbb{T}}

\newcommand{\dt}{\stepsize}

\newcommand{\eucnorm}[1]{\abss{#1}}

\newcommand{\Nsensors}{N}

\begin{document}

\begin{center}
{\bf{\LARGE{Provable imitation learning for control of instability in partially-observed Vlasov--Poisson equations}}}

\vspace*{.2in}
{\large{
 \begin{tabular}{cccc}
  Xiaofan Xia$^{ \dagger}$ &  Qin Li$^{\diamond}$ &
  Wenlong Mou$^{\dagger}$ 
 \end{tabular}

}

\vspace*{.2in}

 \begin{tabular}{c}
 Department of Statistical Sciences, University of Toronto$^{\dagger}$\\
  Department of Mathematics, University of Wisconsin--Madison$^{\diamond}$
 \end{tabular}

}

\end{center}
\begin{abstract}
  We consider the stabilization of Vlasov--Poisson plasma dynamics, a central control problem in nuclear fusion. Our focus is the gap between what an ideal controller would use and what experiments can actually observe: while optimal policy may rely on the full phase-space state, practical feedback is typically limited to sparse macroscopic diagnostics. We therefore study imitation learning methods that distill a fully observed expert policy into controllers operating only on macroscopic measurements. We show the stability guarantees of the learned policy, where the error floor depends on the minimal behavior cloning loss achievable under the observation constraints. We further characterize this minimal loss in terms of a notion of entropy that quantifies the complexity of the initial distribution. Our results demonstrates the theoretical feasibility of learning stabilizing feedback policies for kinetic plasma dynamics from macroscopic observations, and exhibits the adaptivity of the learning approach to low-complexity structures. Through extensive numerical experiments, we validate our theory and show that the learned policies can stabilize the system using only macroscopic observations, within a significantly longer time horizon than non-adaptive baseline controllers.
\end{abstract}

\section{Introduction}
For plasma fusion to be viable, the Lawson criterion requires that the product of plasma density, confinement time, and temperature exceed a critical threshold. Physically, this means that charged particles must remain sufficiently hot and dense for long enough durations for nuclear fusion reactions to occur. Different fusion devices pursue this goal through different trade-offs. For example, the National Ignition Facility (NIF) emphasizes extremely high temperature and density over short durations, while magnetic confinement devices such as tokamaks and stellarators aim to sustain moderate temperature and density over extended confinement times.

Across these settings, a central challenge is stability. Plasma consists of charged particles interacting through self-consistent electromagnetic fields, and these interactions naturally give rise to instabilities that degrade confinement by redistributing or dissipating energy. A primary control objective is therefore to design external electric or magnetic fields that guide particle motion and suppress unstable modes, thereby maintaining a high-energy state over longer time horizons.

A fundamental obstacle in achieving such stabilization arises from a mismatch between the underlying dynamics and available observations. The evolution of plasma is governed by high-dimensional kinetic equations, such as the Vlasov--Poisson system, where instability mechanisms depend sensitively on fine-scale structure in phase space, particularly in the velocity variable. In contrast, most feedback control systems---such as those deployed in tokamaks, including the DIII-D operated by General Atomics in San Diego~\cite{Degrave2022}---rely on sparse, low-dimensional macroscopic observables, including density, temperature, and electromagnetic field measurements. These quantities are accessible to real-time diagnostics and form the basis of existing control strategies.

This discrepancy creates a severe partial observability challenge: distinct velocity-space configurations can produce nearly identical macroscopic measurements while exhibiting drastically different stability behavior. This sensitivity is formalized by classical criteria such as the Penrose condition~\cite{Landau1946,Chen2016}, which characterizes stability in terms of fine-scale features of the velocity distribution that are not captured by macroscopic observables. As a result, kinetic instabilities---driven by velocity-space structure---may not be detectable or controllable using macroscopic observations alone.

This gap motivates the central question of this work:
\begin{quote}
    \textit{Can stabilizing control policies for kinetic plasma dynamics be realized using only sparse, macroscopic observations?}
\end{quote}
We formalize this setting as a control problem under partial observability. Specifically, we consider policies of the form
\[
\pi \big( \Dset [t - t_0, t] \big)
\]
where $\Dset [t - t_0, t]$ denotes the observation history from time $t_0$ to $t$, collected at a limited number of sensors. The key question is whether such policies, operating on low-dimensional observation histories, can encode sufficient information to stabilize inherently high-dimensional kinetic dynamics.

While this problem remains largely open, recent theoretical advances provide important guidance. In particular, prior work~\cite{EINKEMMER2025113904} has derived analytically stabilizing control laws based on spectral analysis of kinetic models. These results establish stabilization guarantees at the kinetic level, but rely on access to detailed phase-space information that is not available in practical settings. Consequently, such controllers are not directly implementable, but they provide high-quality stabilizing strategies under full-state information.

In this work, we leverage these analytically derived controllers as experts with privileged information, and use them to supervise the learning of policies that operate under realistic observation constraints. Concretely, we adopt an imitation learning framework in which trajectories generated by kinetic-level controllers serve as demonstrations, while the learned policy is restricted to macroscopic inputs. Our contribution includes
\begin{itemize}
  \item By formulating stabilization of the Vlasov--Poisson system as an imitation learning problem under partial observability, we design and analyze behavior cloning algorithms that distill the analytically derived kinetic controllers into implementable feedback policies. We prove that the learned policy achieves stabilization guarantees, where the error floor depends on the minimal behavior cloning loss achievable under the observation constraints.
  \item We characterize the minimal behavior cloning loss in terms of the richness of observations and the complexity of the initial distribution of the system state. In particular, we use a scale-sensitive notion of entropy to quantify the complexity of the initial distribution, and show that small behavior cloning loss---and thus effective stabilization---can be achieved when the initial distribution has low entropy. Moreover, for sufficiently rich policy classes, the imitation learning algorithm automatically adapts to such low-entropy structure, achieving stabilization without explicit knowledge of the initial distribution's complexity.
  \item We validate our theoretical findings through numerical experiments on the Vlasov--Poisson system, demonstrating that the learned policies can effectively stabilize the dynamics using only macroscopic observations, within a significantly longer time horizon than non-adaptive baseline controllers that do not learn from data.
\end{itemize}
Our approach bridges the gap between analytically derived control laws and practically implementable feedback policies, by distilling expert knowledge into controllers that rely only on accessible measurements. This framework preserves the structure of the underlying kinetic model while enabling deployment in experimentally relevant regimes. 

\subsection{Related work}
Let us briefly review related literature across three areas: learning-based control in physical sciences and plasma, plasma control and kinetic instability, and imitation learning for control.

\paragraph{Learning-based control in physical science and plasma.}
Machine learning is increasingly used for scientific control problems governed by high-dimensional differential equations, including differentiable-physics control, physics-informed learning, neural mean-field control, operator learning for PDE-constrained control, and neural-operator surrogates for feedback design~\cite{Holl2020DifferentiablePhysics,raissi2019physics,ruthotto2020machine,hwang2022solving,bhan2023neural}. In fusion, learning-based tokamak control dates back to neural-network feedback~\cite{Bishop1995}; recent work studies disruption and locked-mode prediction across devices~\cite{KatesHarbeck2019,Rea2019DIIID,Montes2019Warnings,Akcay2021LockedMode}, connects predictors to tearing-mode and disruption avoidance or mitigation~\cite{Fu2020MLControl,Yang2022HL2A,Murari2024ControlOriented}, and applies deep or offline reinforcement learning to magnetic shape control and instability avoidance~\cite{Degrave2022,Seo2024,Char2023}. These methods demonstrate the value of AI-assisted plasma control, but either treat the plasma dynamics as a black box, or use reduced order models that do not fully capture the kinetic instability behavior. In our work, supervision comes from a kinetic Vlasov--Poisson controller with privileged phase-space information, while deployment uses only sparse macroscopic observations.

\paragraph{Plasma control, inverse problems, and kinetic instability.}
Classical tokamak control combines model-based magnetic, current, shape, and profile regulation with equilibrium or profile reconstruction from diagnostics~\cite{AriolaPironti2016,DeTommasi2019,Lao2005,Kinsey2022,McClenaghan2024}. Much of this pipeline is therefore an inverse problem: the controller acts on reconstructed plasma states, an inference-to-control viewpoint also developed through data assimilation~\cite{Morishita2024}. Kinetic instabilities, however, can depend on velocity-space structure invisible to low-dimensional moments, as reflected in Landau damping, Penrose criteria, and standard kinetic plasma theory~\cite{Landau1946,Penrose1960,Chen2016,mouhot2011landau}. For Vlasov--Poisson systems, control theory has studied exact controllability through localized controls or external force fields~\cite{Glass2003,GlassHanKwan2012}, while recent computational methods design stabilizing fields by constrained optimization, particle-in-cell optimal control, and Fourier--Laplace pole elimination~\cite{Einkemmer2024Constrained,Bartsch2024,EINKEMMER2025113904}. These approaches mostly assume access to the phase space state or the initial state, and do not address the partial observability challenge of learning stabilizing policies from macroscopic measurements.

\paragraph{Imitation learning for control.}
Imitation learning has long served as a control methodology, from autonomous driving by behavioral cloning~\cite{Pomerleau1989} to reductions showing that offline supervised imitation can suffer covariate shift and error compounding~\cite{RossBagnell2010,RossGordonBagnell2011,RossBagnell2014}. Recent theory refines when behavior cloning is statistically sufficient and when continuous-action dynamical systems make offline imitation unstable~\cite{Rajaraman2020,Foster2024,Simchowitz2025}. These guarantees are mostly for finite-dimensional, discrete-time decision processes, with performance governed by horizon dependence, coverage, or learner-induced state distributions. Our analysis instead treats an infinite-dimensional kinetic PDE under partial observation, and deals with phase mixing and instability phenomena that are not present in finite-dimensional systems.

\section{Problem setup}
Let $\torus \mydefn \mathbb{R}/\mathbb{Z}$ be the one-dimensional torus, we consider the system of Vlasov--Poisson equations on $[0, \totaltime] \times \torus \times \real$:
\begin{align}
  \begin{dcases}
    \pt_t \ffunc_t ( x, v) + v \pt_x \ffunc_t (x, v) - \big( \ctrlE_t (x) + \internalE_t ( x) \big) \pt_v \ffunc_t ( x, v) = 0, \\
    \pt_{xx} V_t (x) = 1 - \int \ffunc_t (x, v) dv, \\
    \internalE_t (x) = \pt_x V_t (x).
  \end{dcases}\label{eq:Vlasov--Poisson}
\end{align}
It is easy to verify that any spatial-homogeneous distribution $\ffunc_t (x, v) = \stationary (v)$ is a stationary solution to the above system. However, such stationary solutions are in general unstable. The goal of this paper is to design a control policy $\ctrlE_t (x)$ based on partial observations of the system state to stabilize the dynamics around the stationary solution. We assume that the initial function $\ffunc_0$ is random, following
\begin{align*}
  \ffunc_0^{(1)}, \ffunc_0^{(2)}, \ldots, \ffunc_0^{(\numobs)} \simiid \initDistr,
\end{align*}
where $(\ffunc^{(j)})_{j=1}^{n}$ are the initial conditions for $n$ observed data points, and $\initDistr$ is a distribution over the space of functions on $\torus \times \real$.

At time $t \geq 0$, the density of particles at location $x \in \torus$ is given by
\begin{align*}
  \density_t (x) = \int_{\real} \ffunc_t (x, v) dv.
\end{align*}

\paragraph{Observation model:} Given sensor locations $\{ x_i \}_{i=1}^{\Nsensors}$ and time interval $\dt$, we observe the density at location $x_i$ and time $k \dt$ with additive Gaussian noise:
\begin{align*}
  \rhotilde_{i, k} &= \density_{k \dt} (x_i) + W_{i, k}, \quad W_{i, k} \simiid \mathcal{N} (0, \sigma_\rho^2).
\end{align*}
For simplicity, we assume that the sensors are uniformly placed on the torus. In particular, we assume that the sensor locations are given by
\begin{align*}
  x_i = \frac{i - 1}{\Nsensors}, \quad i = 1, 2, \ldots, \Nsensors.
\end{align*}

\subsection{Symmetry and connection with Tokamak geometry}
It is important to emphasize that the use of the one-dimensional in space and velocity (1D1V) Vlasov–Poisson system is not merely a computational simplification, but arises as a reduced model of the full three-dimensional phase-space dynamics under physically relevant assumptions.

The full system in three spatial and three velocity dimensions is given by
\begin{equation}
\partial_t f + v \cdot \nabla_x f + (E + v \times B)\cdot \nabla_v f = 0,
\end{equation}
with the electric field determined self-consistently through Poisson’s equation. When the external magnetic field is strong and uniform, particle motion exhibits rapid rotation in the velocity components perpendicular to the field direction. In the regime where the magnetic field strength scales as $B = \frac{1}{\epsilon} e_1$ with $\epsilon \ll 1$, this induces a separation of scales between fast rotational dynamics and slow evolution.

A standard asymptotic expansion in $\epsilon$ shows that, at leading order, the distribution function becomes invariant under rotations in the perpendicular velocity plane. Consequently, the dynamics depend only on the parallel velocity component and the spatial coordinate aligned with the magnetic field. Under an additional slab symmetry assumption, the system reduces to the effective one-dimensional kinetic equation, the 1D1V Vlasov–Poisson system that we use.

This reduced model retains the essential velocity-space structure responsible for kinetic instabilities, while providing a tractable setting for studying control under partial observability.

\subsection{Full-information expert policy and its stability guarantees}

If the distribution $\ffunc_t (x, v)$ is fully observed, stabilizing control can be achieved by exact cancellation of the electric field $\ctrlE^*_t (x) = - \internalE_t (x)$, where the function $\internalE$ can be computed by solving the Poisson equation $\pt_{xx} V_t (x) = 1 - \density_t (x)$. In such a case, the equation becomes the free-stream transport equation
\begin{align}
  \partial_t \ffunc_t (x, v) + v \pt_x \ffunc_t (x, v) = 0.\label{eq:free-stream}
\end{align}
Under certain regularity assumptions on the initial condition $\ffunc_0$, it is known~\cite{mouhot2011landau,einkemmer2025control} that the free-stream equation~\Cref{eq:free-stream} enjoys exponential decay of electrical energy
\begin{align}
  \vecnorm{\internalE_t}{\ltwospace (\torus)}^2 \leq c_1 e^{- c_2 t},\label{eq:free-stream-decay}
\end{align}
for some positive constants $c_1, c_2 > 0$ depending on the initial condition $\ffunc_0$.

Though the full-information control policy $\ctrlE^*_t$ stabilizes the system, it cannot be learned directly from partial observations. In particular, the partially-observed control policy relies on sufficiently long observation history to predict the optimal control action. Therefore, we consider an expert policy that runs the uncontrolled Vlasov--Poisson dynamics for a short time interval $[0, \tburn]$ to collect observations, and then switches to the full-information control policy $\ctrlE^*_t$ for $t \geq \tburn$, i.e.,
\begin{align}
  \ctrlE^{\mathrm{expert}}_t (x) = \begin{dcases}
    0, & t \in [0, \tburn], \\
    - \internalE_t (x), & t > \tburn.
  \end{dcases}\label{eq:expert-policy}
\end{align}
To establish the stability of the expert policy, given a parameter $q \ge 2$, we define the weighted space norms
\begin{align*}
  \vecnorm{f}{\mathcal{C}^m_q} &\mydefn \sup_{\bm{1}^\top \alpha = m}~ \sup_{(x, v) \in \torus \times \real} (1 + \eucnorm{v})^{q} \big| \partial^\alpha \ffunc_0 (x, v) \big|, \qquad \mbox{for any } m \in \mathbb{N},\\
  \vecnorm{f}{\mathcal{E}_q (\lambda)} &\mydefn \sum_{m=0}^\infty \frac{\lambda^m}{m!} \vecnorm{f}{\mathcal{C}^m_q}, \qquad \mbox{for any } \lambda > 0.
\end{align*}
We make the following assumption regarding the initial condition $\ffunc_0$.

\begin{assumption}\label{assume:initial-regularity}
  There exist deterministic constants $\lambda_0, B_0 > 0$ and an integer velocity decay parameter $q \ge 2$, such that with probability 1, the random initial condition $\ffunc_0$ satisfies
  \begin{align*}
    \vecnorm{\ffunc_0}{\mathcal{E}_q (\lambda_0)} \leq B_0.
  \end{align*} 
\end{assumption}
Assumption~\ref{assume:initial-regularity} requires the initial condition $\ffunc_0$ to be infinitely differentiable with derivatives growing at most factorially. It also imposes a polynomial decay condition on the velocity variable $v$ to ensure integrability. These conditions are standard in the study of Landau damping~\cite{mouhot2011landau}. For example, if the initial density $\ffunc_0$ factorizes as $\ffunc_0 (x, v) = \density_0 (x) \cdot \mu_0 (v)$, where $\density_0$ is infinitely differentiable and $\mu_0$ is a mixture-of-Gaussian distribution, then Assumption~\ref{assume:initial-regularity} is satisfied.

Under Assumption~\ref{assume:initial-regularity}, the expert policy stabilizes the dynamics as stated in the following proposition.
\begin{proposition}\label{prop:expert-stability}
  Suppose that the initial condition $\ffunc_0$ satisfies Assumption~\ref{assume:initial-regularity}. Then there exist constants $c_1, c_2 > 0$ depending on $\ffunc_0$ such that for the expert policy $\ctrlE^{\mathrm{expert}}_t$, the electrical energy decays exponentially:
  \begin{align*}
    \vecnorm{\internalE_t}{\ltwospace (\torus)}^2 \leq c_1 e^{- c_2 t}, \quad \forall t \geq 0.
  \end{align*}
\end{proposition}
\noindent See \Cref{subsec:proof-expert-stability} for the proof of this proposition. \Cref{prop:expert-stability} shows that the expert policy $\ctrlE^{\mathrm{expert}}_t$ achieves exponential stabilization of the Vlasov--Poisson dynamics, where the decay rate depends on the regularity and velocity decay of the initial condition $\ffunc_0$. This result shows that the phase mixing effect of the free-streaming dynamics~\cite{mouhot2011landau} is robust with respect to the initial transient period $[0, \tburn]$ during which the system evolves under the uncontrolled dynamics. For the rest of the paper, we will focus on learning a policy to achieve similar phenomena using only partial observations.

\section{Imitation learning algorithm and its theoretical guarantees}\label{sec:main-algorithm}
In this section, we present the imitation learning algorithm for learning a partially-observed control policy from the expert trajectories generated by the full-information control policy. We also state the theoretical guarantees of the learned policy.

\subsection{Imitation learning algorithm}
Let us now describe the main algorithm. Given $\mathrm{i.i.d.}$ initial conditions $\{ \ffunc_0^{(j)} \}_{j = 1}^\numobs$, we simulate the controlled dynamics~\eqref{eq:Vlasov--Poisson} with expert policy~\eqref{eq:expert-policy} to obtain the expert trajectories $\{ \ffunc^{(j)}_t (x, v), \internalE^{(j)}_t (x) \}_{j=1}^\numobs$. Recall that for each trajectory, we collect noisy partial observations of the density at sensor locations $\{ x_i \}_{i=1}^N$ and discrete time steps $\{ k \dt \}_{k=0}^{\lfloor \totaltime / \dt \rfloor}$. For notation convenience, we define the pooled observation within time interval $[t_1, t_2]$ for $j$-th trajectory as
\begin{align*}
  \Dset^{(j)} [t_1, t_2] = \Big\{ \big( \rhotilde_{i, k}^{(j)} \big)_{i=1}^N : t_1 \leq k \dt \leq t_2 \Big\},
\end{align*}
When $t_1$ and $t_2$ are both integer multiples of $\dt$, the pooled observation $\Dset^{(j)} [t_1, t_2]$ can be represented as an $N \times \big( t_2 / \dt - t_1 / \dt + 1 \big)$ matrix. The imitation learning algorithm aims to learn a control policy $\ctrlE_t (x)$ that maps the observation history $\Dset [t - \tburn, t]$ to the expert control action $\ctrlE^{\mathrm{expert}}_t (x) = - \internalE_t (x)$ at time $t \geq \tburn$. Given a time-homogeneous policy class
\begin{align*}
  \policyclass \subseteq \Big\{ f : \real^{N \times (\tburn / \dt + 1)}  \to \ltwospace (\torus) \Big\},
\end{align*}
We learn the control policy by solving the empirical risk minimization problem
\begin{align}
  \widehat{\policy} 
  = \arg\min_{\policy \in \policyclass} 
  \widehat{\mathcal{R}}(\policy).
  \label{eq:erm-problem}
\end{align}
Here the empirical risk is defined as
\begin{align}
  \widehat{\mathcal{R}}(\policy)
  \coloneqq
  \frac{1}{\numobs}
  \sum_{j=1}^{\numobs}
  \widehat{\ell}^{(j)}(\policy),
  \label{eq:emp-risk-def}
\end{align}
where the discrete trajectory loss is
\begin{align}
  \widehat{\ell}^{(j)}(\policy)
  \coloneqq
  \frac{\stepsize}{\totaltime-\tburn}
  \sum_{\tburn \le k\stepsize \le \totaltime}
  \Big\|
  \policy\!\left(
    \Dset^{(j)}[k\stepsize-\tburn,k\stepsize]
  \right)
  + \internalE^{(j)}_{k\stepsize}
  \Big\|_{\ltwospace(\torus)}^2.
  \label{eq:discrete-loss}
\end{align}

This optimization problem is a least-squares regression problem using the function class $\policyclass$. When $\policyclass$ is convex, the optimization problem~\eqref{eq:erm-problem} is convex and therefore efficiently solvable. When we parametrize the policy class $\policyclass$ using neural networks, the optimization problem~\eqref{eq:erm-problem} can be solved using stochastic gradient descent. In practice, we can also use softmax or average over time steps to approximate the sum over time steps in~\eqref{eq:erm-problem}, in order to improve computational efficiency.

After obtaining the learned policy $\widehat{\policy}$, we deploy it to control the Vlasov--Poisson dynamics~\eqref{eq:Vlasov--Poisson} by setting
\begin{align*}
  \ctrlE^{\widehat{\policy}}_t (x) = \begin{cases}
    0, & t < \tburn, \\
 \widehat{\policy} \big( \Dset [t - \tburn, t]\big) (x), & t \geq \tburn,
   \end{cases}
\end{align*}
where the observation $\Dset [t - \tburn, t]$ is collected online during the control process.
\subsection{Theoretical guarantees of the learned policy}
Now we are ready to present the theoretical guarantees of the learned policy $\ctrlE^{\widehat{\policy}}_t$. The performance of the learned policy crucially depends on the ``learnability'' of the expert policy, i.e., whether the expert policy can be predicted accurately from the partial observations. Moreover, it also depends on the approximation power of the policy class $\policyclass$.

To characterize these two aspects, we introduce the population risk of a policy $\policy$ as
\begin{align*}
  \mathcal{R} (\policy) = \frac{1}{T - t_0}\int_{\tburn}^{\totaltime} \Exs \Big[ \vecnorm{\policy \big( \Dset^{(j)} [t - \tburn, t]\big) + \internalE^{(j)}_{t}}{\ltwospace (\torus)}^2 \Big] dt.
\end{align*}
where the expectation is taken over the random initial condition and the observation noise.
Equivalently, if $z$ denotes a trajectory sample,
we define the continuous-time trajectory loss
\begin{align}
  \ell(\pi;z)
  \coloneqq
  \frac{1}{T-t_0}\int_{t_0}^{T}
  \big\|\pi\big(\mathcal{D}_z[t-t_0,t]\big)+E_{z}(t)\big\|_{L^2(\T)}^2
  \,dt,
  \label{eq:trajectory-loss-cont}
\end{align}
so that $\mathcal{R}(\pi)=\mathbb{E}_{z}\big[\ell(\pi;z)\big]$.
The approximation error for the expert policy is then quantified by 
\begin{align*}
  \inf_{\policy \in \policyclass} \mathcal{R} (\policy).
\end{align*}
In \Cref{subsec:approx-theory}, we present a concrete example of policy class $\policyclass$ and derive explicit approximation error bounds.

Additionally, we need some technical assumptions on the policy class $\policyclass$ to control its complexity.

\begin{assumption}\label{assume:policy-class}
  For any policy $\policy \in \policyclass$, and any $M_1, M_2 \in \real^{N \times (\tburn / \dt + 1)}$, we have
  \begin{align*}
    \vecnorm{\policy (M_1) - \policy (M_2)}{\mathbb{L}^\infty (\torus)} \leq L \matsnorm{M_1 - M_2}{\infty}.
  \end{align*}
  Furthermore, for any $M \in \real^{N \times (\tburn / \dt + 1)}$, the image of the policy satisfies Lipschitz condition
  \begin{align*}
    \eucnorm{\policy (M) (x_1) - \policy (M) (x_2)} \leq L_1 \eucnorm{x_1 - x_2}, \quad \mbox{for any} ~x_1, x_2 \in \torus.
  \end{align*}
Besides, there exists a compact parameter set $\Theta\subset\R^p$ and a continuous map
$\theta\mapsto \pi_\theta$ such that $\Pi=\{\pi_\theta:\theta\in\Theta\}$.
\end{assumption}
Let us now first state the result on the performance of the learned policy $\ctrlE^{\widehat{\policy}}_t$ in terms of this population risk.

\begin{theorem}\label{thm:main-theorem}
  Under assumptions~\ref{assume:initial-regularity} and~\ref{assume:policy-class}, there exists a function $c_3: [0, T] \to \mathbb{R}_+$ such that the Vlasov--Poisson dynamics~\eqref{eq:Vlasov--Poisson} controlled by the learned policy $\ctrlE^{\widehat{\policy}}_t$ satisfies the electrical energy decay estimate
  \begin{align*}
    \vecnorm{\internalE^{\widehat{\policy}}_t}{\ltwospace (\torus)}^2 \leq c_1 e^{- c_2 t} + c_3 (t) \cdot \bigg\{ \inf_{\policy \in \policyclass} \mathcal{R} (\policy) + \sqrt{\frac{\log \cover (\epsilon, \policyclass, \vecnorm{\cdot}{\infty})}{\numobs}} + \epsilon \bigg\},
  \end{align*}
  where the constants $c_1, c_2 > 0$ are the same as in \Cref{prop:expert-stability}, and $\cover (\epsilon, \policyclass, \vecnorm{\cdot}{\infty})$ is the $\epsilon$-covering number of the policy class $\policyclass$ under the $\mathbb{L}^\infty$ norm.
\end{theorem}
\noindent See \Cref{subsec:proof-main-theorem} for the proof of this theorem.

A few remarks are in order. First, \Cref{thm:main-theorem} shows that the learned policy $\ctrlE^{\widehat{\policy}}_t$ achieves exponential stabilization of the Vlasov--Poisson dynamics, where the error floor depends on the approximation error $\inf_{\policy \in \policyclass} \mathcal{R} (\policy)$ and a standard error term in statistical learning theory that depends on the complexity of the policy class $\policyclass$ and the number of samples $\numobs$. Second, the function $c_3 (t)$ characterizes how the error floor grows with time. In general, $c_3 (t)$ may grow exponentially with time, which comes from Gr\"{o}nwall inequalities and distribution shift in behavior cloning. However, if the approximation error $\inf_{\policy \in \policyclass} \mathcal{R} (\policy)$ is sufficiently small, the theorem still guarantees stability over a relatively long time horizon. It is important direction of future work to study whether the exponential blow-up is unavoidable.

\Cref{thm:main-theorem} establishes the central role of the minimal population risk $\inf_{\policy \in \policyclass} \mathcal{R} (\policy)$ in determining the performance of the learned policy. In the next section, we present a concrete upper bound for this approximation error.

\subsection{Approximation theory for the control policy}\label{subsec:approx-theory}
The bounds in \Cref{thm:main-theorem} relies on how well the expert policy can be approximated using the observation history, quantified as $\inf_{\policy \in \policyclass} \mathcal{R} (\policy)$. In this section, we present a concrete example of policy class $\policyclass$ and derive explicit approximation error bounds. Furthermore, we show how the excess risk of the learned control policy adapts to the complexity of the initial distribution.

To motivate our theory, let us consider two extreme cases of the initial condition $\ffunc_0$ and their implications on the learnability of the expert policy.
\begin{itemize}
  \item If the initial density $\ffunc_0$ is deterministic, then the expert control action at time $t$ is exactly determined by solving the Vlasov--Poisson equation~\eqref{eq:Vlasov--Poisson} up to time $t$. The learning algorithm does not need any observation to predict the expert action. In such a case, the problem degenerates to learning a time-dependent function $\ctrlE^{\mathrm{expert}}_t (x)$.
  \item If the initial distribution $\initDistr$ has large entropy (i.e. an arbitrary density satisfying \Cref{assume:initial-regularity} without additional structures), the electric field $\internalE_t$ at time $t$ may still be predictable from the observation history $\Dset [t - \tburn, t]$. However, the best prediction error is limited by the number of sensors and observation frequency.
\end{itemize}
It is natural to use entropy to characterize such a difference in the learnability of the expert policy. As $\initDistr$ is defined over space of functions, we use the following notion of \emph{$\varepsilon$-resolution entropy} to characterize its complexity. In particular, we define
\begin{align*}
  H_\varepsilon (\initDistr) \mydefn  - \int \log \initDistr \big( \ball_{\ltwospace}(\ffunc, \varepsilon) \big) d \initDistr (f), \qquad \mbox{for any } \varepsilon > 0.
\end{align*}
Such a notion of entropy is a scale-sensitive generalization of the classical Shannon entropy, and it quantifies the complexity of the distribution $\initDistr$ at scale $\varepsilon$.

Besides, the learnability of the expert policy depends crucially on the spatial resolution of the observations. To ensure that the macroscopic density can be stably inverted from discrete sensor data, we make the following mild assumption regarding the sensor geometry. 
  We assume that the spatial sensors are placed at fixed uniform locations 
\begin{equation}
  \label{assume:sensor-distribute}
  x_j = j/N \in \torus \mbox{ for $j = 1, \dots, N$}.
\end{equation}

Under this setup, we are ready to present the approximation theory for the control policy.
\begin{theorem}\label{thm:approximation}
  There exists a policy class $\policyclass$ satisfying \Cref{assume:policy-class}, such that for any $p \leq q - 2$ and any initial distribution $\initDistr$ whose support satisfies \Cref{assume:initial-regularity}, the minimal population risk $\inf_{\policy \in \policyclass} \mathcal{R} (\policy)$ is upper bounded by
  \begin{align*}
    \inf_{\policy \in \policyclass} \mathcal{R} (\policy) \leq c \big( \stepsize^2 + e^{- c_0 \Nsensors} \big) + c \inf_{k \in \mathbb{N}_+, \varepsilon > 0}\Big\{   (k \stepsize)^{2p} + \varepsilon^2 + \frac{\sigma^2}{k} \cdot \frac{H_{\varepsilon} (\initDistr) + \log (\totaltime / \varepsilon)}{N} \Big\},
  \end{align*}
  where $c, c_0 > 0$ are constants depending on the regularity parameters in \Cref{assume:initial-regularity} and the integer constant $p$.
\end{theorem}
\noindent See \Cref{subsec:proof-approximation} for the proof of this theorem.

A few remarks are in order. The error bound consists of three parts. The first term $c \big( \stepsize^2 + e^{- c_0 \Nsensors} \big)$ characterizes the error due to the discrete-time observation and the spatially discrete sensors. They are unavoidable even for noiseless sensor readings. The second and third parts exhibits a trade-off through the choice of the parameter $k$ -- by our construction, the policies in the class $\Pi$ uses observation history of length $k \stepsize$ to predict the control action at time $t$. A longer observation history (i.e. larger $k$) allows the policy to reduce the variance term $\sigma^2 / k$ by averaging over more observations, but it also increases the bias term $(k \stepsize)^{2p}$ due to the temporal variation of the control action. The term $(k \stepsize)^{2p}$ is due to the the degradation of the utility of older data. On the other hand, for regular enough initial conditions, we can take $p$ to be large, and the bias term can be made negligible. The last part of the error takes the form
\begin{align*}
  \inf_{\varepsilon > 0} \Big\{ \varepsilon^2 + \frac{\sigma^2}{k} \cdot \frac{H_{\varepsilon} (\initDistr) + \log (\totaltime / \varepsilon)}{N} \Big\},
\end{align*}
which is standard in statistical learning theory, corresponding to the Bayes risk of a nonparametric estimation problem with $N$ samples and noise variance $\sigma^2 / k$, where the prior distribution is given by $\initDistr$. If we know the prior distribution $\initDistr$ and can sample from it, this is often the best possible error for learning the initial phase-space density $f_0$. However, the imitation learning setup does not allow us to collect direct samples from $f_0$, and we can only indirectly learn $f_0$ through the observation history. Furthermore, the learning algorithm only uses sample trajectories from the initial distribution $\nu_0$, without explicit knowledge of the distribution itself. Our result nevertheless shows that such desirable error is still achievable using indirect observations.

It is worth noting that the policy class $\policyclass$ does not depend on the initial distribution $\initDistr$. In other words, we can use a single learning model to learn control policies for a wide range of initial distributions, and the performance of the learned policy will automatically adapt to the complexity of the initial distribution as characterized by the $\varepsilon$-resolution entropy $H_\varepsilon (\initDistr)$. This shows the advantage of imitation learning approach over standard methods based on Bayesian inverse problems, which require explicit knowledge of the prior distribution $\initDistr$ to achieve the optimal error. Finally, we remark that the policy class $\policyclass$ constructed in the proof of \Cref{thm:approximation} mimicking an exponential weight aggregation estimator~\cite{dalalyan2008aggregation}, which can be implemented using a two-layer neural network with a sufficiently large hidden layer.

\section{Simulations Results}\label{sec:simulations}
In this section, we empirically evaluate our proposed control policy on the one-dimensional Vlasov--Poisson system.
\subsection{Experimental Setup}\label{subsec:sim-setup}
\paragraph{Dynamics and Physical Parameters.}
We solve the Vlasov--Poisson system on a periodic spatial domain $x\in[0, L_x)$ with length $L_x = 10\pi$, and a truncated velocity domain $v\in[-v_{\max},v_{\max}]$ with $v_{\max}=6$. The phase space is discretized on a high-resolution uniform grid with $n_x = n_v = 1024$. The system is integrated using a semi-Lagrangian scheme with a time step $\Delta t = 0.02$. We evolve the system up to a final time $T = 80$, which consists of a burn-in period of $\tburn = 1$ during which no control is applied.
\paragraph{Equilibrium and Initial Perturbation.}
We consider the classical two-stream Maxwellian equilibrium, defined as:
\begin{equation}
  \mu(v;\bar v) = \frac{1}{2\sqrt{2\pi}} \exp\Big(-\frac{(v-\bar v)^2}{2}\Big) + \frac{1}{2\sqrt{2\pi}} \exp\Big(-\frac{(v+\bar v)^2}{2}\Big),
\end{equation}
where we fix the stream separation parameter at $\bar v = 2.4$. To challenge our control policy, we construct the initial state using a dual-mode perturbation:
\begin{equation}
  f_0(x,v) = \mu(v;\bar v) \big( 1 + \varepsilon \eta(x) \big),
\end{equation}
where the spatial perturbation $\eta(x)$ superposes a macroscopic low-frequency mode and a microscopic high-frequency mode:
\begin{equation}
  \eta(x) = \cos(\frac{k_1 2\pi x}{L_x}   + \phi_1) + \cos(\frac{k_2 2\pi x}{L_x} + \phi_2).
\end{equation}
We fix the amplitude $\varepsilon = 0.05$, and the mode indices $k_1 = 1$ and $k_2 = 5$. The phase shifts $\phi_1, \phi_2 \sim \mathrm{Unif}(0, 2\pi)$ are randomized for each trajectory to evaluate the robustness of the policy.
\paragraph{Extreme Sparse Sensing.}
To rigorously evaluate the robustness of our approach, we deliberately push the system into an extremely sparse observation regime. Density measurements are recorded by merely $N = 4$ equally spaced sensors along the spatial domain. At any discrete time $t_k$, the controller receives a temporal observation window of length $K = 50$:
\begin{equation}
\mathcal D_j \;=\; \Big(\widetilde\rho_{i,j-\ell}\Big)_{\substack{1\le i\le N\\0\le \ell\le K-1}} \in \mathbb{R}^{N\times K}.
\end{equation}
\subsection{Neural Network Architecture and Training Details}\label{subsec:sim-nn}
To map the sparse observation window $\mathcal D_j$ to the control field $H_{t_j}(x)$, we design the following neural network architecture.
\paragraph{Network Architecture.}
We first augment the static density measurements $\widetilde{\rho}_{i,j - \ell}$ with their first-order temporal differences $\Delta \widetilde{\rho}_{i,j - \ell}$. The augmented input is then processed by a causal Temporal Convolutional Network (TCN) to extract local temporal patterns, followed by a Multi-Head Attention (MHA) module to capture global temporal dependencies over the window length $K$. The latent features are mean-pooled and mapped via a multi-layer perceptron (MLP) to $2M_f$ truncated Fourier coefficients (with $M_f = 8$ in our experiment), which can be straightforwardly transformed into the spatial control field $H_{t_j}(x)$ via the inverse Fourier transform.
\paragraph{Scale-Aware Hybrid Loss Function.}
Since our training dataset is generated from the unforced free-streaming evolution, the internal electric field undergoes severe amplitude decay over time due to Landau damping. Consequently, the late-stage electric fields are orders of magnitude smaller than the initial transients. A standard Mean Squared Error (MSE) objective would overwhelmingly bias the network towards the early-stage high-amplitude dynamics, entirely neglecting the fine-grained precision required for long-term stabilization.\\
To resolve this temporal scale disparity, we train the network using a scale-aware hybrid loss function:
\begin{equation}
    \mathcal{L} = \alpha \mathcal{L}_{\text{abs}} + \beta \mathcal{L}_{\text{rel}},
\end{equation}
where $\alpha$ and $\beta$ are balancing hyperparameters. Let $c_{k,j}$ denote the ground-truth Fourier coefficients for mode $k$ at the $j$-th step within the prediction window, and let $\hat{c}_{k,j}$ denote the corresponding network predictions. The absolute and relative error terms are defined as:
\begin{equation}
    \mathcal{L}_{\text{abs}} = \frac{1}{K} \sum_{j=1}^K \sum_{k=1}^{2M_f} w_k \big| \hat{c}_{k,j} - c_{k,j} \big|^2, \qquad
    \mathcal{L}_{\text{rel}} = \frac{1}{K} \sum_{j=1}^K \sum_{k=1}^{2M_f} w_k \frac{\big| \hat{c}_{k,j} - c_{k,j} \big|^2}{|c_{k,j}|^2}.
\end{equation}
where $w_k \propto k^{-1}$ is a static decay weight applied to the $k$-th Fourier mode.

\subsection{Reference Baselines}\label{subsec:sim-baselines}
To assess the contribution of our learned controller, we compare its performance against three distinct physical and mathematical baselines.
\paragraph{(B0) Uncontrolled Dynamics.}
We set the external control field to zero globally, i.e., $H_t(x) \equiv 0$. This baseline illustrates the intrinsic instability of the perturbed two-stream equilibrium. 
\paragraph{(B1) Instantaneous Spectral Poisson Reconstruction.}
This baseline reconstructs the control field using only instantaneous sparse observations ($K=1$, $N=4$). We transform the subsampled density into the Fourier domain, solve the discrete Poisson equation, and zero-pad the high-frequency spectrum to the full $n_x=1024$ resolution. An inverse FFT yields the reconstructed field $\widetilde{E}_{t}(x)$, and the control is applied as $H_{t}(x) = - \widetilde{E}_{t}(x)$.
\paragraph{(B2) Expert Policy.}
This controller computes the exact internal electric field $E_t^{\text{true}}(x)$ from the full-resolution density grid without any sensor sparsity, applying the optimal stabilizing control after burn-in $H_t(x) = - E_t^{\text{true}}(x) \mathbb{I}_{t \ge t_0}$.
\subsection{Main Results: Stabilization via Sparse Sensing}\label{subsec:sim-results}

We evaluate the stabilization performance using two primary metrics: the macroscopic endogenous electric energy $\mathcal{E}(t) = \frac{1}{2} \int_0^{L_x} |E_t(x)|^2 \, dx$, and the microscopic phase-space distribution $f_t(x,v)$.

\subsubsection{Stabilization under Clean Observations.}
\begin{figure}[htbp]
    \centering
    \includegraphics[width=0.98\linewidth]{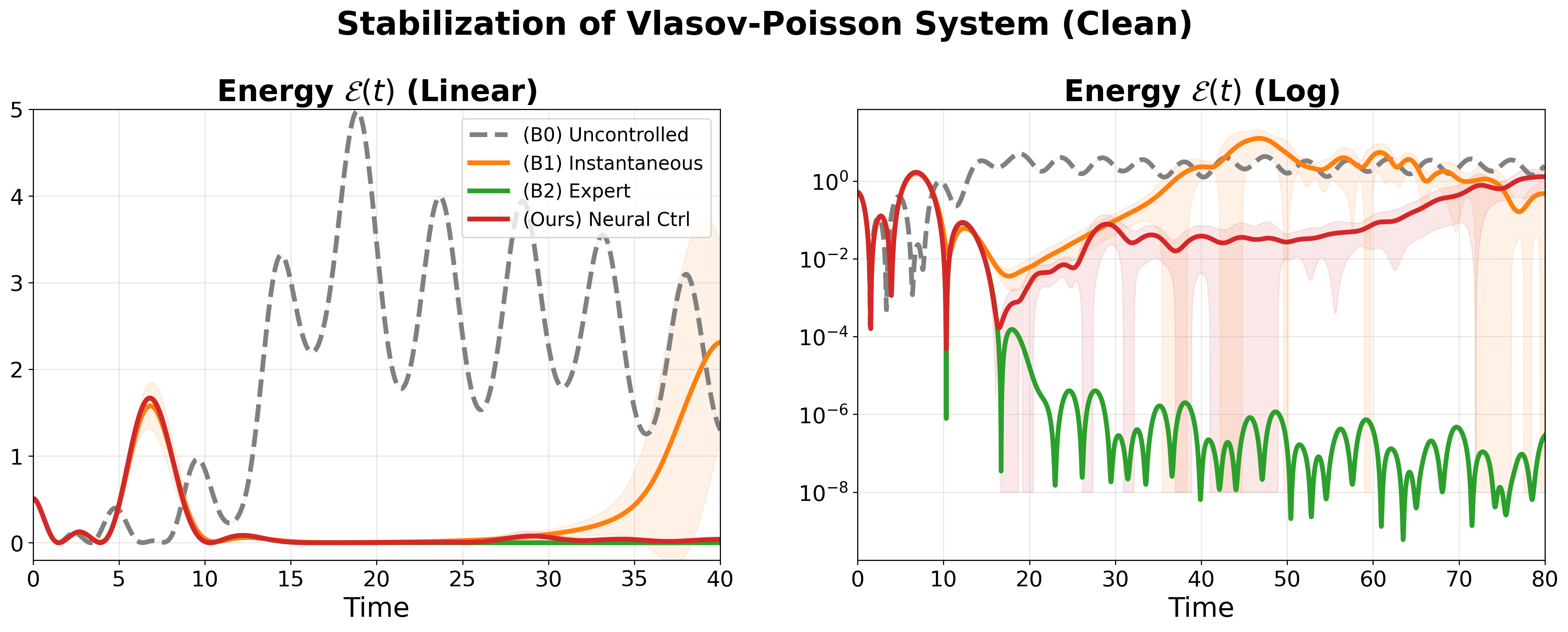}
    \caption{\textbf{Electric-energy stabilization under clean observations.} The Neural Controller overcomes the aliasing-induced divergence of the instantaneous baseline (B1) and closely tracks the Expert upper bound over the main control window.}
    \label{fig:clean-stabilization-energy}
\end{figure}

\begin{figure}[htbp]
    \centering
    \includegraphics[width=0.98\linewidth]{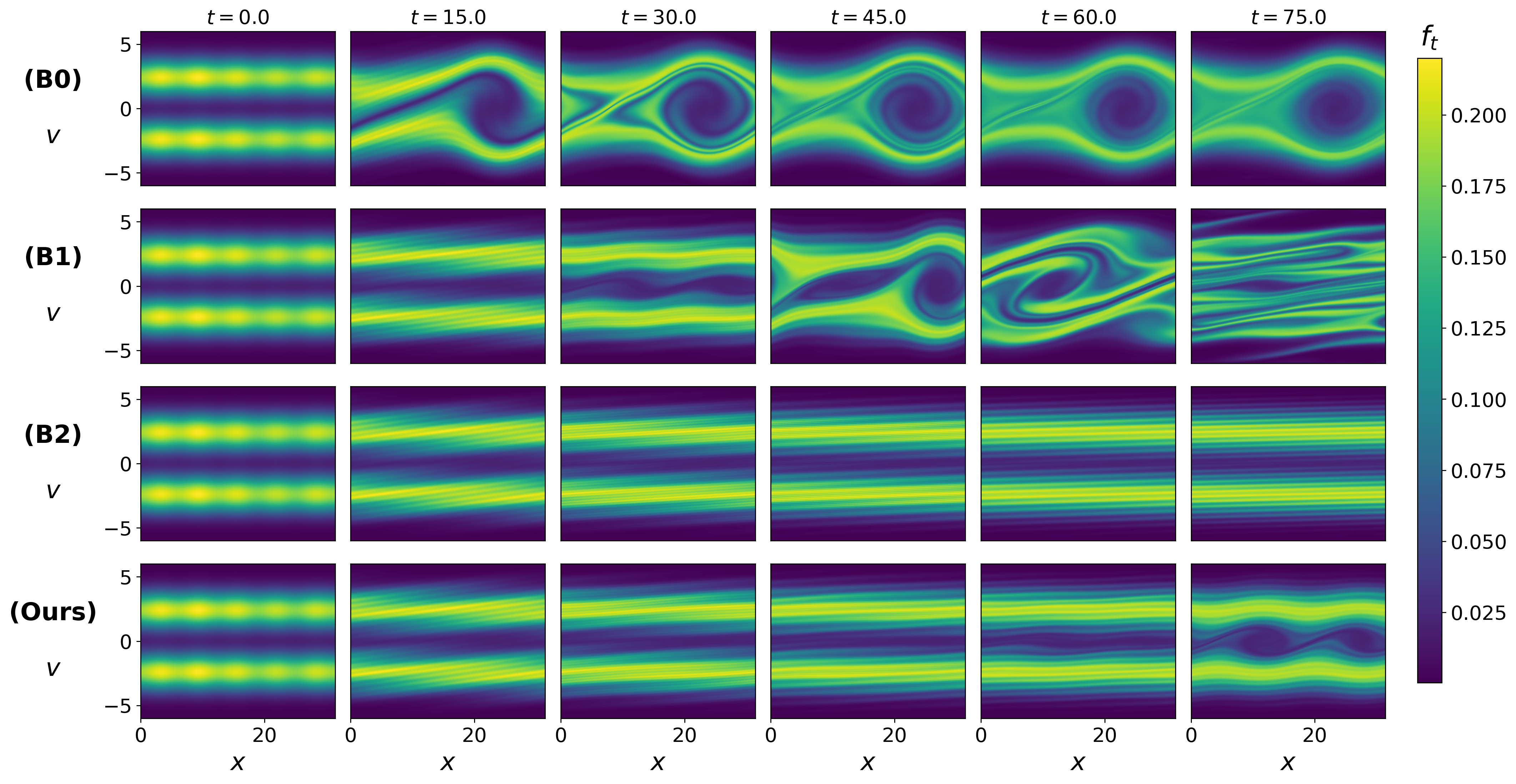}
    \caption{\textbf{Phase-space stabilization under clean observations.} The snapshots show that our method suppresses vortex formations and restores the uniform equilibrium, matching the qualitative behavior of the Expert policy.}
    \label{fig:clean-stabilization-snapshots}
\end{figure}

Figure~\ref{fig:clean-stabilization-energy} presents the closed-loop energy evolution under noise-free conditions. The uncontrolled baseline (B0) exhibits exponential growth followed by non-linear saturation. The instantaneous spectral baseline (B1) initially suppresses the instability, but quickly diverges at $t = 40$ due to severe aliasing errors from the sparse observations. In contrast, our Neural Controller significantly delays the onset of instability, maintaining stable control up to $t = 70$. Figure~\ref{fig:clean-stabilization-snapshots} further confirms at the phase-space level that our method suppresses vortex formations and restores the uniform equilibrium, closely mirroring the expert policy's performance.

\subsubsection{Robustness to Observational Noise and Implicit Regularization}\label{subsec:robustness}

\begin{figure}[htbp]
    \centering
    \includegraphics[width=0.98\linewidth]{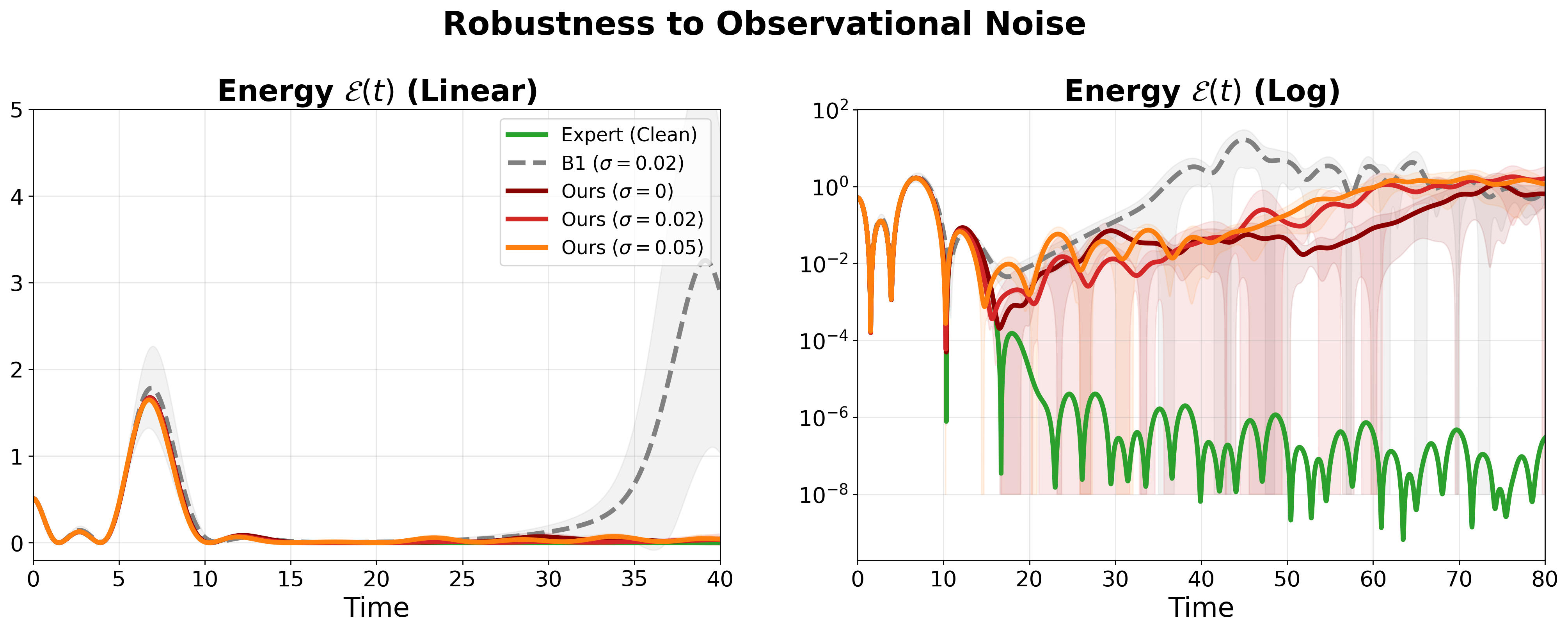}
    \caption{\textbf{Electric-energy robustness against observational noise.} The Neural Controller significantly delays the onset of instability compared to the B1 baseline, essentially doubling the stable control window across the tested noise scales.}
    \label{fig:noise-robustness-energy}
\end{figure}

\begin{figure}[htbp]
    \centering
    \includegraphics[width=0.98\linewidth]{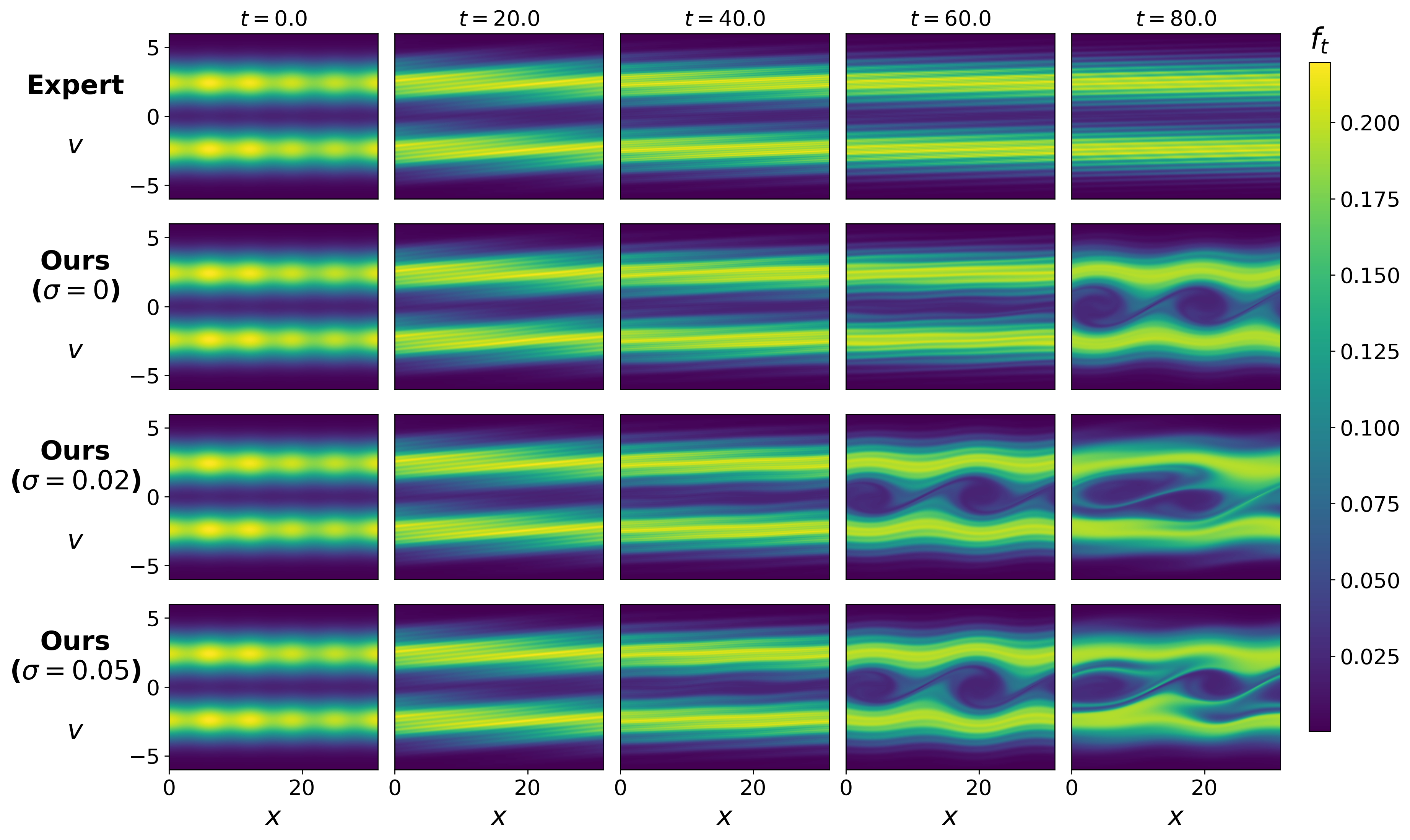}
    \caption{\textbf{Phase-space robustness against observational noise.} The snapshots reveal a decoupling effect: severe training noise ($\sigma=0.05$) implicitly regularizes the network, resulting in smoother large-scale structures during the failure regime ($t=80$) compared to the sharp filamentations seen in the moderate-noise ($\sigma=0.02$) case.}
    \label{fig:noise-robustness-snapshots}
\end{figure}

Figure~\ref{fig:noise-robustness-energy} presents the closed-loop energy evolution under varying levels of observational noise. Compared to the B1 baseline, our Neural Controller significantly delays the onset of instability, maintaining robust energy suppression up to $t = 50$ across different noise scales. Figure~\ref{fig:noise-robustness-snapshots} shows the corresponding phase-space snapshots and reveals an implicit regularization effect from severe training noise.

\section{Proofs}\label{sec:proofs}
We collect the proofs of the main results in this section. Some technical lemmas are deferred to the appendix for clarity.
\subsection{Proof of \Cref{prop:expert-stability}}\label{subsec:proof-expert-stability}
First, we prove the following lemma on the decay of the electric field for the free–stream transport equation with general initial conditions. This type of results are classical in the Landau damping literature~\cite{mouhot2011landau}, but we provide a self-contained proof for completeness.
\begin{lemma}\label{lemma:free-stream-decay-general}
Let $f$ solve the free-stream transport equation on $\T\times\R$
\begin{equation}\label{eq:freestream}
\partial_t f + v \partial_x f = 0,\qquad f|_{t=0}=f_0 .
\end{equation}
Assume that $f_0$ satisfies Assumption~\ref{assume:initial-regularity} with parameters
$\lambda_0,B_0$, and in addition the total mass is normalized:
\begin{equation}\label{eq:mass-normalization}
\int_{\T\times\R} f_0(x,v)\,dv\,dx = 1 .
\end{equation}
Then there exist constants $c_1,c_2>0$ depending only on $(\lambda_0,B_0)$ such that
\begin{equation}\label{eq:freestream-E-decay}
\|E_t\|^2_{L^2(\T)} \le c_1 e^{-c_2 t},\qquad \forall t\ge 0.
\end{equation}
\end{lemma}
\noindent See \Cref{subsubsec:proof-free-stream-decay-general} for the proof of this lemma.

The following result guarantees that within a short time interval, the distribution remains in the Gaussian mixture class if the initial condition is in the class, even for the uncontrolled Vlasov--Poisson dynamics.
\begin{lemma}\label{lemma:short-time-vp}
Suppose the initial condition $\ffunc_0$ satisfies Assumption~\ref{assume:initial-regularity} with an initial analyticity radius $\lambda_0 > 0$, a velocity decay parameter $q \ge 3$, and a uniform analytic norm bound $\vecnorm{\ffunc_0}{\mathcal{E}_q(\lambda_0)} \leq B_0$. Let $\ffunc_t$ be the solution to the uncontrolled 1D Vlasov--Poisson system on $t \in [0, \tburn]$.
Then, there exist deterministic constants $T^* > 0$ and $\kappa > 0$, depending only on $\lambda_0$, $B_0$, and $q$, such that for any chosen burn-in time $\tburn \leq T^*$, the solution $\ffunc_t$ remains analytic up to time $\tburn$. Specifically, defining the dynamically shrinking radius
\begin{equation*}
    \lambda_t \mydefn \frac{\lambda_0 - \kappa t}{1+t},
\end{equation*}
we have $\lambda_{\tburn} > 0$, and the analytic norm of the solution is strictly bounded by the initial constant:
\begin{equation}
  \sup_{0 \le t \le \tburn} \vecnorm{\ffunc_t}{\mathcal{E}_q(\lambda_t)} \leq B_0.
  \label{eq:local-solution-regularity-analytic}
\end{equation}
\end{lemma}

\noindent See \Cref{subsubsec:proof-short-time-vp} for the proof of this lemma.

Combining \Cref{lemma:free-stream-decay-general} and \Cref{lemma:short-time-vp}, we conclude that for a burn-in period satisfying
\begin{align*}
  \tburn \leq \min\Big\{ T^*, \frac{\lambda_0}{2\kappa} \Big\},
\end{align*}
we have that $\vecnorm{\ffunc_t}{\mathcal{E}_q(\tfrac{\lambda_0}{2 (1 + \tburn)})} \leq B_0$ for all $t \in [0, \tburn]$, which implies the exponential decay of the electric field for $t \geq \tburn$ by \Cref{lemma:free-stream-decay-general}. This completes the proof of \Cref{prop:expert-stability}.

\subsubsection{Proof of \Cref{lemma:free-stream-decay-general}}\label{subsubsec:proof-free-stream-decay-general}
The free-stream equation admits the explicit solution
\begin{align}\label{eq:fs-sol-explicit}
f_t(x,v)=f_0(x-tv,v).
\end{align}
For $k\in\Z$, denote the $x$-Fourier coefficients
\[
\widehat{g}(k):=\int_{\T} g(x)\,e^{-2\pi i k x}\,dx.
\]
Writing the Poisson equation $\partial_x E_t = 1-\rho_t$ in Fourier variables gives, for $k\neq 0$,
\begin{equation}\label{eq:Ek-rhok-clean-1d}
\widehat{E_t}(k)=-\frac{1}{2\pi i k}\,\widehat{\rho_t}(k).
\end{equation}
Hence by Parseval,
\begin{equation}\label{eq:E-L2-Fourier-clean-1d}
\|E_t\|_{L^2(\T)}^2
=\sum_{k\neq 0}|\widehat{E_t}(k)|^2
= \frac{1}{4\pi^2}\sum_{k\neq 0}\frac{|\widehat{\rho_t}(k)|^2}{k^2}.
\end{equation}
Define the mixed Fourier transform of $f_0$ by
\[
\widehat f_0(k,\eta)
:=\int_{\T}\int_{\R} f_0(x,v)\,e^{-2\pi i(k x+\eta v)}\,dv\,dx,
\qquad k\in\Z,\ \eta\in\R.
\]
Using \eqref{eq:fs-sol-explicit} and the change of variables $y=x-tv$, we have
\begin{equation}\label{eq:rhohat-fhat-clean-1d}
\widehat{\rho_t}(k)
=\int_{\T}\int_{\R} f_0(x-tv,v)\,e^{-2\pi i k x}\,dv\,dx
=\widehat f_0(k,kt).
\end{equation}
Since
\[
\|f_0\|_{\mathcal E(\lambda_0)}
=\sum_{m=0}^\infty \frac{\lambda_0^m}{m!}\,\|f_0\|_{\mathcal C^m}
\le B_0,
\]
since all summannds are nonnegative, it follows that for every $m\in\mathbb{N}$,
\begin{equation}\label{eq:Cm-from-E-1d}
\|f_0\|_{\mathcal C^m}\le B_0\, m!\,\lambda_0^{-m}.
\end{equation}
Fix $k\in\Z$ and $\eta\in\R\setminus\{0\}$. By the polynomial velocity weight in $\|f_0\|_{\mathcal C^m}$, the boundary terms at $|v|\to\infty$ vanish,
so integrating by parts $m$ times in $v$ yields
\[
\widehat f_0(k,\eta)
=(2\pi i\eta)^{-m}\int_{\T}\int_{\R} \partial_v^{m} f_0(x,v)\,
e^{-2\pi i(k x+\eta v)}\,dv\,dx,
\]
and therefore
\begin{equation}\label{eq:fhat-ibp-bound-1d}
|\widehat f_0(k,\eta)|
\le (2\pi|\eta|)^{-m}\int_{\T}\int_{\R} |\partial_v^{m} f_0(x,v)|\,dv\,dx.
\end{equation}
Using the weight built into $\|\cdot\|_{\mathcal C^m_q}$ and $\int_{\R}(1+|v|)^{-q}\,dv<\infty$ (since $q \ge 2 > 1$),
we obtain
\[
\int_{\T}\int_{\R} |\partial_v^{m} f_0(x,v)|\,dv\,dx
\le C_*\,\|f_0\|_{\mathcal C^m_q},
\]
where $C_*>0$ is a universal constant. Combining with
\eqref{eq:fhat-ibp-bound-1d} and \eqref{eq:Cm-from-E-1d} yields
\begin{equation}\label{eq:fhat-m-bound-clean-1d}
|\widehat f_0(k,\eta)|
\le C_*\,B_0\, m!\,(\lambda_0|\eta|)^{-m},
\qquad \forall m\in\mathbb{N}.
\end{equation}
Fix $\delta = \frac{\lambda_0}{2}$, multiplying by $\delta^m/m!$ and summing over $m\ge 0$ yields
\[
e^{\delta|\eta|}\,|\widehat f_0(k,\eta)|
\le C_*B_0\sum_{m=0}^\infty\Big(\frac{\delta}{\lambda_0}\Big)^m
=\frac{C_*B_0}{1-\delta/\lambda_0} = 2 C_*B_0.
\]
Thus there exist constants $C_0,c_0>0$ depending only on $(\lambda_0,B_0)$ such that
\begin{equation}\label{eq:fhat-exp-clean-1d}
|\widehat f_0(k,\eta)|\le C_0\,e^{-c_0|\eta|},\qquad \forall k\in\Z,\ \eta\in\R.
\end{equation}
In particular, by \eqref{eq:rhohat-fhat-clean-1d}, for $k\neq 0$,
\[
|\widehat{\rho_t}(k)|=|\widehat f_0(k,kt)|\le C_0 e^{-c_0|k|t}.
\]
Substituting into \eqref{eq:E-L2-Fourier-clean-1d} gives
\begin{equation}\label{eq:E-sum-clean-1d}
\|E_t\|_{L^2(\T)}^2
\le \frac{C_0^2}{4\pi^2}\sum_{k\neq 0}\frac{e^{-2c_0|k|t}}{k^2}.
\end{equation}
Since $|k|\ge 1$ for $k\neq 0$,
\[
\sum_{k\neq 0}\frac{e^{-2c_0|k|t}}{k^2}
\le 2 e^{-2c_0 t}\sum_{k\neq 0}\frac{1}{k^2}
= \frac{\pi^2}{3}\,e^{-2c_0 t}.
\]
Therefore, $\|E_t\|_{L^2(\T)}^2 \le c_1 e^{-c_2 t}$ for all $t\ge 0$, for some $c_1,c_2>0$ depending only on $(\lambda_0,B_0)$.
\subsubsection{Proof of Lemma~\ref{lemma:short-time-vp}}\label{subsubsec:proof-short-time-vp}
Define the shifted distribution:
\begin{equation*}
  g_t(x,v) \defn \ffunc_t(x+vt, v).
\end{equation*}
Applying the chain rule yields $\partial_t g_t = (\partial_t \ffunc_t + v \partial_x \ffunc_t)|_{(x+vt, v)}$. Thus, the PDE \eqref{eq:Vlasov--Poisson} transforms into a purely forced transport equation:
\begin{equation}\label{eq:g-pde}
  \partial_t g_t(x,v) = \widetilde{E}_t(x,v) \big( \partial_v - t \partial_x \big) g_t(x,v),
\end{equation}
where the shifted electric field is defined as $\widetilde{E}_t(x,v) \defn E_t(x+vt)$. To bound the nonlinear term on right hand side, we firstly establish a Banach algebra property tailored to our functional space. By the multi-index Leibniz rule, for any multi-index $\alpha$ with $|\alpha| = m$, the derivative of a product could be expanded as:
\begin{equation*}
  \partial^\alpha (U V) = \sum_{\beta \le \alpha} \binom{\alpha}{\beta} \partial^\beta U \, \partial^{\alpha-\beta} V.
\end{equation*}
Multiplying by the velocity weight $(1+|v|)^q$ and applying the triangle inequality, we have
\begin{equation*}
  (1+|v|)^q \big|\partial^\alpha (U V)\big| \le \sum_{\beta \le \alpha} \binom{\alpha}{\beta} \big|\partial^\beta U\big| \cdot (1+|v|)^q \big|\partial^{\alpha-\beta} V\big|.
\end{equation*}
To pass to the supremum bounds, we group the summation terms by their derivative order $k = |\beta|$. For each term in the sum, we bound the factors by their respective global suprema: $|\partial^\beta U| \le \|U\|_{\mathcal{C}_0^k}$ and $(1+|v|)^q |\partial^{\alpha-\beta} V| \le  \| V\|_{\mathcal{C}_q^{m-k}}$. Factoring these uniform bounds out of the inner summation yields:
\begin{equation*}
  (1+|v|)^q \big|\partial^\alpha (U V)\big| \le \sum_{k=0}^m \Bigg( \sum_{\substack{\beta \le \alpha \\ |\beta|=k}} \binom{\alpha}{\beta} \Bigg) \| U\|_{\mathcal{C}_0^k} \| V\|_{\mathcal{C}_q^{m-k}}.
\end{equation*}
Using the identity $\sum_{\beta \le \alpha, |\beta|=k} \binom{\alpha}{\beta} = \binom{|\alpha|}{k} = \binom{m}{k}$ and taking the supremum over $(x,v)$ and the maximum over all multi-indices $|\alpha| = m$ on the left-hand side, we obtain the rigorous uniform bound at derivative order $m$:
\begin{equation*}
  \|U V\|_{\mathcal{C}_q^m} \le \sum_{k=0}^m \binom{m}{k} \| U\|_{\mathcal{C}_0^k} \|V\|_{\mathcal{C}_q^{m-k}}.
\end{equation*}
Multiplying both sides by $\frac{\lambda^m}{m!}$ and summing over all $m \ge 0$, the $m!$ in the denominator perfectly cancels the factorial from the binomial expansion $\binom{m}{k} = \frac{m!}{k!(m-k)!}$. By regrouping the summation indices via the Cauchy product formula, we conclude the Banach algebra property:
\begin{align}\label{eq:banach-algebra}
  \|U V\|_{\mathcal{E}_q(\lambda)} 
  &\le \sum_{m=0}^\infty \frac{\lambda^m}{m!} \sum_{k=0}^m \frac{m!}{k!(m-k)!} \| U\|_{\mathcal{C}_0^k} \| V\|_{\mathcal{C}_q^{m-k}} \nonumber \\
  &= \sum_{m=0}^\infty \sum_{k=0}^m \left( \frac{\lambda^k}{k!} \| U\|_{\mathcal{C}_0^k} \right) \left( \frac{\lambda^{m-k}}{(m-k)!} \| V\|_{\mathcal{C}_q^{m-k}} \right) \nonumber \\
  &= \Bigg( \sum_{k=0}^\infty \frac{\lambda^k}{k!} \| U\|_{\mathcal{C}_0^k} \Bigg) \Bigg( \sum_{j=0}^\infty \frac{\lambda^j}{j!} \| V\|_{\mathcal{C}_q^{j}} \Bigg) \nonumber \\
  &= \|U\|_{\mathcal{E}_0(\lambda)} \|V\|_{\mathcal{E}_q(\lambda)}.
\end{align}
We first estimate the analytic norm of the macroscopic field $\widetilde{E}_t$. Note that $\partial_x \widetilde{E}_t(x,v) = (\partial_x E_t)(x+vt)$ and $\partial_v \widetilde{E}_t(x,v) = t (\partial_x E_t)(x+vt)$. Assuming $t \leq 1$, any mixed derivative satisfies $|\partial_{x,v}^\alpha \widetilde{E}_t| \leq |\partial_x^{|\alpha|} E_t(x+vt)|$. Therefore, the unweighted analytic norm of the field is bounded by the series of its pure spatial derivatives:
\begin{equation*}
  \|\widetilde{E}_t\|_{\mathcal{E}_0(\lambda)} \le \sum_{m=0}^\infty \frac{\lambda^m}{m!} \| E_t\|_{\mathcal{C}_0^m}.
\end{equation*}
For $m \ge 2$, the spatial derivatives annihilate the background constant $1$, yielding:
\begin{equation*}
  \|E_t\|_{\mathcal{C}_0^m} = \| \rho_t\|_{\mathcal{C}_0^{m-1}} \le \sup_{x,v}\int_{\real} |\partial_x^{m-1} g_t| \, dv \leq C_q \max_{|\beta|=m-1} \sup_{x,v} (1+|v|)^q |\partial_{x,v}^\beta g_t| = C_q \|g_t\|_{\mathcal{C}_q^{m-1}}.
\end{equation*}
where $C_q \defn \int (1+|v|)^{-q} dv < \infty$ for $q \ge 3$. \\
For $m = 1$, the constant remains in the bound:
\begin{equation*}
  \| E_t\|_{\mathcal{C}_0^1} = \|\partial_x E_t\|_{\mathbb{L}^{\infty}} \le 1 + \|\rho_t\|_{\mathbb{L}^{\infty}} \le 1 + C_q \|g_t\|_{\mathcal{C}_q^0}.
\end{equation*}
For $m = 0$, the zero-mean condition of the electric field on the torus yields a Poincaré-type inequality. Bounding the field by the integral of its gradient gives:
\begin{equation*}
  \|E_t\|_{\mathcal{C}_0^0} \le C_0 \|\partial_x E_t\|_{\mathbb{L}^{\infty}} \le C_0(1 + \|\rho_t\|_{\mathbb{L}^{\infty}}) \le C_0(1 + C_q \|g_t\|_{\mathcal{C}_q^0}).
\end{equation*}
Substituting these three cases back into the analytic norm series, we obtain for all $\lambda \le \lambda_0$:
\begin{align}\label{eq:field-bound}
  \|\widetilde{E}_t\|_{\mathcal{E}_0(\lambda)} 
  &\leq C_0(1 + C_q \|g_t\|_{\mathcal{C}_q^0}) + \lambda(1 + C_q \|g_t\|_{\mathcal{C}_q^0}) + \sum_{m=2}^\infty \frac{\lambda^m}{m!} C_q \| g_t\|_{\mathcal{C}_q^{m-1}} \nonumber \\
  &= (C_0 + \lambda) + (C_0 + \lambda) C_q \|g_t\|_{\mathcal{E}_q(\lambda)} + \lambda C_q \sum_{k=1}^\infty \frac{\lambda^k}{(k+1)!} \| g_t\|_{\mathcal{C}_q^{k}} \nonumber \\
  &\leq (C_0 + \lambda) + (C_0 + 2\lambda) C_q \|g_t\|_{\mathcal{E}_q(\lambda)} \nonumber \\
  &\leq C_E \big( 1 + \|g_t\|_{\mathcal{E}_q(\lambda)} \big),
\end{align}
where $C_E \defn \max(C_0+\lambda_0, \, C_0 C_q + 2\lambda_0 C_q)$ is a constant depends only on $\lambda_0$. Here we used the inequality $\frac{1}{(k+1)!} \le \frac{1}{k!}$ to absorb the higher-order tail into the full analytic norm $\|g_t\|_{\mathcal{E}_q(\lambda)}$.
\\
Next, we construct the Lyapunov energy estimate. Let the dynamic radius be $\lambda(t) \coloneqq \lambda_0 - \kappa t$. To compute the time derivative of the norm $\|g_t\|_{\mathcal{E}_q(\lambda(t))}$, we compute the upper right Dini derivative , which we still denote as $\frac{d}{dt}$ for brevity. Using $\sup |A| - \sup |B| \le \sup |A - B|$, can we conclude $D^+_t \sup |\partial^\alpha g_t| \le \sup |\partial_t \partial^\alpha g_t|$. Applying this rule to the series yields:
\begin{equation}
  \frac{d}{dt} \|g_t\|_{\mathcal{E}_q(\lambda(t))} \leq - \kappa \|\nabla_{x,v} g_t\|_{\mathcal{E}_q(\lambda(t))} + \|\partial_t g_t\|_{\mathcal{E}_q(\lambda(t))}. \label{eq:lyapunov-deriv}
\end{equation}
Applying the Banach algebra property \eqref{eq:banach-algebra} to the PDE \eqref{eq:g-pde}, the second term is bounded by:
\begin{align*}
  \|\partial_t g_t\|_{\mathcal{E}_q(\lambda(t))} 
  &\leq \|\widetilde{E}_t\|_{\mathcal{E}_0(\lambda(t))} \|(\partial_v - t \partial_x) g_t\|_{\mathcal{E}_q(\lambda(t))} \\
  &\leq C_E \big(1 + \|g_t\|_{\mathcal{E}_q(\lambda(t))}\big) \cdot 2 \|\nabla_{x,v} g_t\|_{\mathcal{E}_q(\lambda(t))}.
\end{align*}
Substituting this into \eqref{eq:lyapunov-deriv}, we establish the fundamental differential inequality:
\begin{equation*}
  \frac{d}{dt} \|g_t\|_{\mathcal{E}_q(\lambda(t))} \leq \Big( 2 C_E \big(1 + \|g_t\|_{\mathcal{E}_q(\lambda(t))}\big) - \kappa \Big) \|\nabla_{x,v} g_t\|_{\mathcal{E}_q(\lambda(t))}.
\end{equation*}
We fix the decay rate as $\kappa \coloneqq 4 C_E (1 + B_0)$. Define the following stopping time for the bootstrap argument:
\begin{equation*}
  T^* \coloneqq \sup \Big\{ T \in [0, \tburn] \;\Big|\; \sup_{t \in [0, T]} \|g_t\|_{\mathcal{E}_q(\lambda(t))} \leq B_0 \Big\}.
\end{equation*}
Since the initial condition satisfies $\|g_0\|_{\mathcal{E}_q(\lambda_0)} = \|\ffunc_0\|_{\mathcal{E}_q(\lambda_0)} \leq B_0$ and the map $t \mapsto \|g_t\|_{\mathcal{E}_q(\lambda(t))}$ is continuous, we have $T^* > 0$. \\
For any $t \in [0, T^*)$, the bootstrap hypothesis $\|g_t\|_{\mathcal{E}_q(\lambda(t))} \leq B_0$ ensures that the bracket in the differential inequality is strictly bounded away from zero:
\begin{equation*}
  2 C_E \big(1 + \|g_t\|_{\mathcal{E}_q(\lambda(t))}\big) - \kappa \leq 2 C_E (1 + B_0) - 4 C_E (1 + B_0) = -2 C_E (1 + B_0) < 0.
\end{equation*}
Consequently, the differential inequality reduces to $\frac{d}{dt} \|g_t\|_{\mathcal{E}_q(\lambda(t))} \leq 0$ on $[0, T^*)$. This implies that the analytic norm is monotonically non-increasing on this interval, yielding $\|g_t\|_{\mathcal{E}_q(\lambda(t))} \leq \|g_0\|_{\mathcal{E}_q(\lambda_0)} \leq B_0$. \\
Assume that $T^* < \tburn$. By continuity, the norm must exactly reach the boundary at $T^*$, meaning $\|g_{T^*}\|_{\mathcal{E}_q(\lambda(T^*))} = B_0$. However, because the derivative is non-positive whenever the norm is at or below $B_0$, the norm can never strictly grow to cross this threshold if it starts at $\leq B_0$. Thus, the bound is completely stabilized, and the maximal time $T^*$ cannot be strictly less than $\tburn$ without yielding a mathematical contradiction. Therefore, we conclude that $T^* = \tburn$, ensuring that $\|g_t\|_{\mathcal{E}_q(\lambda_t)} \leq B_0$ strictly holds for the entire interval $t \in [0, \tburn]$.\\
Finally, we map the bounds back to the Eulerian coordinates. Since $\ffunc_t(x,v) = g_t(x-vt, v)$, the chain rule dictates that any $m$-th order mixed derivative of $\ffunc_t$ is a linear combination of derivatives of $g_t$ multiplied by factors up to $t^m$. Hence, $|D^m \ffunc_t| \leq (1+t)^m \max_{|\beta|=m} |D^\beta g_t|$. This implies:
\begin{equation*}
  \|\ffunc_t\|_{\mathcal{E}_q(\lambda_t)} \leq \|g_t\|_{\mathcal{E}_q(\lambda_t(1+t))}.
\end{equation*}
By equating the dynamic radii $\lambda_t(1+t) = \lambda_0 - \kappa t$, we obtain exactly $\lambda_t = \frac{\lambda_0 - \kappa t}{1+t}$. We choose $T^* = \min(1, \frac{\lambda_0}{2\kappa})$. For any chosen burn-in time $\tburn \leq T^*$, we guarantee that the Eulerian analytic radius remains strictly positive: $\lambda_{\tburn} \ge \frac{\lambda_0}{4} > 0$. The Eulerian analytic norm satisfies $\|\ffunc_{\tburn}\|_{\mathcal{E}_q(\lambda_{\tburn})} \leq B_0$, which completes the proof.

\subsection{Proof of \Cref{thm:main-theorem}}\label{subsec:proof-main-theorem}
The proof consists of two parts: we first show an excess risk bound for the empirical risk minimizer $\widehat{\policy}$, and then we bound the stability of the learned policy using such excess risk bound. The two parts of the results are given in the following lemmas.

\begin{lemma}\label{lemma:excess-risk-bound}
Under the setup of \Cref{thm:main-theorem}, we denote the trajectory dataset $S = \{z^{(i)}\}_{i =1}^{n}$. Then for any $\delta\in(0,1)$, there exist deterministic constants
\[
C_{\mathrm{disc}} = C_{\mathrm{disc}}(\delta, B_0,\lambda_0,L,\sigma_\rho,\totaltime,\tburn,n,N,\eta,C_{\mathrm{bias}})
\quad\text{and}\quad
B = B(\delta, B_0,\lambda_0,L,\sigma_\rho,\totaltime,\tburn,n,N,\eta,C_{\mathrm{bias}})
\]
such that with probability at least $1-\delta$,
\begin{align*}
\mathcal{R}(\widehat{\pi})
\le
\inf_{\pi\in\Pi}\mathcal{R}(\pi)
\;+\; 2\,\mathfrak{R}_n(\mathcal{L}\circ\Pi)
\;+\; B\sqrt{\frac{2\log(8/\delta)}{n}}
\;+\; 2\,C_{\mathrm{disc}}\,\eta.
\end{align*}
where
\begin{align*}
\mathfrak{R}_n(\mathcal{L}\circ\Pi)
:=
\E_{S,\varepsilon}
\left[
\sup_{\pi\in\Pi}
\frac{1}{n}
\sum_{j=1}^n
\varepsilon_j
\ell(\pi,z^{(j)})
\right],
\end{align*}
and $\varepsilon_1,\dots,\varepsilon_n$ are i.i.d.\ Rademacher random variables
taking values in $\{-1,+1\}$ with equal probability. Moreover, one may take $B$ of the form
\[
B =
\Big( L\big(C_{\mathrm{phy}}+\sigma_\rho\sqrt{2\log\!\tfrac{4\,nN(\lfloor \totaltime/\eta\rfloor+1)}{\delta}}\big) + C_{\mathrm{bias}}
\Big)^2,
\]
where $C_{\mathrm{phy}}$ is the constant given in Lemma~\ref{lemma:free-stream-decay-general} and $C_{\mathrm{bias}}$ is the finite quantity defined in the proof below.
\end{lemma}
\noindent The proof of \Cref{lemma:excess-risk-bound} is given in \Cref{subsubsec:proof-excess-risk-bound}.

\begin{lemma}\label{lemma:stability-via-excess-risk}
  Under the setup of \Cref{thm:main-theorem}, for any policy $\policy \in \policyclass$. Let $\epsilon_{\mathrm{stat}} = \mathcal{R}(\policy)$ be the population risk of the learned policy. For any $\delta \in (0,1)$, there exist constants $C_1, C_2, C_3 > 0$ depending on the $B_0, \lambda_0$, such that with probability at least $1 - \delta$, for all $t \ge \tburn$ satisfying $C_3\,\exp\!\big(C_3(1+t)^{11}\big)\,\frac{\epsilon_{\mathrm{stat}}^2}{\delta} \le \frac12$, the corresponding solution $\ffunc^{\policy}$ to the Vlasov--Poisson equation~\eqref{eq:Vlasov--Poisson} satisfies
  \begin{align*}
    \vecnorm{\internalE_t^{\policy}}{\ltwospace (\torus)} \le
C_1 e^{-C_2 t}
+ C_1\,\frac{\epsilon_{\mathrm{stat}}}{\sqrt{\delta}} (1+t)^{2} \exp(C_1(1+t)^{11})
+ C_1\,\frac{\epsilon^2_{\mathrm{stat}}}{\delta} (1+t)^{5} \exp(C_1(1+t)^{11}).
  \end{align*}
\end{lemma}
\noindent The proof of \Cref{lemma:stability-via-excess-risk} is given in \Cref{subsubsec:proof-stability-via-excess-risk}.

\subsubsection{Proof of \Cref{lemma:excess-risk-bound}}\label{subsubsec:proof-excess-risk-bound}

In this proof we denote $\hat R(\pi):=\frac1n\sum_{j=1}^n \ell(\pi,z^{(j)})$ and $\hat R_{\eta}(\pi) = \frac1n\sum_{j=1}^n \hat\ell_\eta(\pi,z^{(j)})$.\\
We start by defining a good event on which the noise and loss are uniformly bounded. 
Let $N$ be the number of sensors and let $K$ be the number of time points used inside each
window.
Fix $\delta_0\in(0,1)$ to be specified later, define
\[
\mathcal{E}_{\mathrm{noise}}
\coloneqq
\Big\{
\max_{1\le j\le n}\max_{1\le i\le N}\max_{0\le k\le \lfloor \totaltime/\eta\rfloor}
|W^{(j)}_{i,k}|
\le
\sigma_\rho\sqrt{2\log\frac{4nN(\lfloor \totaltime/\eta\rfloor+1)}{\delta_0}}
\Big\}.
\]
By a standard tail bound for Gaussian random variables and union bound, we have
\[
\mathbb{P}(\mathcal{E}_{\mathrm{noise}})\ge 1-\delta_0.
\]
On $\mathcal{E}_{\mathrm{noise}}$, fix any $\pi\in\Pi$ and any trajectory sample $z$.
For $t\in[t_0,T]$, write
\[
J(\pi,z,t)\coloneqq \big\|\pi(\mathcal D^z[t-t_0,t])+E^z(t)\big\|_{\mathbb{L}^2(\T)}^2,
\qquad \text{so that}\qquad
\ell(\pi,z)=\frac{1}{T-t_0}\int_{t_0}^T J(\pi,z,t)\,dt.
\]
Since $\T$ is bounded, we have $\|h\|_{\mathbb{L}^2(\T)}\le \|h\|_{\mathbb{L}^\infty(\T)}$. Hence
\begin{align*}
\sqrt{J(\pi,z,t)}
&\le \|\pi(\mathcal D^z[t-t_0,t])\|_{\mathbb{L}^\infty(\T)} + \|E^z(t)\|_{\mathbb{L}^\infty(\T)} .
\end{align*}
We next bound the two terms on the right-hand side. The compact parameterization in Assumption~\ref{assume:policy-class} implies that $C_{\mathrm{bias}}\coloneqq \sup_{\pi\in\Pi}\|\pi(0)\|_{\mathbb{L}^\infty(\T)} $ is finite. For the policy term, by Assumption~\ref{assume:policy-class} and the triangle inequality,
\begin{align}
\|\pi(\mathcal D^z[t-t_0,t])\|_{\mathbb{L}^\infty(\T)}
&\le \|\pi(\mathcal D^z[t-t_0,t])-\pi(0)\|_{\mathbb{L}^\infty(\T)}
     +\|\pi(0)\|_{\mathbb{L}^\infty(\T)}\notag\\
&\le L\,\|\mathcal D^z[t-t_0,t]\|_{\infty} +  C_{\mathrm{bias}}. \label{eq:pi-output-bound}
\end{align}
We furtherly bound $\|\mathcal D^z[t-t_0,t]\|_{\infty}$, using the triangle inequality yields:
\begin{align}
\|\mathcal D^z[t-t_0,t]\|_{\infty}
&=\max_{1\le i\le N}\max_{1 \le k\le K} |\widetilde\rho_{i,k}|
\le \sup_{s\in[0,T]}\|\rho_s^z(\cdot)\|_{\mathbb{L}^\infty(\T)}
   + \max_{i,k}|W_{i,k}|\nonumber\\
&\le \sup_{s\in[0,T]}\|\rho_s^z(\cdot)\|_{\mathbb{L}^\infty(\T)}
+\sigma_\rho\sqrt{2\log\frac{4nN(\lfloor T/\eta\rfloor+1)}{\delta_0}},
\label{eq:D-bound}
\end{align}
We then give bounds for $\rho$ and $E$. By Lemma~\ref{lemma:free-stream-decay-general}, for the free-streaming trajectories we have
deterministic constants
$C_{\mathrm{phy}}=C(B_0,\lambda_0) $
such that for all trajectories $z$ and all $s\in[0,T]$,
\begin{equation}\label{eq:PDE-sup-bounds}
\|\rho^z_s(\cdot)\|_{\mathbb{L}^\infty(\T)}\le C_{\mathrm{phy}},\qquad
\|E_s^z(\cdot)\|_{\mathbb{L}^\infty(\T)}\le C_{\mathrm{phy}} .
\end{equation}
Combining \eqref{eq:pi-output-bound}--\eqref{eq:PDE-sup-bounds}, on $\mathcal E_{\mathrm{noise}}$ we get
for all $t\in[t_0,T]$,
\[
\sqrt{J(\pi,z,t)}
\le \Bigg(L\Big(C_{\mathrm{phy}}+\sigma_\rho\sqrt{2\log\frac{4nN(\lfloor T/\eta\rfloor+1)}{\delta_0}}\Big)+C_{\mathrm{bias}}\Big)
+ C_{\mathrm{phy}}\Bigg).
\]
Squaring both sides yields on $\mathcal E_{\mathrm{noise}}$, we have
\begin{equation}\label{eq:B-delta-explicit}
J(\pi,z,t)\le B
\quad\text{and hence}\quad
0\le \ell(\pi,z)\le B,\qquad 0\le \hat\ell_\eta(\pi,z)\le B,
\end{equation}
where 
\[
B
\coloneqq
\Big(
L\Big(C_{\mathrm{phy}}+\sigma_\rho\sqrt{2\log\frac{4nN(\lfloor T/\eta\rfloor+1)}{\delta_0}}\Big)
+ C_{\mathrm{bias}}
\Big)^2.
\] 
We then bound the discretization error between $\ell$ and $\hat\ell_\eta$ by using the following technical Lemma.
\begin{lemma}
\label{lem:J-time-lipschitz}
Under the setup of Lemma~\ref{lemma:excess-risk-bound}, let the time grid be
\[
t_k \coloneqq t_0+k\eta,\qquad k=0,1,\dots,K,\qquad K\coloneqq \Big\lfloor \frac{T-t_0}{\eta}\Big\rfloor,
\]
and denote the grid cells $I_k\coloneqq [t_k,t_{k+1})$ for $k=0,\dots,K-1$. On the noise good event $\mathcal E_{\mathrm{noise}}$
defined in the proof of Lemma~\ref{lemma:excess-risk-bound},
there exists a deterministic constant $C$, such that for each trajectory $z$ and each $\pi\in\Pi$ 
\[
|J(\pi,z,t)-J(\pi,z,s)|
\le C\,|t-s|,
\qquad
\forall\,k\in\{0,\dots,K-1\},\ \forall\,t,s\in I_k,\ \forall\,\pi\in\Pi,\ \forall\,z.
\]
Where the constant $C$ depends on $B_0,\lambda_0,t_0,T,\sigma_\rho,n,N,\eta,\delta_0$.
\end{lemma}
See \Cref{subsubsec:proof-J-time-lipschitz} for the proof of \Cref{lem:J-time-lipschitz}. 
\\By \Cref{lem:J-time-lipschitz}, on $\mathcal E_{\mathrm{noise}}$
the map $t \mapsto J(\pi, z,t) $
is Lipschitz continuous on $[t_0,T]$ uniformly in $z$ and $\pi$.
Therefore, for each $\pi\in\Pi$ and each trajectory $z$,
we have
\begin{equation}\label{eq:disc-err}
\big|\ell(\pi,z)-\hat\ell_\eta(\pi,z)\big|
\le
C_{\mathrm{disc}}\,\eta,
\end{equation}
where $C_{\mathrm{disc}}=C/(T-t_0)$ and $C$ is the Lipschitz constant
in \Cref{lem:J-time-lipschitz}.
Consequently, for any $\pi\in\Pi$,
\[
\big|\hat R(\pi)-\hat R_{\eta}(\pi)\big|
=
\left|\frac1n\sum_{j=1}^n \big(\ell(\pi,z^{(j)})-\hat\ell_\eta(\pi,z^{(j)})\big)\right|
\le C_{\mathrm{disc}}\eta.
\]
In particular,
\begin{equation}\label{eq:disc-two-policies}
\hat R(\widehat\pi)\le \hat R_{\eta}(\widehat\pi)+C_{\mathrm{disc}}\eta,
\qquad
\hat R_{\eta}(\pi)\le \hat R(\pi)+C_{\mathrm{disc}}\eta.
\end{equation}
Let $\pi^\star\in\arg\min_{\pi\in\Pi} R(\pi)$.
Since $\widehat\pi$ minimizes $\hat R_{\eta}$, we have $\hat R_{\eta}(\widehat\pi)\le \hat R_{\eta}(\pi^\star)$.
Using \eqref{eq:disc-two-policies},
\[
\hat R(\widehat\pi)
\le
\hat R_{\eta}(\widehat\pi)+C_{\mathrm{disc}}\eta
\le
\hat R_{\eta}(\pi^\star)+C_{\mathrm{disc}}\eta
\le
\hat R(\pi^\star)+2C_{\mathrm{disc}}\eta.
\]
Therefore,
\begin{align}
R(\widehat\pi)-R(\pi^\star)
&=
\big(R(\widehat\pi)-\hat R(\widehat\pi)\big)
+
\big(\hat R(\widehat\pi)-\hat R(\pi^\star)\big)
+
\big(\hat R(\pi^\star)-R(\pi^\star)\big)\notag \\
&\le
\big(R(\widehat\pi)-\hat R(\widehat\pi)\big)
+
\big(\hat R(\pi^\star)-R(\pi^\star)\big)
+
2C_{\mathrm{disc}}\eta\notag \\
&\le
2\sup_{\pi\in\Pi}\big|R(\pi)-\hat R(\pi)\big|
+
2C_{\mathrm{disc}}\eta. \label{eq:R-Factorization}
\end{align}
It remains to control $\sup_{\pi\in\Pi}|R(\pi)-\hat R(\pi)|$. Work conditionally on $\mathcal{E}_{\mathrm{noise}}$.
Define
\[
\Phi(S)\coloneqq \sup_{\pi\in\Pi}\big(R(\pi)-\hat R(\pi)\big),
\qquad
S=\{z^{(1)},\dots,z^{(n)}\}.
\]
If we replace one sample $z^{(j)}$ by an independent copy $z^{(j)\prime}$, denote the new sample set by $S'$. Then since on $\mathcal{E}_{\mathrm{noise}}$ the loss is bounded in $[0,B]$, we have
\[
|\Phi(S)-\Phi(S')|
\le
\sup_{\pi\in\Pi}\frac1n\big|\ell(\pi,z^{(j)})-\ell(\pi,z^{(j)\prime})\big|
\le \frac{B}{n}.
\]
By McDiarmid's inequality, for any $\epsilon>0$,
\[
\mathbb{P}\Big(\Phi(S)-\E[\Phi(S)\mid \mathcal{E}_{\mathrm{noise}}]\ge \epsilon\ \Big|\ \mathcal{E}_{\mathrm{noise}}\Big)
\le
\exp\Big(-\frac{2n\epsilon^2}{B^2}\Big).
\]
Taking $\epsilon=B\sqrt{\frac{\log(4/\delta)}{2n}}$ yields that with conditional probability given $\mathcal{E}_{\mathrm{noise}}$ at least $1-\delta/2$,
\begin{equation}\label{eq:mcdiarmid}
\Phi(S)
\le
\E[\Phi(S)\mid \mathcal{E}_{\mathrm{noise}}]
+
B\sqrt{\frac{\log(4/\delta)}{2n}}.
\end{equation}
A standard symmetrization argument gives
\[
\E[\Phi(S)\mid \mathcal{E}_{\mathrm{noise}}]
\le
2\,\Rad_n(\mathcal{L}\circ\Pi),
\]
Combining with \eqref{eq:mcdiarmid} and  applying the argument to $-\ell$,
we obtain that on $\mathcal{E}_{\mathrm{noise}}$, with conditional probability at least
$1-\delta/2$,
\[
\sup_{\pi\in\Pi}\big|R(\pi)-\hat R_S(\pi)\big|
\le
2\,\Rad_n(\mathcal{L}\circ\Pi)
+
B\sqrt{\frac{\log(4/\delta)}{2n}}.
\]
Plugging the above bound into \Cref{eq:R-Factorization} gives on $\mathcal{E}_{\mathrm{noise}}$ we have
\[
R(\widehat\pi)
\le
R(\pi^\star)
+
2\,\Rad_n(\mathcal{L}\circ\Pi)
+
B\sqrt{\frac{2\log(4/\delta)}{n}}
+
2C_{\mathrm{disc}}\eta.
\]
Finally choose $\delta_0=\delta/2$ in the definition of $\mathcal{E}_{\mathrm{noise}}$.
Then $\mathbb{P}(\mathcal{E}_{\mathrm{noise}}^c)\le \delta/2$ so the desired inequality holds with overall probability at least $1-\delta$.

\subsubsection{Proof of \Cref{lemma:stability-via-excess-risk}}\label{subsubsec:proof-stability-via-excess-risk}

In the proof of \Cref{lemma:stability-via-excess-risk}, we will use the following technical lemma, which controls the spatial density for a transport equation under a perturbation in the $v$--direction.
\begin{lemma}\label{lemma:transport-perturbed-rho-bound}
  Consider the perturbed transport equation:
  \begin{align}\label{eq:perturbed-transport}
    \partial_t \ffunc_t + v \pt_x \ffunc_t + \delta_t(x) \pt_v \ffunc_t = 0.
  \end{align}
   Suppose the initial condition $\ffunc_0$ satisfies Assumption~\ref{assume:initial-regularity} with parameters $\lambda_0, B_0$ and $\int_{\mathbb T\times\mathbb R} f_0\,dv\,dx=1$.
  Define $D(t) \coloneqq (\int_0^t \vecnorm{\delta_\tau(\cdot)}{\mathbb{L}^{\infty}(\torus)}^2\,d\tau)^{1/2}$. Then there exist constants $C_1,C_2$ depending only on $B_0, \lambda_0$ such that for all $t\ge0$,
  \begin{align*}
  \|\rho_t(\cdot)-1\|_{\mathbb{L}^\infty(\mathbb T)}
  \le C_1e^{-C_2t}
+
C_1 D(t) (1 + t)^2
+
C_1\,(1+D(t)^3 + D(t)^3 t^2) D(t)^2t^3.
\end{align*}
\end{lemma}
See \Cref{subsubsec:proof-transport-perturbed-rho-bound} for the proof of this lemma.\\
Let the expert (free-stream) density window be $\mathcal D^{\mathrm{fs}}[t-t_0,t]$.
Define the \emph{expert-evaluated mismatch}
\[
\delta^{\mathrm{exp}}_t(x)
:=
H^\pi\!\big(\mathcal D^{\mathrm{fs}}[t-t_0,t],t\big)(x) + E_t^{\mathrm{fs}}(x).
\]
By the definition of the population risk $\mathcal{R}$, we have
\begin{equation}\label{eq:exp-risk-mean}
\mathbb E\Big[
\int_{t_0}^T \|\delta_t^{\mathrm{exp}}\|_{\mathbb{L}^2(\T)}^2\,dt
\Big]
\le \epsilon_{\mathrm{stat}}^2.
\end{equation}
Let
\[
Z
:=
\int_{t_0}^T
\|\delta_t^{\mathrm{exp}}\|_{\mathbb{L}^2(\T)}^2\,dt.
\]
By Markov's inequality, for any $\delta \in (0,1)$, define the event
\[
\mathcal G_\delta
:=
\Big\{
Z \le \epsilon_{\mathrm{stat}}^2 / \delta
\Big\}.
\]
Then
\begin{equation}\label{eq:good-event}
\mathbb P(\mathcal G_\delta)
\ge 1-\delta.
\end{equation}
In the remainder of the proof we work on the event $\mathcal G_\delta$.
Along the policy trajectory, define
\[
\delta_t(x):=H^\pi\!\big(\mathcal D[t-t_0,t],t\big)(x)+E_t(x),
\qquad
D(t)^2:=\int_{t_0}^t\|\delta_s\|_{\mathbb{L}^\infty(\T)}^2\,ds.
\]
Decompose
\begin{equation}\label{eq:delta-decomp-patch}
\delta_t
=
\underbrace{\Big(H^\pi(\mathcal D,t)-H^\pi(\mathcal D^{\mathrm{fs}},t)\Big)}_{=:S_t^{(1)}}
+
\underbrace{\Big(H^\pi(\mathcal D^{\mathrm{fs}},t)+E_t^{\mathrm{fs}}\Big)}_{=\delta_t^{\mathrm{exp}}}
+
\underbrace{(E_t-E_t^{\mathrm{fs}})}_{=:S_t^{(2)}},
\end{equation}
where we simplify the notation by denote $\mathcal D=\mathcal D[t-t_0,t]$ and
$\mathcal D^{\mathrm{fs}}=\mathcal D^{\mathrm{fs}}[t-t_0,t]$.\\
We firstly bound the term $S_t^{(1)}$.
By \Cref{assume:policy-class} we have
\[
\|S_t^{(1)}\|_{\mathbb{L}^\infty(\T)}
\le L\,\|\mathcal D-\mathcal D^{\mathrm{fs}}\|_\infty.
\]
Since the window matrix is formed by discrete samples of the partial density observations, we have (if we define $\| A\|_{\infty} := \max_{i,j}|A_{i,j}|$ for matrix $A$)
\[
\|\mathcal D-\mathcal D^{\mathrm{fs}}\|_\infty
\le \sup_{s\in[t-t_0,t]}\|\rho_s-\rho_s^{\mathrm{fs}}\|_{\mathbb{L}^\infty(\T)}.
\]
Hence
\begin{equation}\label{eq:shift1}
\|S_t^{(1)}\|_{\mathbb{L}^\infty(\T)}
\le C \sup_{s\in[t-t_0,t]}\|\rho_s-\rho_s^{\mathrm{fs}}\|_{\mathbb{L}^\infty(\T)}.
\end{equation}
We then turn to the term $S_t^{(2)}$.
Since $\partial_x(E_t-E_t^{\mathrm{fs}})=-(\rho_t-\rho_t^{\mathrm{fs}})$, and both electric fields have zero mean, interpolation inequality on $\T$ yields
\begin{equation}\label{eq:shift2}
\|S_t^{(2)}\|_{\mathbb{L}^\infty(\T)}
=
\|E_t-E_t^{\mathrm{fs}}\|_{\mathbb{L}^\infty(\T)}
\le C\|\partial_x(E_t-E_t^{\mathrm{fs}})\|_{\mathbb{L}^2(\T)}
= C\|\rho_t-\rho_t^{\mathrm{fs}}\|_{\mathbb{L}^2(\T)}
\le C\|\rho_t-\rho_t^{\mathrm{fs}}\|_{\mathbb{L}^\infty(\T)}.
\end{equation}
Fix any horizon $T>0$ such that $C\,\exp\!\big(C(1+T)^{11}\big)\,\frac{\epsilon_{\mathrm{stat}}^2}{\delta} \le \frac12,$, define $T^* = \sup \{t \in [t_0, T] \mid D(t) \le 1\}$.
By applying \Cref{lemma:transport-perturbed-rho-bound} on bootstrap interval $[t_0, T^*]$, we have for $t\in[t_0,T^*]$,
\begin{equation}\label{eq:rho-diff-blackbox}
\|\rho_t-\rho_t^{\mathrm{fs}}\|_{\mathbb{L}^\infty(\T)}
\le
C\Big((1+t)^{2}D(t) + (1+t)^5 D(t)^2\Big).
\end{equation}
We then give a bound for $D(t).$ Define
\[
y(t):=D(t)^2=\int_{t_0}^t \|\delta_s\|_{\mathbb{L}^\infty(\T)}^2\,ds,
\qquad\text{so that}\qquad
y'(t)=\|\delta_t\|_{\mathbb{L}^\infty(\T)}^2.
\]
We still work on the bootstrap interval $[t_0, T^*]$ where $y(t)\le 1$ on it. By $(a+b+c)^2\le 3(a^2+b^2+c^2)$ and \eqref{eq:delta-decomp-patch}, we obtain for $t\in[t_0,T^*]$,
\begin{equation}\label{eq:delta-infty-split}
\|\delta_t\|_{\mathbb{L}^\infty}^2
\le 3\|S_t^{(1)}\|_{\mathbb{L}^\infty}^2 + 3\|\delta_t^{\exp}\|_{\mathbb{L}^\infty}^2 + 3\|S_t^{(2)}\|_{\mathbb{L}^\infty}^2.
\end{equation}
Using \eqref{eq:shift1}--\eqref{eq:rho-diff-blackbox} and the monotonicity of $D(t)$, we obtain for $t\in[t_0,T^*]$,
\begin{align*}
\|S_t^{(1)}\|_{\mathbb{L}^\infty}^2 + \|S_t^{(2)}\|_{\mathbb{L}^\infty}^2 \le C \|\rho_t-\rho_t^{\mathrm{fs}}\|_{\mathbb{L}^\infty}^2
\le C\Big((1+t)^{2}D(t) + (1+t)^5D(t)^2\Big)^2.
\end{align*}
Since on the bootstrap region we have $D(t)\le 1$, we could further simplify the above bound by
\[
(1+t)^2D(t) + (1+t)^5D(t)^2 \;\le\; C(1+t)^{5} D(t),
\]
and \eqref{eq:delta-infty-split} simplifies to
\begin{equation}\label{eq:g-infty-simplified}
\|\delta_t\|_{\mathbb{L}^\infty}^2
\le 3 \|\delta_t^{\mathrm{exp}}\|^2_{\mathbb{L}^\infty(\T)} + C(1+t)^{10}D(t)^2.
\end{equation}
We then bound for the term involving $\delta_t^{\exp}$. Agmon inequality in one dimension gives
\[
\|\delta_t^{\exp}\|_{\mathbb{L}^\infty(\T)}^2 \le C\|\delta_t^{\exp}\|_{\mathbb{L}^2(\T)}\,\|\partial_x\delta_t^{\exp}\|_{\mathbb{L}^2(\T)}.
\]
Moreover, $\partial_x\delta_t^{\exp}=\partial_x H^\pi(\mathcal D^{\mathrm{fs}},t)+\partial_x E_t^{\mathrm{fs}}$.
By \Cref{assume:policy-class}, $\|\partial_x H^\pi(\mathcal D^{\mathrm{fs}},t)\|_{\mathbb{L}^2(\T)}\le L_1$, and since
$\partial_xE_t^{\mathrm{fs}}=1-\rho_t^{\mathrm{fs}}$ in $d=1$, we have
$\|\partial_xE_t^{\mathrm{fs}}\|_{\mathbb{L}^2(\T)}=\|1-\rho_t^{\mathrm{fs}}\|_{\mathbb{L}^2(\T)}\le C$ for any $t$ by \Cref{prop:expert-stability}.
Hence, we have
\begin{equation}\label{eq:exp-infty}
\|\delta_t^{\exp}\|_{\mathbb{L}^\infty(\T)}^2 \le C\,\|\delta_t^{\exp}\|_{\mathbb{L}^2(\T)}^2,
\end{equation}
 where $C$ depends only on $L_1$.
Plugging \eqref{eq:exp-infty}
into \eqref{eq:g-infty-simplified}, we obtain on the bootstrap region $[t_0, T^*]$,
\begin{align}\label{eq:y-ode}
y'(t)=\|\delta_t\|_{\mathbb{L}^\infty(\T)}^2
\le C \|\delta_t^{\exp}\|_{\mathbb{L}^2(\T)}^2 + C(1+t)^{10}y(t),
\qquad t\in[t_0,T^*),
\end{align}
Integrating the above ode and applying Gr\"onwall gives
\begin{equation}\label{eq:y-gronwall}
y(t)\le
C\int_{t_0}^t
\|\delta_s^{\mathrm{exp}}\|_{\mathbb{L}^2(\T)}^2\,
\exp\!\Big(C\int_s^t(1+\tau)^{10}d\tau\Big)\,ds,
\qquad t\in[t_0,T^*).
\end{equation}
Since $\int_s^t(1+\tau)^{10}d\tau \le C(1+t)^{11}$, we obtain the bound
\begin{equation}\label{eq:y-gronwall-coarse}
y(t)\le
C\,\exp\!\big(C(1+t)^{11}\big)\,
\int_{t_0}^t \|\delta_s^{\mathrm{exp}}\|_{\mathbb{L}^2(\T)}^2\,ds,
\qquad t\in[t_0,T^*).
\end{equation}
Since on $\mathcal{G}_\delta$, we have $\int_{t_0}^t \|\delta_s^{\mathrm{exp}}\|_{\mathbb{L}^2(\T)}^2\,ds \le \epsilon_{\mathrm{stat}}^2 / \delta$, we finally obtain
\begin{equation}\label{eq:y-final}
y(t)
\le
C\,\exp\!\big(C(1+t)^{11}\big)\,\frac{\epsilon_{\mathrm{stat}}^2}{\delta},
\qquad t\in[t_0,T^*).
\end{equation}
If $\epsilon_{\mathrm{stat}}$ and $T$ satisfy
\begin{equation}\label{eq:basecase-small}
C\,\exp\!\big(C(1+T)^{11}\big)\,\frac{\epsilon_{\mathrm{stat}}^2}{\delta} \le \frac12,
\end{equation}
then \eqref{eq:y-final} implies $y(t)\le 1/2$ on $[t_0,T^*)$. The continuity of $y(t)$ implies $T^*=T$ and $D(t) \le 1 $ holds on $[t_0,T]$. We finish the proof by applying \Cref{lemma:transport-perturbed-rho-bound} with $D(t) \le C \exp(C (1+t)^{11}) \frac{\epsilon_{\mathrm{stat}}}{\sqrt\delta}$.

\subsection{Proof of \Cref{thm:approximation}}\label{subsec:proof-approximation}
We prove the theorem constructively by providing a concrete instance of the policy class $\policyclass$ that achieves the desired approximation error bound, uniformly for any initial distribution $\ffunc_0$ in the support of $\Prob_0$. The construction consists of two steps: first we construct a single policy $\policy_0$ that achieves certain pointwise bias and variance bounds, and then we obtain the final policy $\policy$ by an exponential weight algorithm to further reduce the error.

First, let $\mathrm{Dir} (x) \mydefn \frac{\sin (\Nsensors x / 2)}{\sin (x / 2)}$ be the Dirichlet kernel, we define the linear subspace
\begin{align*}
  \LinSpace_{\Nsensors} \mydefn \big\{ \textstyle\sum_{i = 1}^{\Nsensors} c_i \mathrm{Dir} (x - x_i) ~: ~ c \in \real^{\Nsensors} \big\},
\end{align*}
which is the space of trigonometric polynomials of degree at most $\frac{\Nsensors - 1}{2}$ that interpolate the values at the sensor locations. We also define the projection operator $P_{\Nsensors}: \ltwospace (\torus) \rightarrow \LinSpace_{\Nsensors}$ as
\begin{align*}
  P_{\Nsensors} f := \tfrac{1}{\Nsensors} \textstyle\sum_{i = 1}^{\Nsensors} f(x_i) \mathrm{Dir} (x - x_i).
\end{align*}
Clearly, $P_{\Nsensors}$ is a linear operator, and we have the isometry property $\vecnorm{P_{\Nsensors} f}{\ltwospace (\torus)}^2 = \frac{1}{\Nsensors} \sum_{i = 1}^{\Nsensors} |f(x_i)|^2$ for any $f \in \ltwospace (\torus)$. The following lemma shows that the expert policy $\ctrlE_t^{\mathrm{expert}}$ can be well approximated by its projection $P_{\Nsensors} \ctrlE_t^{\mathrm{expert}}$ onto $\LinSpace_{\Nsensors}$.
\begin{lemma}\label{lemma:projection-approximation}
  For any $t \in [\tburn, \totaltime]$, we have
  \begin{align*}
    \vecnorm{(I - P_{\Nsensors}) \ctrlE_t^{\mathrm{expert}}}{\ltwospace} \leq \varepsilon_{\mathrm{tr}},
  \end{align*}
\end{lemma}
\noindent See \Cref{subsubsec:proof-projection-approximation} for the proof of \Cref{lemma:projection-approximation}.

For any $t \in [\tburn, \totaltime]$, and any control function $\ctrlE \in \LinSpace_N$, \Cref{lemma:projection-approximation} along with the Pythagorean theorem gives
\begin{align}
  \vecnorm{\ctrlE_t - \ctrlE_t^{\mathrm{expert}}}{\ltwospace}^2 = \vecnorm{\ctrlE_t - P_N \ctrlE_t^{\mathrm{expert}}}{\ltwospace}^2 + \vecnorm{(I - P_N) \ctrlE_t^{\mathrm{expert}}}{\ltwospace}^2 \leq \vecnorm{\ctrlE_t - P_N \ctrlE_t^{\mathrm{expert}}}{\ltwospace}^2 + \varepsilon_{\mathrm{tr}}^2.\label{eq:pythagorean-thm-in-approximation-decomp}
\end{align}
Therefore, it suffices to construct a policy $\policy$ that approximates $\ctrlE_t^{\mathrm{expert}}$ well within the subspace $\LinSpace_N$ to achieve the desired approximation error bound. The following lemma constructs such a policy $\policy_0$, but its estimation error may be large. We will later apply an exponential weight algorithm to further reduce the error and obtain the final policy $\policy$.
\begin{lemma}\label{lemma:building-block-in-policy-class-construction}
  Let $p = q - 2 \ge 1$ be the estimation order determined by the velocity decay parameter $q$ in Assumption~\ref{assume:initial-regularity}. For any $\Delta \in [\stepsize, \tburn]$, there exists a policy $\policy_0: \real^{N \times (\tburn / \stepsize + 1)} \rightarrow \LinSpace_N$, such that conditionally on the initial distribution $\ffunc_0$, for any $t \in [\tburn, \totaltime]$ and each sensor location $x_i$, $i = 1, \ldots, N$, the expectation of the policy output satisfies the pointwise bias bound:
  \begin{equation*}
    \abss{ \Exs \big[ \policy_0 (\Dset[t - \tburn, t]) \big] (x_i) - \ctrlE_t^{\mathrm{expert}} (x_i)} \leq C \big( \varepsilon_{\mathrm{tr}} + \Delta^p \big).
  \end{equation*}
  and the estimator is Gaussian with variance bound
  \begin{equation*}
    \policy_0 (\Dset[t - \tburn, t])  (x_i) - \Exs \big[\policy_0 (\Dset[t - \tburn, t]) \big] (x_i) \sim \mathcal{N}\big(0, v_{t,i}^2\big), \quad \mbox{with} \quad
    v_{t,i}^2 \leq C' \sigma^2 \frac{\stepsize}{\Delta},
  \end{equation*}
  for constants $C, C' > 0$ depending only on $p$ and the initial regularity constants.

  Furthermore, the policy $\policy_0$ is Lipschitz with respect to its input in the following sense: for any $Y, Y' \in \real^{N \times (\tburn / \stepsize + 1)}$, we have
  \begin{align*}
    \vecnorm{\policy_0 (Y) - \policy_0 (Y') }{\mathbb{L}^\infty(\torus)} \leq \frac{2^p - 1}{\pi} (1 + \ln M) \|Y - Y'\|_{\infty}.
  \end{align*}
\end{lemma}
\noindent The proof of \Cref{lemma:building-block-in-policy-class-construction} is given in \Cref{subsubsec:proof-building-block-in-policy-class-construction}.

Taking this lemma as given, we proceed with proof of \Cref{thm:approximation}. The proof is based on explicit construction of the policy class $\policyclass$ and the policy $\policy \in \policyclass$ that achieves the desired approximation error bound. To start with, we note that by combining \Cref{lemma:short-time-vp} and \Cref{lemma:free-stream-decay-general}, we have
\begin{align*}
  \sup_{x \in \torus} \abss{\ctrlE^{\mathrm{expert}}_t (x)} + \abss{\nabla_x \ctrlE^{\mathrm{expert}}_t (x)}  \leq L, \quad \forall t \in [\tburn, \totaltime],
\end{align*}
for some constant $L > 0$ depending only on parameters in Assumption~\ref{assume:initial-regularity}. 

Based on this, we define the exponential weight function
\begin{align*}
  \weightFun_{\temperature, \nu} (f; y) := \frac{\exp (- \Nsensors \vecnorm{f - y}{\ltwospace}^2 / \temperature) d \nu (f) }{\int \exp (- \Nsensors \vecnorm{h - y}{\ltwospace}^2 / \temperature) d \nu (h)},
\end{align*}
for a base measure $\nu$ on $\ltwospace (\torus)$ and a temperature parameter $\temperature > 0$. The policy class $\policyclass_\temperature$ is then defined as
\begin{align*}
  \policyclass_\temperature &:= \Big\{ M \mapsto \int f d w_{\temperature, \nu} (f; \policy_0 (M)) ~:~ \support (\nu) \subseteq \Fclass_{BL} (L) \cap \LinSpace_\Nsensors\Big\}.
\end{align*}
In other words, $\policyclass_\temperature$ is the class of policies obtained by applying exponential weight aggregation to the base policy $\policy_0$ with respect to any base measure $\nu$ supported on the bounded Lipschitz function class $\Fclass_{BL} (L)$.
The tuning parameter $\temperature$ will be chosen later.

The following lemma summarizes its basic regularity properties.
\begin{lemma}\label{lemma:regularity-of-policy-class}
  The function class $\policyclass_\temperature$ satisfies \Cref{assume:policy-class}. In particular, for any $\policy \in \policyclass_\temperature$, we have
  \begin{align*}
    \pi (\Dset) &\in \mathcal{F}_{BL} (L), \quad \mbox{and}\\
    \vecnorm{\pi (\Dset_1) - \pi (\Dset_2)}{\infty} &\leq \frac{c N^2 \log N}{\tau} \|\Dset_1 - \Dset_2\|_{\infty},
  \end{align*}
  where $c > 0$ is a constant depending only on $p$, and $\Dset, \Dset_1, \Dset_2$ are any input data matrices in $\real^{N \times (\tburn / \stepsize + 1)}$.
\end{lemma}
\noindent See \Cref{subsubsec:proof-regularity-of-policy-class} for the proof.

Now we construct the policy $\policy \in \policyclass_\temperature$ that achieves the desired approximation error bound. To start with, for any measurable set $\mathcal{A} \subseteq \LinSpace_\Nsensors$, we define the projected occupancy measure $\occupmsr$ and its discretized version $\nu_+$ as
\begin{align*}
  \occupmsr (\mathcal{A}) \mydefn \frac{1}{\totaltime - \tburn} \int_{\tburn}^{\totaltime} \Prob (P_\Nsensors (\ctrlE_t^{\mathrm{expert}}) \in \mathcal{A}) dt, \quad \nu_+ (\mathcal{A}) \mydefn \frac{\stepsize}{\totaltime - \tburn} \sum_{m = \lceil \tburn / \stepsize \rceil}^{\lfloor \totaltime / \stepsize \rfloor} \Prob (P_\Nsensors (\ctrlE^{\mathrm{expert}}_{m \stepsize}) \in \mathcal{A})
\end{align*}
We then define the policy $\policy$ to be a policy in $\Pi_\temperature$ induced by the discretized measure $\nu_+$ as the base measure $\nu$ in the exponential weight function. In the following, we bound its approximation error. To start with, by Parseval's identity, for any $f \in  \LinSpace_N$, we have
\begin{align*}
  \vecnorm{f}{\ltwospace}^2 = \frac{1}{N} \sum_{|k| \leq  (\Nsensors - 1) / 2}  |\widehat{f}(k)|^2 = \frac{1}{N} \sum_{i = 1}^\Nsensors f (x_i)^2 =: \vecnorm{f}{\Nsensors}^2.
\end{align*}
Define the idealized noise-free control action
\begin{align*}
  \ctrlE_t^* \mydefn \Exs \big[ \policy_0 \big(\Dset[t - \tburn, t]) \big]
\end{align*}
By the bias bound in \Cref{lemma:building-block-in-policy-class-construction}, we have
\begin{align*}
  \ltwonorm{\ctrlE_t^* - P_N \ctrlE_t^{\mathrm{expert}}} \leq C (\varepsilon_{\mathrm{tr}} + \Delta^p).
\end{align*}
By \Cref{lem:rho-time-derivative} and the definition of $\ctrlE_t^{\mathrm{expert}}$, we have
\begin{align*}
  \ltwonorm{\ctrlE_t^{\mathrm{expert}} - \ctrlE_{\lfloor t / \stepsize \rfloor \stepsize}^{\mathrm{expert}}} \leq c_1 \stepsize.
\end{align*}

On the other hand, by the noise characterization in \Cref{lemma:building-block-in-policy-class-construction}, denote $\zeta_t \mydefn \policy_0 (\Dset[t - \tburn, t]) - \ctrlE_t^*$, we have $\zeta_t (x_i) \sim \mathcal{N} (0, v_{t,i}^2)$ with $v_{t,i}^2 \leq C' \sigma^2 \frac{\stepsize}{\Delta}$, independently for each sensor location $x_i$.

By Theorem 1 of \cite{dalalyan2008aggregation}, by taking the temperature parameter
\begin{align*}
  \temperature = 4 C' \sigma^2 \frac{\stepsize}{\Delta},
\end{align*}
we have the estimation error bound for any distribution $q$ over $\ltwospace (\torus)$
\begin{align}
  \Exs \big[ \ltwonorm{\policy_{\temperature, \nu_+} \big(\Dset ([t - \tburn, t]) \big) - \ctrlE_t^*}^2 \mid \ffunc_0 \big] \leq \int \Exs \big[ \ltwonorm{f - \ctrlE_t^*}^2 \big] q(df) + \frac{\temperature}{\Nsensors} \kull{q}{\nu_+}.\label{eq:dalalyan-aggregation-bound}
\end{align}
We now analyze the right-hand-side of \Cref{eq:dalalyan-aggregation-bound}. In particular, given a time $t$ and an initial condition $\ffunc_0$, we choose $q$ as
\begin{align*}
  q_t (\mathcal{A} \mid \ffunc_0) \mydefn \nu_+ \big(\mathcal{A} \cap \ball_{\ltwospace} (P_N \ctrlE_{\lfloor t / \stepsize \rfloor \stepsize}^{\mathrm{expert}}, \varepsilon) \big) / \nu_+ \big( \ball_{\ltwospace} (P_N \ctrlE_{\lfloor t / \stepsize \rfloor \stepsize}^{\mathrm{expert}}, \varepsilon) \big),
\end{align*}
for any subset $\mathcal{A} \subseteq \LinSpace_\Nsensors$. In other words, $q$ is the conditional distribution of $\nu_+$ restricted to an $\varepsilon$-ball around $P_N \ctrlE_{\lfloor t / \stepsize \rfloor \stepsize}^{\mathrm{expert}}$. By the definition of $\ltwospace$ ball and \Cref{lemma:building-block-in-policy-class-construction}, we have
\begin{align*}
  \int \Exs \big[ \ltwonorm{\ctrlE - \ctrlE_t^*}^2 \big] q_t (d \ctrlE \mid \ffunc_0)\leq 2 \varepsilon^2 +  c\varepsilon_{\mathrm{tr}}^2 + c\Delta^{2p} + (c_1 \stepsize)^2.
\end{align*}
On the other hand, we have
\begin{align*}
  \kull{q_t (\cdot \mid \ffunc_0)}{\nu_+} = - \log \nu_+ \big( \ball_{\ltwospace} (P_N \ctrlE_t^{\mathrm{expert}}, \varepsilon) \big),
\end{align*}
and consequently, by averaging over the initial distribution $\initDistr$ and the time interval $[\tburn, \totaltime]$, we have the identity.
\begin{align*}
   \int  \frac{1}{\totaltime - \tburn} \int_{\tburn}^{\totaltime}  \kull{q_t (\cdot \mid \ffunc_0)}{\nu_+} dt~ d \initDistr (\ffunc_0) = H_{\varepsilon} (\nu_+).
\end{align*}
In the following lemma, we bound the $\varepsilon$-resolution entropy of the projected occupancy measure $\occupmsr$ using that of the initial measure $\initDistr$.
\begin{lemma}\label{lemma:resolution-entropy-bound}
 Under the setup of \Cref{thm:approximation}, for any $\varepsilon > 0$, we have
 \begin{align*}
  H_\varepsilon (\nu_+) \leq H_{\varepsilon / C} (\initDistr) + \log (\totaltime / \stepsize).
 \end{align*}
\end{lemma}
\noindent See \Cref{subsubsec:proof-resolution-entropy-bound} for the proof of \Cref{lemma:resolution-entropy-bound}.

Putting these bounds together, and substituting into \Cref{eq:dalalyan-aggregation-bound}, we have
\begin{multline*}
  \int_{\tburn}^{\totaltime} \Exs \big[ \ltwonorm{\policy_{\temperature, \nu_+} \big(\Dset ([t - \tburn, t]) \big) - \ctrlE_t^{\mathrm{expert}}}^2 \big] dt\\
   \leq c \Big\{ \varepsilon_\mathrm{tr}^2 +  \Delta^{2p} + \stepsize^2 \Big\} + c \inf_{\varepsilon > 0} \Big\{ \varepsilon^2 + \sigma^2 \frac{\stepsize}{\Delta} \cdot \frac{H_{\varepsilon} (\initDistr) + \log (\totaltime / \stepsize)}{\Nsensors} \Big\}.
\end{multline*}
Letting $k = \Delta / \stepsize$, we complete the proof by optimizing over $\varepsilon$ and $k$.

\subsubsection{Proof of \Cref{lemma:projection-approximation}}\label{subsubsec:proof-projection-approximation}
Recall that the expert policy corresponds to the true electric field, $H_t^{\mathrm{expert}} = E_t$. By the Poisson equation, its Fourier coefficients are given by $\widehat{E}_t(k) = \frac{\widehat{\rho}_t(k)}{-2\pi i k}$ for $k \neq 0$, and $\widehat{E}_t(0) = 0$. We decompose $E_t$ into a bandlimited polynomial of degree $M$ and a high-frequency tail:
\begin{equation*}
    E_t(x) = E_t^{(M)}(x) + E_t^{>M}(x) \defn \sum_{0 < |k| \le M} \widehat{E}_t(k) e^{2\pi i k x} + \sum_{|k| > M} \widehat{E}_t(k) e^{2\pi i k x}.
\end{equation*}
Since the number of sensors satisfies $N > 2M$, the Dirichlet projection operator $P_N$ acts as the exact identity operator on the subspace of trigonometric polynomials up to degree $M$. Consequently, $P_N E_t^{(M)} = E_t^{(M)}$, which leads to the error decomposition:
\begin{equation*}
    (I - P_N) \ctrlE_t^{\mathrm{expert}} = (I - P_N) \big(E_t^{(M)} + E_t^{>M}\big) = E_t^{>M} - P_N E_t^{>M}.
\end{equation*}
Applying the triangle inequality, we have $\vecnorm{(I - P_N) \ctrlE_t^{\mathrm{expert}}}{\ltwospace} \le \vecnorm{E_t^{>M}}{\ltwospace} + \vecnorm{P_N E_t^{>M}}{\ltwospace}$.\\
For the first term, we apply Parseval's identity. Leveraging the intrinsic smoothing effect of the Poisson operator where $|k| > M \ge 1$, we obtain:
\begin{align*}
    \vecnorm{E_t^{>M}}{\ltwospace}^2 &= \sum_{|k| > M} \frac{|\widehat{\rho}_t(k)|^2}{4\pi^2 k^2} \\
    &\le \frac{1}{4\pi^2 (M+1)^2} \sum_{|k| > M} |\widehat{\rho}_t(k)|^2 \\
    &= \frac{1}{4\pi^2 (M+1)^2} \vecnorm{\rho_t - \rho_t^{(M)}}{\ltwospace}^2.
\end{align*}
By Lemma~\ref{lemma:spatial-regularity-truncation}, the density truncation is bounded by $\varepsilon_{\mathrm{tr}}$, yielding $\vecnorm{E_t^{>M}}{\ltwospace} < \frac{1}{2\pi (M+1)} \varepsilon_{\mathrm{tr}}$.\\
For the second term, the pointwise bound $|P_N E_t^{>M}(x_j)| \le \sum_{|k|>M} |\widehat{E}_t(k)|$ combined with the exact isometry of $P_N$ immediately yields:
\begin{equation*}
    \vecnorm{P_N E_t^{>M}}{\ltwospace} \le \sum_{|k|>M} |\widehat{E}_t(k)|.
\end{equation*}
Applying the Poisson equation $\widehat{E}_t(k) = \frac{\widehat{\rho}_t(k)}{-2\pi i k}$ and the analytic regularity condition $|\widehat{\rho}_t(k)| \le B_0 e^{-\lambda_0 |k|}$, we bound this $\ell_1$-norm via a geometric series:
\begin{align*}
    \vecnorm{P_N E_t^{>M}}{\ltwospace} &\le 2 \sum_{k=M+1}^{\infty} \frac{B_0 e^{-\lambda_0 k}}{2\pi k} \\
    &\le \frac{B_0}{\pi (M+1)} \sum_{k=M+1}^{\infty} e^{-\lambda_0 k} \\
    &= \underbrace{\frac{B_0 e^{-\lambda_0}}{\pi (1 - e^{-\lambda_0})}}_{ \defn C_A} \frac{e^{-\lambda_0 M}}{M+1}.
\end{align*}
Recall from Lemma~\ref{lemma:spatial-regularity-truncation} that $M = \mathcal{O}(\ln(1/\varepsilon_{\mathrm{tr}}))$, which inherently provides $e^{-\lambda_0 M} \propto \varepsilon_{\mathrm{tr}}$. By absorbing the deterministic constant $C_A$ into the logarithmic definition of the truncation degree $M$, we strictly bound the aliasing error by $\mathcal{O}(\varepsilon_{\mathrm{tr}})$. Combining this with the continuous truncation error, we secure the uniform bound:
\begin{equation*}
    \vecnorm{(I - P_N) \ctrlE_t^{\mathrm{expert}}}{\ltwospace} \le \vecnorm{E_t^{>M}}{\ltwospace} + \vecnorm{P_N E_t^{>M}}{\ltwospace} \leq \varepsilon_{\mathrm{tr}},
\end{equation*}
which completes the proof.

\subsubsection{Proof of \Cref{lemma:building-block-in-policy-class-construction}}\label{subsubsec:proof-building-block-in-policy-class-construction}
In this section, we quantify how well the macroscopic density $\rho_t$ can be reconstructed from a noisy temporal window of spatial density sensors.

\begin{lemma}\label{lemma:spatial-regularity-truncation}
  For any $\varepsilon_{\mathrm{tr}} > 0$, there exists an integer $M \ge 1$ depending only on $\varepsilon_{\mathrm{tr}}$, $\lambda_0$, and $B_0$, such that for any initial condition $\ffunc_0$ satisfying Assumption~\ref{assume:initial-regularity}, the truncated Fourier series $\rho_t^{(M)}(x) \defn \sum_{k=-M}^{M} \widehat{\rho}_t(k) e^{2\pi i k x}$ of the density $\rho_t$ satisfies
  \begin{equation*}
    \sup_{t\in[0,\totaltime]} \big\|\rho_t - \rho_t^{(M)}\big\|_{\mathbb{L}^2(\torus)} \le \varepsilon_{\mathrm{tr}}.
  \end{equation*}
Moreover, one could set $M = \left \lceil \frac{1}{2c_0} \ln(C_1 / \varepsilon_{\mathrm{tr}}^2)\right \rceil$, where $C_1$ and $c_0$ are the constants in \Cref{lemma:free-stream-decay-general} that depend only on $(\lambda_0,B_0)$.
\end{lemma}
 See \Cref{subsubsec:proof-spatial-regularity-truncation} for the proof of this lemma. \\
Let $\mathcal{K} \defn \{-M, \dots, M\}$ be the set of retained frequencies, with total dimension $J \defn 2M + 1$. Define the discrete sampling matrix $\Phi \in \mathbb{C}^{N \times J}$ by
\begin{equation*}
  \Phi_{jk} \;\defn\; e^{2\pi i k x_j}, \qquad j = 1,\dots,N,\ \ k \in \mathcal{K}.
\end{equation*}

\begin{lemma}
  \label{lemma:exact-isometry}
  Suppose the number of sensors satisfies $N > 2M$, where the sensor locations are given by \Cref{assume:sensor-distribute}. Then, the normalized Gram matrix $\frac{1}{N}\Phi^\ast \Phi$ is exactly the identity matrix $I_J$. Consequently, for all $u \in \mathbb{C}^J$, we have the exact isometry
  \begin{equation*}
    \frac{1}{N}\,\|\Phi u\|_2^2 \;=\; \|u\|_2^2.
  \end{equation*}
  Furthermore, the matrix $\Phi$ has full column rank, and its least-squares reconstruction operator  $\Phi^\dagger = (\Phi^\ast \Phi)^{-1}\Phi^\ast$ has an operator norm strictly equal to $1/\sqrt{N}$.
\end{lemma}
See \Cref{subsubsec:proof-exact-isometry} for the proof of this lemma. \\
We then construct the desired estimator. Recall that we place $N$ sensors at uniform locations $x_1,\dots,x_N \in \torus$, and at discrete times
\[
  t_k = t - k \Delta t, \qquad k = 0,1,\dots,K-1,
\]
we observe noisy point samples of the density
\begin{equation}
  Y_{i,k}
  = \rho_{t_k}(x_i) + \varepsilon_{i,k},
  \qquad
  \varepsilon_{i,k} \simiid \mathcal{N}(0,\sigma^2),
  \quad
  i = 1,\dots,N,\; k = 0,\dots,K-1.
  \label{eq:window-observation}
\end{equation}
Under the expert policy, the system follows the free-stream dynamics for $t \ge t_0$. Our analysis uses the fact that the macroscopic density is sufficiently smooth in time on the history window $[t-K\Delta t,t]$, as quantified by the following lemma.

\begin{lemma}
  \label{lem:rho-time-derivative}
  Suppose the initial condition $f_0$ satisfies Assumption~\ref{assume:initial-regularity} with a velocity decay parameter $q \ge 3$. Then for any integer $1 \le \ell \le q - 2$, there exists a constant $C_\ell > 0$, depending only on $\ell$, $q$, and the initial regularity constants $(\lambda_0, B_0)$, such that for all $s \in [t-K\Delta t, t]$ and all $x \in \torus$, the macroscopic density satisfies
  \[
    \big|\partial_s^\ell \rho_s(x)\big| \;\le\; C_\ell.
  \]
\end{lemma}
See \Cref{subsec:proof-rho-time-derivative} for the proof of this lemma.\\
Now fix a window $[t-K\Delta t,t]$ such that Lemma~\ref{lem:rho-time-derivative} holds. Denote $p \defn q - 2 $, we will concretely construct an $\mathcal{O}(\Delta t^p)$ estimator. \\
To ensure that we can construct a large number of independent estimators without reusing any noisy observation, we define a strictly separated range for the base index $k$. Let $\mathcal{K}_p$ be the set of indices:
\begin{equation*}
  \mathcal{K}_p \defn \Big\{ \lfloor \frac{p-1}{p^2} K \rfloor + 1, \dots, \lfloor \frac{1}{p} K \rfloor \Big\}.
\end{equation*}
For any $k \in \mathcal{K}_p$, the time index multiples $\{k, 2k, \dots, pk\}$ reside in strictly disjoint intervals, guaranteeing that for any two distinct $k_1, k_2 \in \mathcal{K}_p$, their required historical indices are totally disjoint. \\For each sensor $i$ and each $k \in \mathcal{K}_p$, we form the $p$-th order extrapolation estimator using binomial weights over $p$ historical points:
\begin{equation}
  Z_{i,k}
  \;\defn\; \sum_{j=1}^p (-1)^{j-1} \binom{p}{j} Y_{i, j k}.
  \label{eq:local-extrap-estimator}
\end{equation}
\begin{lemma}\label{lem:local-extrap}
  Under the conditions of Lemma~\ref{lem:rho-time-derivative}, for each sensor $i$ and each $k \in \mathcal{K}_p$, the estimator defined in \eqref{eq:local-extrap-estimator} satisfies
  \begin{align}
    \big|\E[Z_{i,k}] - \rho_t(x_i)\big| &\le C_p\, (k\Delta t)^p, \label{eq:local-bias}\\
    \Var(Z_{i,k}) &= V_p \sigma^2, \label{eq:local-var}
  \end{align}
  where $C_p > 0$ depends only on $p$ and the initial regularity, and the variance multiplier is exactly $V_p = \binom{2p}{p} - 1$.
\end{lemma}
See \Cref{subsec:proof-local-extrap} for the proof.
To obtain a single, low-variance estimate of $\rho_t(x_i)$ at sensor $i$, we average the individual estimators over the entirely disjoint family of historical index sets defined by $\mathcal{K}_p$. Let $J_p \defn |\mathcal{K}_p|$ be the total number of independent estimators available in the temporal window. We define the \emph{$p$-th order window estimator} at sensor $i$ by
\begin{equation*}
  \widetilde{Y}_i \defn \frac{1}{J_p} \sum_{k \in \mathcal{K}_p} Z_{i,k}.
\end{equation*}

\begin{lemma}\label{lem:sensor-bias-variance}
  Suppose the window size is sufficiently large such that $K \ge 2p^2$, ensuring $J_p \ge K / (2p^2) > 0$. For each sensor $i=1,\dots,N$, the averaged estimator $\widetilde{Y}_i$ satisfies
  \begin{align}
    \big|\E[\widetilde{Y}_i] - \rho_t(x_i)\big| &\le C_B (\Delta t K)^p, \label{eq:sensor-bias}\\
    \Var(\widetilde{Y}_i) &\le C_V \frac{\sigma^2}{K}, \label{eq:sensor-var}
  \end{align}
  where the deterministic constants are explicitly given by $C_B = C_p / p^p$ (with $C_p$ defined in Lemma~\ref{lem:local-extrap}) and $C_V = 2p^2 \big[\binom{2p}{p} - 1\big]$.
\end{lemma}See \Cref{subsec:proof-sensor-bias-variance} for the proof.
Collecting the estimates into a vector $\widetilde{Y} = (\widetilde{Y}_1,\dots,\widetilde{Y}_N)^\top$, and letting $\Phi \in \mathbb{C}^{N \times (2M+1)}$ be the exact isometry sampling matrix from Lemma~\ref{lemma:exact-isometry}, we reconstruct the Fourier coefficients via least-squares:
\begin{equation*}
  \widetilde{c} \defn (\Phi^\ast \Phi)^{-1} \Phi^\ast \widetilde{Y}.
\end{equation*}
We define the output of our base policy $\policy_0 (\Dset[t - \tburn, t])$ as the electric field derived from these coefficients:
\begin{equation}
  \widetilde{E}_t(x) \defn \sum_{0 < |k| \le M} \frac{\widetilde{c}_k}{-2\pi i k} e^{2\pi i k x}.
\end{equation}
We now establish the pointwise bias and Gaussian variance bounds for $\widetilde{E}_t(x_i)$. We begin by expressing the window estimator vector as
\begin{equation*}
  \widetilde{Y} = \Phi c_t + b + \zeta,
\end{equation*}
where $c_t$ is the vector of true Fourier coefficients $\{c_t(k) \mydefn \widehat\rho_t(k)\}_{0 < |k| \le M}$ of the truncated density $\rho_t^{(M)}$, $b$ is the deterministic bias satisfying $\|b\|_\infty \le C_B \tburn^p$, and $\zeta$ is the independent zero-mean Gaussian noise vector with $\E\|\zeta\|_2^2 \le N C_V \sigma^2 / K$.\\
By Lemma~\ref{lemma:exact-isometry}, the reconstruction matrix is $\Phi^\dagger = \frac{1}{N}\Phi^\ast$ with its operator norm being exactly $\|\Phi^\dagger\|_{\mathrm{op}} = 1/\sqrt{N}$. For the coefficient bias $\E[\widetilde{c}] - c_t = \Phi^\dagger b$, we bound its $\ell_2$ norm as:
\begin{equation*}
  \|\E[\widetilde{c}] - c_t\|_2 \le \|\Phi^\dagger\|_{\mathrm{op}} \|b\|_2 \le \frac{1}{\sqrt{N}} \sqrt{N} \|b\|_\infty = \|b\|_\infty \le C_B \tburn^p.
\end{equation*}
To translate this coefficient bias bound into the electric field bias bound, we use the Poisson equation $\widehat{E}_t(k) = c_t(k) / (-2\pi i k)$ and the Cauchy-Schwarz inequality. The total pointwise expectation bias at sensor $x_i$ decomposes into the estimation error of the low-frequency modes and the spatial truncation error of the high-frequency modes:
\begin{align*}
  \abss{ \Exs \big[ \widetilde{E}_t (x_i) \big] - \ctrlE_t^{\mathrm{expert}} (x_i)} 
  &\le \abss{ \sum_{0 < |k| \le M} \frac{\Exs[\widetilde{c}_k] - c_t(k)}{-2\pi i k} e^{2\pi i k x_i} } + \abss{ \sum_{|k| > M} \frac{c_t(k)}{-2\pi i k} e^{2\pi i k x_i} } \\
  &\le \Big( \sum_{0 < |k| \le M} \big| \Exs[\widetilde{c}_k] - c_t(k) \big|^2 \Big)^{\frac{1}{2}} \Big( \sum_{k \neq 0} \frac{1}{4\pi^2 k^2} \Big)^{\frac{1}{2}} \\
  &\quad + \Big( \sum_{|k| > M} |c_t(k)|^2 \Big)^{\frac{1}{2}} \Big( \sum_{k \neq 0} \frac{1}{4\pi^2 k^2} \Big)^{\frac{1}{2}}.
\end{align*}
Recognizing that the constant series evaluates to $\sum_{k \neq 0} \frac{1}{4\pi^2 k^2} = \frac{1}{12} < 1$, the first term is strictly bounded by the $\ell_2$ coefficient error $\|\E[\widetilde{c}] - c_t\|_2$. The second term is bounded by the spatial truncation error $\|\rho_t - \rho_t^{(M)}\|_{L^2(\torus)} \le \varepsilon_{\mathrm{tr}}$ established in Lemma~\ref{lemma:spatial-regularity-truncation}. Consequently, we obtain the uniform pointwise bias bound:
\begin{equation*}
  \abss{ \Exs \big[ \widetilde{E}_t (x_i) \big] - \ctrlE_t^{\mathrm{expert}} (x_i)} 
  \le \|\E[\widetilde{c}] - c_t\|_2 + \varepsilon_{\mathrm{tr}} 
  \le C_B \tburn^p + \varepsilon_{\mathrm{tr}}.
\end{equation*}
For the variance, the coefficient error is purely driven by noise: $\widetilde{c} - \E[\widetilde{c}] = \Phi^\dagger \zeta$. Its total variance satisfies
\begin{equation*}
  \E\|\widetilde{c} - \E[\widetilde{c}]\|_2^2 = \E\|\Phi^\dagger \zeta\|_2^2 \le \|\Phi^\dagger\|_{\mathrm{op}}^2 \E\|\zeta\|_2^2 \le \frac{1}{N} \Big(N \cdot C_V \frac{\sigma^2}{K}\Big) = C_V \frac{\sigma^2}{K}.
\end{equation*}
Since the final output $\widetilde{E}_t(x_i)$ is a linear combination of the independent Gaussian noise vector $\zeta$, its pointwise fluctuation is exactly a zero-mean Gaussian random variable. Because the Poisson multiplier satisfies $1/(4\pi^2 k^2) < 1$, the pointwise variance $v_{t,i}^2$ is bounded by the total variance of the coefficients:
\begin{equation*}
  v_{t,i}^2 \defn \E \Big[ \big| \widetilde{E}_t(x_i) - \Exs[\widetilde{E}_t(x_i)] \big|^2 \Big] \le \E\|\widetilde{c} - \E[\widetilde{c}]\|_2^2 \le C_V \frac{\sigma^2}{K}.
\end{equation*}
Recalling that the number of sampled windows is $K = \tburn / \stepsize$, we conclude that
\begin{equation*}
  v_{t,i}^2 \le C_V \sigma^2 \frac{\stepsize}{\tburn},
\end{equation*}which completes the proof of the bias and variance bounds for the policy $\pi_0$.\\
Finally, we establish the Lipschitz continuity of the policy $\pi_0$ with respect to its input observation matrix. Let $Y^{(1)}, Y^{(2)} \in \mathbb{R}^{N \times (K+1)}$ be any two input historical density observations, and denote their difference by $\Delta Y \defn Y^{(1)} - Y^{(2)}$.\\
Since the policy $\pi_0$ is a composition of linear operators, we trace the input difference $\Delta Y$ through the reconstruction steps. For the $p$-th order extrapolation estimator defined in \eqref{eq:local-extrap-estimator}, the difference at sensor $i$ for index $k$ satisfies:
\begin{align*}
  |\Delta Z_{i,k}| &= \Big| \sum_{j=1}^p (-1)^{j-1} \binom{p}{j} \Delta Y_{i, jk} \Big| \\
  &\le \sum_{j=1}^p \binom{p}{j} |\Delta Y_{i, jk}| \\
  &\le \Big( \sum_{j=1}^p \binom{p}{j} \Big) \|\Delta Y\|_{\infty} = (2^p - 1) \|\Delta Y\|_{\infty}.
\end{align*}
Because the window estimator $\widetilde{Y}_i$ is a simple arithmetic mean of $Z_{i,k}$ over the set $\mathcal{K}_p$, this uniform bound is perfectly preserved:
\begin{equation*}
  |\Delta \widetilde{Y}_i| \le (2^p - 1) \|\Delta Y\|_{\infty}, \qquad \forall i = 1,\dots,N.
\end{equation*}
Next, we bound the difference in the reconstructed Fourier coefficients $\Delta \widetilde{c} = \Phi^\dagger \Delta \widetilde{Y}$. Recalling from Lemma~\ref{lemma:exact-isometry} that $\Phi^\dagger = \frac{1}{N}\Phi^\ast$, the coefficient difference for any mode $k$ is bounded by:
\begin{align*}
  |\Delta \widetilde{c}_k| &= \Big| \frac{1}{N} \sum_{j=1}^N e^{-2\pi i k x_j} \Delta \widetilde{Y}_j \Big| \\
  &\le \frac{1}{N} \sum_{j=1}^N \big| e^{-2\pi i k x_j} \big| \cdot |\Delta \widetilde{Y}_j| \\
  &\le \max_{1 \le j \le N} |\Delta \widetilde{Y}_j| \le (2^p - 1) \|\Delta Y\|_{\infty}.
\end{align*}
Finally, we translate this coefficient difference into the spatial $\mathbb{L}^\infty$ norm of the reconstructed electric field. Using the triangle inequality and the bounds on $|\Delta \widetilde{c}_k|$, we have for any $x \in \torus$:
\begin{align*}
  \big| \pi_0(Y^{(1)})(x) - \pi_0(Y^{(2)})(x) \big| &= \Big| \sum_{0 < |k| \le M} \frac{\Delta \widetilde{c}_k}{-2\pi i k} e^{2\pi i k x} \Big| \\
  &\le \sum_{0 < |k| \le M} \frac{|\Delta \widetilde{c}_k|}{2\pi |k|} \\
  &\le \frac{(2^p - 1)}{\pi} \|\Delta Y\|_{\infty} \sum_{k=1}^M \frac{1}{k}.
\end{align*}
Applying the standard harmonic series bound $\sum_{k=1}^M \frac{1}{k} \le 1 + \ln M$, we obtain the uniform spatial bound:
\begin{equation*}
  \big\| \pi_0(Y^{(1)}) - \pi_0(Y^{(2)}) \big\|_{\mathbb{L}^\infty(\torus)} \le \frac{2^p - 1}{\pi} (1 + \ln M) \|Y^{(1)} - Y^{(2)}\|_{\infty}.
\end{equation*}
Since $M$ is uniquely determined by $\varepsilon_{\mathrm{tr}}, \lambda_0,$ and $B_0$ , the coefficient $C_L \defn \frac{2^p - 1}{\pi} (1 + \ln M)$ is a deterministic finite constant. Therefore, the policy $\pi_0$ is globally Lipschitz continuous from the discrete observation space to the output functional space $\mathbb{L}^\infty(\torus)$. Because the domain $\torus$ is bounded, this inherently implies the $L^2(\torus)$ Lipschitz continuity as well, which completes the proof.

\subsubsection{Proof of \Cref{lemma:regularity-of-policy-class}}\label{subsubsec:proof-regularity-of-policy-class}
By definition, the measure $\nu$ that defines any exponential weight policy in $\Pi_\tau$ is supported on the space of bounded Lipschitz functions $\mathcal{F}_{BL} (L)$. Therefore, the image of any $\pi \in \Pi_\tau$ is a mixture of bounded Lipschitz functions, which is itself a bounded Lipschitz function, i.e., $\pi(\Dset) \in \mathcal{F}_{BL} (L)$ for all $\pi \in \Pi_\tau$. It remains to verify the Lipschitzness of the policy itself with respect to the input dataset $\Dset$.

Let $\pi \in \Pi_\tau$ be a policy defined by the measure $\nu$, and let $\Dset_1, \Dset_2$ be any two datasets. By \Cref{lemma:building-block-in-policy-class-construction}, we have
\begin{align*}
  \vecnorm{\pi_0 (\Dset_1) - \pi_0 (\Dset_2)}{\infty} \leq \frac{2^p - 1}{\pi} (1 + \ln M) \|\Dset_1 - \Dset_2\|_{\infty}.
\end{align*}
Since the image of $\pi_0$ lies in the finite-dimensional subspace $\LinSpace_N$, we let $\pi_0 (\Dset_i) = \sum_{j=1}^N a_{j}^{(i)} \phi_j$ for $i=1,2$, where $\{\phi_j\}$ is the Fourier basis. By Parseval's identity, we have
\begin{align*}
  \vecnorm{a^{(1)} - a^{(2)}}{N} =  \vecnorm{\pi_0 (\Dset_1) - \pi_0 (\Dset_2)}{\ltwospace} \leq \vecnorm{\pi_0 (\Dset_1) - \pi_0 (\Dset_2)}{\infty}.
\end{align*}
On the other hand, we have
\begin{align*}
  \nabla_a \int f \, d w_{\tau, \nu} (f; a^\top \phi)
  = \frac{2N}{\tau} \int f \cdot \phi w_{\tau, \nu} (f; a^\top \phi) - \frac{2N}{\tau} \int f \, d w_{\tau, \nu} (f; a^\top \phi) \cdot \int \phi \, d w_{\tau, \nu} (f; a^\top \phi).
\end{align*}
Since $\mathrm{supp}(\nu) \subseteq \mathcal{F}_{BL}(L)\cap\LinSpace_N$, we have
\begin{align*}
  \vecnorm{ \nabla_a \int f \, d w_{\tau, \nu} (f; a^\top \phi)}{2} \leq \frac{4N}{\tau} \cdot L \sqrt{N},
\end{align*}
and consequently, we have
\begin{align*}
  \vecnorm{\pi (\Dset_1) - \pi (\Dset_2)}{\infty} &= \vecnorm{\int f \, d w_{\tau, \nu} (f; a^{(1)\top} \phi) - \int f \, d w_{\tau, \nu} (f; a^{(2)\top} \phi)}{\infty} \\
  &\leq \sqrt{N} \cdot \vecnorm{\int f \, d w_{\tau, \nu} (f; a^{(1)\top} \phi) - \int f \, d w_{\tau, \nu} (f; a^{(2)\top} \phi)}{\ltwospace} \\
  &\leq \sqrt{N} \cdot \frac{4N}{\tau} \cdot L \sqrt{N} \cdot \vecnorm{a^{(1)} - a^{(2)}}{2} \\
  &\leq \frac{4N^2 L}{\tau} \cdot \frac{2^p - 1}{\pi} (1 + \ln M) \|\Dset_1 - \Dset_2\|_{\infty},
\end{align*}
which proves the Lipschitz continuity of $\pi$ with respect to the input dataset $\Dset$.

\subsubsection{Proof of \Cref{lemma:resolution-entropy-bound}}\label{subsubsec:proof-resolution-entropy-bound}
We start with the following lemma that characterizes the key properties of the resolution entropy functional $H_\varepsilon$.
\begin{lemma}\label{lemma:entropy-properties}
The resolution entropy functional $H_\varepsilon$ satisfies the following properties:
\begin{itemize}
  \item Data processing inequality: for any measure $\Prob$ and an $L$-Lipschitz mapping $\phi$, we have
  \begin{align*}
    H_\varepsilon (\phi_\sharp \Prob) \leq H_{\varepsilon / L} (\Prob)
  \end{align*}
  \item Chain rule: for a collection of measures $(\Prob_a)_{a \in \mathcal{A}}$ and a distribution $\mu$ over $\mathcal{A}$ the mixture measure satisfies
  \begin{align*}
    H_\varepsilon \Big( \int_{\mathcal{A}} \Prob_a d\mu(a) \Big) \leq \int_{\mathcal{A}} H_\varepsilon (\Prob_a) d\mu(a) + I \big( A; X \big),
  \end{align*}
  where $A \sim \mu$ is the random index and $X \sim \Prob_A$ is the random variable drawn from the corresponding measure.
\end{itemize}
\end{lemma}
\noindent See \Cref{subsec:app-proof-lemma-entropy-properties} for the proof of \Cref{lemma:entropy-properties}.

Taking this lemma as given, we proceed to prove \Cref{lemma:resolution-entropy-bound}. First, we define the un-projected distributions as
\begin{align*}
  \widetilde{\nu}_t (\mathcal{A}) \mydefn \Prob (\ctrlE_t^{\mathrm{expert}} \in \mathcal{A}), \quad \mbox{for any }\mathcal{A} \subseteq \ltwospace, \quad \mbox{and} \qquad \widetilde{\nu}_+ \mydefn \frac{\stepsize}{\totaltime - \tburn} \sum_{m = \lceil \tburn / \stepsize \rceil}^{\lfloor \totaltime / \stepsize \rfloor} \widetilde{\nu}_{m \stepsize}.
\end{align*}
Since the projection operator $P_\Nsensors$ is $1$-Lipschitz, by the data processing inequality in \Cref{lemma:entropy-properties}, we have
\begin{align*}
 H_\varepsilon (\nu_+) = H_\varepsilon \big((P_\Nsensors)_\sharp \widetilde{\nu}_+ \big) \leq H_\varepsilon (\widetilde{\nu}_+).
\end{align*}
Next, we apply the chain rule in \Cref{lemma:entropy-properties} to the mixture distribution $\widetilde{\nu}_+$, which is a uniform mixture of the distributions $\{\widetilde{\nu}_{m \stepsize}\}$
\begin{align*}
  H_\varepsilon (\widetilde{\nu}_+) &\leq \frac{\stepsize}{\totaltime - \tburn} \sum_{m = \lceil \tburn / \stepsize \rceil}^{\lfloor \totaltime / \stepsize \rfloor} H_\varepsilon (\widetilde{\nu}_{m \stepsize}) + I \big( m; \density_{m\stepsize} \big) \\
  &\leq \frac{\stepsize}{\totaltime - \tburn} \sum_{m = \lceil \tburn / \stepsize \rceil}^{\lfloor \totaltime / \stepsize \rfloor} H_\varepsilon (\widetilde{\nu}_{m \stepsize}) + \log (\totaltime / \stepsize).
\end{align*}
In order to bound each term in the sum, we need the following lemma that controls the sensitivity of the expert control $\ctrlE_t^{\mathrm{expert}}$ with respect to the initial condition $f_0$.
\begin{lemma}\label{lemma:expert-policy-regularity}
  Let $\ffunc_0^{(1)}$ and $\ffunc_0^{(2)}$ be two initial conditions satisfying Assumption~\ref{assume:initial-regularity}, and let $\ctrlE_t^{\mathrm{expert}, (1)}$ and $\ctrlE_t^{\mathrm{expert}, (2)}$ be the corresponding expert controls defined by \eqref{eq:expert-policy}. Then we have the following uniform bound for all $t \in [\tburn, \totaltime]$:
  \begin{align*}
    \ltwonorm{\ctrlE_t^{\mathrm{expert}, (1)} - \ctrlE_t^{\mathrm{expert}, (2)}} \leq C \ltwonorm{f_0^{(1)} - f_0^{(2)}}.
  \end{align*}
\end{lemma}
\noindent See \Cref{subsec:app-proof-lemma-expert-policy-regularity} for the proof of \Cref{lemma:expert-policy-regularity}.

By \Cref{lemma:expert-policy-regularity}, the mapping $\ffunc_0 \mapsto \ctrlE_t^{\mathrm{expert}}$ is $C$-Lipschitz, and hence the distribution $\widetilde{\nu}_{m \stepsize}$ is the pushforward of the initial distribution $\Prob$ under a $C$-Lipschitz mapping. Applying the data processing inequality in \Cref{lemma:entropy-properties} again, we have
\begin{align*}
  H_\varepsilon (\widetilde{\nu}_{m \stepsize}) \leq H_{\varepsilon / C} (\Prob),
\end{align*}
for all $m$. Substituting this bound back into the previous inequality, we arrive at
\begin{align*}
  H_\varepsilon (\nu_+) \leq H_{\varepsilon / C} (\initDistr) + \log (\totaltime / \stepsize),
\end{align*}
which completes the proof of \Cref{lemma:resolution-entropy-bound}.

\section{Discussion}\label{sec:discussion}
In this paper, we have established a theoretical framework for understanding imitation learning for controlling the stability of the Vlasov-Poisson system. We show that small behavior cloning error yields stability control within a finite time horizon, and also provide a criterion for the possibility of achieving small BC error with partial observations. Through the theoretical analysis, we developed behavior cloning algorithms that provably adapts to the complexity of the initial distribution, a key factor that determines the learnability of the control problem. Numerical experiments validate our theoretical findings and demonstrate the practical effectiveness of our proposed algorithms.

Our work opens up several avenues for future research. First, it is important future direction to explore the capacity of other imitation learning algorithms, such as DAgger and its variants, especially for long-time control of the Vlasov-Poisson system. Second, while our analysis and experiments focus on the one-dimensional torus case, it is of great interest to extend our results to $3+3$-dimensional real-world tokamak geometry. Additionally, it would be valuable to investigate the interplay between imitation learning and reinforcement learning for controlling plasma stability, and to develop algorithms that go beyond the free-streaming expert dynamics considered in this paper. Finally, though this work focuses on the Vlasov--Poisson system, the theoretical framework and algorithmic insights may be applicable to a broader class of high-dimensional control problems governed by kinetic equations, and it would be valuable to explore such extensions in future work.

\bibliographystyle{alpha}
\bibliography{references}

\appendix
\section{Technical results used in the proof of \Cref{thm:main-theorem}}

\subsection{Proof of \Cref{lem:J-time-lipschitz}}\label{subsubsec:proof-J-time-lipschitz}

Fix a trajectory $z$ and a policy $\pi\in\Pi$. Throughout we work on the event $\mathcal E_{\mathrm{noise}}$.
Since the expert dynamics is free-streaming, the solution is explicit:
\begin{equation}\label{eq:fs-explicit}
f_t^z(x,v)=f^z_{0}(x-tv,v),\qquad t\in[0,T].
\end{equation}
By Assumption~\ref{assume:initial-regularity}, for every integer $m\ge0$,
\[
\max_{|\alpha|\le m}\ \sup_{x,v}(1+|v|)^q\,|\partial_{x,v}^\alpha f_{0}^z(x,v)|
\le \frac{m!}{\lambda_0^m}\,B_0 \text{ for all } z.
\]
Using \eqref{eq:fs-explicit} and the chain rule, any mixed derivative
$\partial_{x,v}^\alpha f^z(t,x,v)$ with $|\alpha|\le 1$ is a finite linear combination of terms
$t^r\,(\partial_{x,v}^\beta f_{0}^z)(x-tv,v)$ with $r\le 1$ and $|\beta|\le 1$.
Therefore there exists a deterministic constant $C_1 \;=\; C_1(B_0,\lambda_0,T)$ such that, for all $t\in[0,T]$ and all $z$,
\begin{align}\label{eq:K4-from-analytic}
\max_{|\alpha|\le 1}\ \sup_{t,x,v}(1+|v|)^q\,|\partial_{x,v}^\alpha f_t^z(x,v)|\le C_1(B_0,\lambda_0,T). 
\end{align}
From \eqref{eq:K4-from-analytic} and the integrability of $(1+|v|)^{-q}$, we get for all $z$
\begin{equation}\label{eq:rho-sup}
\sup_{t\in[0,T]}\|\rho_t^z(\cdot)\|_{\mathbb{L}^\infty(\T)} \le C\,C_1,
\qquad
\sup_{t\in[0,T]}\|\partial_x\rho_t^z(\cdot)\|_{\mathbb{L}^\infty(\T)} \le C\,C_1,
\end{equation}
For the time derivative, using the free-stream equation $\partial_t f^z=-v\partial_x f^z$,
\[
\partial_t\rho_t^z(x)=\int_\R \partial_t f_t^z(x,v)\,dv
= -\partial_x \int_\R v\, f_t^z(x,v)\,dv.
\]
Hence, applying \eqref{eq:K4-from-analytic} to $\partial_x f$ and by the integrability of $|v|(1+|v|)^{-q}$ (since $q \ge 2 > 1$), we get for all $z$,
\begin{equation}\label{eq:dt-rho}
\sup_{t\in[0,T]}\|\partial_t\rho_t^z(\cdot)\|_{\mathbb{L}^\infty(\T)} \le C\,C_1.
\end{equation}
Since $\partial_x E_t^z=1-\rho_t^z$ and $E_t^z$ has zero mean, interpolation inequality on $\T$ gives
\begin{equation}\label{eq:E-and-dtE-bds}
\|E_t^z\|_{\mathbb{L}^\infty(\T)}
\le C\|\rho_t^z-1\|_{\mathbb{L}^\infty(\T)} \le CC_1,
\qquad
\|\partial_t E_t^z\|_{\mathbb{L}^\infty(\T)}
\le C\|\partial_t\rho_t^z(\cdot)\|_{\mathbb{L}^\infty(\T)} \le CC_1.
\end{equation}
Define
\[
\delta(\pi, z,t)\coloneqq \pi(\mathcal D^z[t-t_0,t])+E^z(t),
\qquad \text{so that}\qquad
J(\pi,z,t)=\|\delta(\pi,z,t)\|_{\mathbb{L}^2(\T)}^2.
\]
We first bound the increment of $\delta$ by sepearately bounding the increments of $\pi(\mathcal D^z[t-t_0,t],t)$ and $E^z(t)$.Recall that $t_k = t_0 + k\eta$ and $I_k=[t_k,t_{k+1})$.
Fix $k$ and suppose $t,s\in I_k$. Then we have $\mathcal D^z[t-t_0,t]$ and $\mathcal D^z[s-t_0,s]$
are identical, so for all $t,s \in I_k$, we have
\[
\|\pi(\mathcal D^z[t-t_0,t])
-
\pi(\mathcal D^z[s-t_0,s])\|_{\mathbb{L}^\infty(\T)}
= 0.
\]
Moreover, by~\eqref{eq:E-and-dtE-bds}, we obtain
\[
\|E_t^z(\cdot)-E_s^z(\cdot)\|_{\mathbb{L}^\infty(\T)}
\le
C C_1 |t-s|,
\]
Combining the two bounds above gives
\begin{equation}\label{eq:delta-Lip}
\|\delta(\pi,z,t)-\delta(\pi,z,s)\|_{\mathbb{L}^\infty(\T)}
\le C C_1 |t-s|,
\end{equation}
On $\mathcal E_{\mathrm{noise}}$, by \Cref{eq:B-delta-explicit} we have there exists a deterministic constant $B$, depending only on
\[
(B_0,\lambda_0,T,L,C_{\mathrm{bias}},\sigma_\rho,n,N,\eta,\delta_0),
\]
such that
\begin{equation}\label{eq:delta-unif}
\sup_{t\in[t_0,T]}
\|\delta(\pi,z,t)\|_{\mathbb{L}^2(\T)}
\le
B,
\qquad
\forall\,\pi\in\Pi,\ \forall\,z.
\end{equation}
Finally, using
\[
|\|a\|_2^2-\|b\|_2^2|
\le
(\|a\|_2+\|b\|_2)\|a-b\|_2,
\qquad
\|f\|_2\le\|f\|_\infty \text{ on } \T,
\]
together with~\eqref{eq:delta-Lip}--\eqref{eq:delta-unif}, we obtain
\[
|J(\pi,z,t)-J(\pi,z,s)|
\le
2B CC_1 |t-s|,
\qquad
\forall\,t,s\in I_k.
\]
This proves the piecewise Lipschitz continuity of $J(\pi,z,t)$ in $t$ uniformly in $\pi$ and $z$, with Lipschitz constant $2BCC_1$.

\subsection{Proof of \Cref{lemma:transport-perturbed-rho-bound}}\label{subsubsec:proof-transport-perturbed-rho-bound}
For any $(t,x,v)$, let $(X(s),V(s))_{0\le s\le t}$ be the backward characteristic:
\[
\dot X(s)=V(s),\qquad \dot V(s)=\delta_s(X(s)),\qquad X(t)=x,\ V(t)=v.
\]
Define the free-streaming path $X_{\rm f}(s):=x-(t-s)v$ and the offsets
\[
\Delta v:=V(0)-v,\qquad \Delta x:=X(0)-X_{\rm f}(0)=X(0)-(x-tv).
\]
By Cauchy--Schwarz,
\[
\int_0^t \|\delta_s\|_{\mathbb{L}^\infty(\T)}\,ds
\le \Big(\int_0^t \|\delta_s\|_{\mathbb{L}^\infty(\T)}^2\,ds\Big)^{1/2}\, t^{1/2}
= D(t)\,t^{1/2}.
\]
Integrating $\dot V=\delta_s(X(s))$ gives
\[
\Delta v=\int_0^t \delta_{s}(X(s))\,ds,
\qquad\Rightarrow\qquad
|\Delta v|\le \int_0^t \vecnorm{\delta_{s}}{\mathbb{L}^\infty(\T)}\,ds \le D(t)\,t^{1/2}.
\]
Similarly, integrating $\dot X=V$ gives
\[
\Delta x=X(0)-(x-tv)=\int_0^t (V(\tau)-v)\,d\tau
=\int_0^t\int_\tau^t \delta_s(X(s))\,ds\,d\tau
=\int_0^t (t - s) \,\delta_s(X(s))\,ds,
\]
hence by Cauchy--Schwarz, we have
\[
|\Delta x|
\le \int_0^t (t-s)\|\delta_s\|_{\mathbb{L}^\infty(\T)}\,ds
\le \Big(\int_0^t (t-s)^2 ds\Big)^{1/2}\Big(\int_0^t \|\delta_s\|_{\mathbb{L}^\infty(\T)}^2 ds\Big)^{1/2}
= \frac{t^{3/2}}{\sqrt3}\,D(t).
\]
In particular,
\begin{equation}\label{eq:dx-dv-bounds}
|\Delta v|\le D(t)t^{1/2},\qquad |\Delta x|\le D(t)t^{3/2}.
\end{equation}
Along characteristics the solution is constant, so
\begin{equation}\label{eq:repr-f}
f_t(x,v)=f_0\big(X(0),V(0)\big)=f_0\big(x-tv+\Delta x,\ v+\Delta v\big).
\end{equation}
Assumption~\ref{assume:initial-regularity} states that
\[
\|f_0\|_{\mathcal E_q(\lambda_0)}=\sum_{m=0}^{\infty}\frac{\lambda_0^m}{m!}\|f_0\|_{\mathcal C^m_q}\le B_0,
\qquad
\|f_0\|_{\mathcal C^m_q}
:=\sup_{0\le a+b\le m}\ \sup_{(x,v)\in\mathbb T\times\mathbb R}(1+|v|)^q\,|\partial_x^a\partial_v^b f_0(x,v)|.
\]
Therefore, for each $m\ge 0$,
\begin{equation}\label{eq:Cm-bound}
\|f_0\|_{\mathcal C^m_q}\le B_0\,\frac{m!}{\lambda_0^m},
\end{equation}
because all terms in the series are nonnegative.
Consequently, for any integers $a,b\ge0$,
\begin{equation}\label{eq:L1-derivative}
\sup_{x\in\mathbb T}\int_{\mathbb R}\big|\partial_x^a\partial_v^b f_0(x,v)\big|\,dv
\le
\|f_0\|_{\mathcal C^{a+b}_q}\int_{\mathbb R}(1+|v|)^{-q}\,dv
\le
C_\ast\,B_0\,\frac{(a+b)!}{\lambda_0^{a+b}},
\end{equation}
where $C_\ast:=\int_{\mathbb R}(1+|v|)^{-q}\,dv<\infty$ is an absolute constant (since $q \ge 2 > 1$).\\
Define the free-streaming density
\[
\rho^{\rm fs}_t(x):=\int_{\mathbb R} f_0(x-tv,v)\,dv.
\]
Then, by adding and subtracting $\rho^{\rm fs}$, we obtain
\begin{equation}\label{eq:split}
|\rho_t(x)-1|
\le
|\rho^{\rm fs}_t(x)-1|
+
|\rho_t(x)-\rho^{\rm fs}_t(x)|.
\end{equation}
We start by estimating the first term in \eqref{eq:split}.
Define
\[
\rho_0(x):=\int_{\real} f_0(x,v)\,dv,
\qquad
\widehat{\rho_0}(k):=\int_{\torus}\rho_0(x)\,e^{-2\pi i k x}\,dx,
\qquad k\in\mathbb Z,
\]
so that $\widehat{\rho_0}(0)=\int_{\torus\times\real} f_0\,dv\,dx=1$.
For $k\neq 0$, define the mixed Fourier transform of $f_0$ by
\[
\widehat f_0(k,\eta)
:=\int_{\torus\times\real} f_0(x,v)\,e^{-2\pi i(kx+\eta v)}\,dx\,dv,
\qquad \eta\in\real,
\]
so that $\widehat{\rho_0}(k)=\widehat f_0(k,0)$.
Under free streaming,
\[
\rho^{\mathrm{fs}}_t(x)=\int_{\real} f_0(x-tv,v)\,dv,
\]
and for each $k\neq 0$,
\[
\widehat{\rho^{\mathrm{fs}}_t}(k)
=\int_{\torus\times\real} f_0(y,v)\,e^{-2\pi i(k y + k t v)}\,dy\,dv
=\widehat f_0(k,kt).
\]
Hence, using the Fourier series on $\torus$ and $\widehat{\rho^{\mathrm{fs}}_t}(0)=1$,
\begin{equation}\label{eq:rho-fs-fourier-sum}
\rho^{\mathrm{fs}}_t(x)-1
=\sum_{k\in\mathbb Z\setminus\{0\}}\widehat f_0(k,kt)\,e^{2\pi i k x},
\qquad
\|\rho^{\mathrm{fs}}_t-1\|_{L^\infty(\torus)}
\le \sum_{k\neq 0}\big|\widehat f_0(k,kt)\big|.
\end{equation}
We next bound $\widehat f_0(k,\eta)$ using Assumption~\ref{assume:initial-regularity}.
For any $m\in\mathbb N$, integrating by parts $m$ times in $v$ yields
\[
(2\pi i\eta)^m\widehat f_0(k,\eta)
=\int_{\torus\times\real}\partial_v^m f_0(x,v)\,e^{-2\pi i(kx+\eta v)}\,dx\,dv,
\]
hence
\[
|\eta|^m\,|\widehat f_0(k,\eta)|
\le (2\pi)^{-m}\|\partial_v^m f_0\|_{\mathbb{L}^1(\torus\times\real)}.
\]
Using the velocity weight in $\|\cdot\|_{\mathcal C^m_q}$ and
$\int_{\real}(1+|v|)^{-q}dv<\infty$ (since $q \ge 2 > 1$), we have
\[
\|\partial_v^m f_0\|_{\mathbb{L}^1(\torus\times\real)}
\le C_\ast\,\|f_0\|_{\mathcal C^m_q},
\]
where $C_\ast>0$ is universal.
By \eqref{eq:Cm-bound}, $\|f_0\|_{\mathcal C^m}\le B_0\,m!\,\lambda_0^{-m}$, hence
\[
|\eta|^m\,|\widehat f_0(k,\eta)|
\le C_\ast B_0\,m!\,(2\pi\lambda_0)^{-m},
\qquad \forall m\in\mathbb N.
\]
Fix $\delta = \pi\lambda_0$. Multiplying by $\delta^m/m!$ and summing over $m\ge0$ yields
\[
e^{\delta|\eta|}\,|\widehat f_0(k,\eta)|
\le C_\ast B_0\sum_{m=0}^\infty\Big(\frac{\delta}{2\pi\lambda_0}\Big)^m
=\frac{C_\ast B_0}{1-\delta/(2\pi\lambda_0)} = 2C_\ast B_0.
\]
Consequently, there exist constants $C_0,c_0>0$ depending only on $(\lambda_0,B_0)$ such that
\[
|\widehat f_0(k,\eta)|\le C_0\,e^{-c_0|\eta|},
\qquad \forall k\in\mathbb Z,\ \eta\in\mathbb R.
\]
Applying this bound with $\eta=kt$ in \eqref{eq:rho-fs-fourier-sum} gives
\[
\|\rho^{\mathrm{fs}}_t-1\|_{\mathbb{L}^\infty(\torus)}
\le C_0\sum_{k\neq 0}e^{-c_0|k|t}
=2C_0\sum_{k=1}^\infty e^{-c_0 k t}
=\frac{2C_0 e^{-c_0 t}}{1-e^{-c_0 t}}.
\]
For $t\ge 1$,
\[
\frac{2C_0 e^{-c_0 t}}{1-e^{-c_0 t}}
\le C_1 e^{-c_0 t},
\]
with $C_1:=\frac{2C_0 }{1 - e^{-c_0}}$.
For $t\in[0,1]$, we simply use
\[
\|\rho_t^{\mathrm{fs}}\|_{L^\infty(\T)}
\le \sup_{x\in\T}\int_{\R}|f_0(x-tv,v)|\,dv
= \sup_{x\in\T}\int_{\R}|f_0(x,v)|\,dv
\le C(\lambda_0,B_0),
\]
and hence $\|\rho_t^{\mathrm{fs}}-1\|_{\mathbb{L}^\infty(\T)}\le C(\lambda_0,B_0)$ on $[0,1]$. Combining the two cases yields
\begin{align}\label{eq:rho-fs-bound}
   \|\rho_t^{\mathrm{fs}}-1\|_{\mathbb{L}^\infty(\torus)}\le C_1 e^{- c_2 t}
\end{align}
 for all $t\ge0$ for some constants $C_1,c_2>0$ depending only on $(\lambda_0,B_0)$.
It remains to bound the second term in \eqref{eq:split} uniformly in $x$:
\[
\rho_t(x)-\rho_t^{\rm fs}(x)
=
\int_{\mathbb R}
\Big(
f_0(x-tv+\Delta x,\,v+\Delta v)-f_0(x-tv,\,v)
\Big)\,dv.
\]
Fix $(t,x)$ and for each $v\in\mathbb R$ denote the base point $(x_0,v_0):=(x-tv,v)$.
Applying the multivariate Taylor formula for $f_0$ at $(x_0,v_0)$ to first order yields
\begin{align}
f_0(x_0+\Delta x,\,v_0+\Delta v)
&= f_0(x_0,v_0)
+ \partial_x f_0(x_0,v_0)\,\Delta x
+ \partial_v f_0(x_0,v_0)\,\Delta v
+ R_1(t,x,v),
\label{eq:taylor-1st}
\end{align}
where the remainder admits the integral representation
\begin{equation}\label{eq:R1-int}
R_1(t,x,v)
=
\sum_{a+b=2}\frac{1}{a!\,b!}\,(\Delta x)^a(\Delta v)^b
\int_0^1 (1-s)\,
\partial_x^a\partial_v^b f_0(x_0+s\Delta x,\ v_0+s\Delta v)\,ds .
\end{equation}
Subtracting $f_0(x_0,v_0)$, integrating in $v$, and taking absolute values yield
\begin{align}\label{eq:rho-pert-1st}
|\rho_t(x)-\rho_t^{\rm fs}(x)|
&\le
\int_{\mathbb R} |\Delta x|\,|\partial_x f_0(x-tv,v)|\,dv
\;+\;
\int_{\mathbb R}|\Delta v|\,|\partial_v f_0(x-tv,v)|\,dv
\;+\;
\int_{\mathbb R}|R_1(t,x,v)|\,dv .
\end{align}
As for the linear term, applying the elementary inequality $\int |h(v)|\, |g(v)|\,dv\le \|h\|_{\mathbb{L}^\infty_v}\int |g(v)|\,dv$ and \eqref{eq:dx-dv-bounds}, we obtain
\begin{align}\label{eq:first-order-bound}
\int_{\mathbb R}|\Delta x(t,x,v)|\,|\partial_x f_0(x-tv,v)|\,dv
&\le \|\Delta x(t,x,\cdot)\|_{\mathbb{L}^\infty_v}\int_{\mathbb R}|\partial_x f_0(x-tv,v)|\,dv
\\
&\le D(t)t^{3/2} \int_{\mathbb R}|\partial_x f_0(x-tv,v)|\,dv,
\nonumber\\
\int_{\mathbb R}|\Delta v(t,x,v)|\,|\partial_v f_0(x-tv,v)|\,dv
&\le \|\Delta v(t,x,\cdot)\|_{\mathbb{L}^\infty_v}\int_{\mathbb R}|\partial_v f_0(x-tv,v)|\,dv
\\
&\le D(t)t^{1/2} \int_{\mathbb R}|\partial_v f_0(x-tv,v)|\,dv.
\nonumber
\end{align}
Using the $\mathbb{L}^1_v$ derivative bound \eqref{eq:L1-derivative} with $(a,b)=(1,0)$ and $(0,1)$ gives
\begin{equation}\label{eq:first-order-final}
\int_{\mathbb R}
|\Delta x|\,|\partial_x f_0(x-tv,v)|\,dv
+
\int_{\mathbb R}
|\Delta v|\,|\partial_v f_0(x-tv,v)|\,dv
\le
C(\lambda_0,B_0)\Big[D(t)t^{1/2} + D(t)t^{3/2}\Big].
\end{equation}
As for the remainder $R_1$, we bound it pointwise using \eqref{eq:Cm-bound} with $m = 2$.
Recall
\[
R_1(t,x,v)=\sum_{a+b=2}\frac{1}{a!\,b!}(\Delta x)^a(\Delta v)^b
\int_0^1(1-s)\,
\partial_x^a\partial_v^b f_0(x_0+s\Delta x,\ v_0+s\Delta v)\,ds,
\qquad (x_0,v_0)=(x-tv,v).
\]
Taking absolute values, integrating in $v$, and using $(1-s)\le 1$ yield
\begin{align}
\int_{\mathbb R}|R_1(t,x,v)|\,dv
&\le
\sum_{a+b=2}\frac{1}{a!\,b!}
\int_{\mathbb R}|\Delta x(t,x,v)|^a|\Delta v(t,x,v)|^b
\int_0^1
\big|
\partial_x^a\partial_v^b f_0(x_0+s\Delta x,\ v_0+s\Delta v) 
\big|\,ds\,dv . \nonumber\\
&\le C(\lambda_0,B_0)\int_{\mathbb R}(|\Delta x|^2 + |\Delta x| |\Delta v| + |\Delta v|^2)\int_0^1 (1+|v + s \Delta v|)^{-3} ds\,dv. \label{eq:R1-start}
\end{align}
For each fixed $v$ we have
\[
1+|v+s\Delta v|\ge \frac{1+|v|}{1+|s\Delta v|}\qquad\Rightarrow\qquad
(1+|v+s\Delta v|)^{-3}\le (1+|s\Delta v|)^3(1+|v|)^{-3}\le (1+|\Delta v|)^3(1+|v|)^{-3}.
\]
Plugging this into \eqref{eq:R1-start} and using $\int_0^1 ds=1$ gives
\begin{align}\label{eq:R1-mid}
\int_{\mathbb R}|R_1(t,x,v)|\,dv
&\le
C(\lambda_0,B_0)\int_{\mathbb R}(|\Delta x|^2 + |\Delta x| |\Delta v| + |\Delta v|^2)(1+|\Delta v|)^{3}(1+|v|)^{-3} \,dv.
\end{align}
Now use \eqref{eq:dx-dv-bounds} and $\int_{\R} (1 + |v|)^{-3} dv <\infty$ to bound the integral, we have:
\begin{align}\label{eq:R1-final}
\int_{\mathbb R}|R_1(t,x,v)|\,dv \le C(\lambda_0,B_0)(1 + D(t)t^{1/2})^3 D(t)^2 t^3.
\end{align}
Combining \eqref{eq:first-order-final} and \eqref{eq:R1-final} with \eqref{eq:rho-pert-1st} gives
\begin{equation}\label{eq:rho-pert-final}
\|\rho_t(\cdot)-\rho_t^{\rm fs}(\cdot)\|_{\mathbb{L}^\infty(\mathbb T)}
\le
C(\lambda_0,B_0)\Big[D(t)t^{1/2} + D(t)t^{3/2}\Big]
+
C(\lambda_0,B_0)\,(1+D(t)t^{1/2})^3D(t)^2t^3.
\end{equation}
Finally, recalling \eqref{eq:split} and \eqref{eq:rho-fs-bound}, we conclude
\[
\|\rho_t(\cdot)-1\|_{\mathbb{L}^\infty(\mathbb T)}
\le
C(\lambda_0,B_0)e^{-c' t}
+
C(\lambda_0,B_0)\Big[D(t)t^{1/2} + D(t)t^{3/2}\Big]
+
C(\lambda_0,B_0)\,(1+D(t)t^{1/2})^3D(t)^2t^3.
\]
Finally, using $t^{1/2}\le 1+t$ and $t^{3/2}\le 1+t^2$ for all $t\ge0$, we could rewrite the bound in terms of polynomial expressions in $t$ as stated in the lemma, which completes the proof.

\section{Technical results used in the proof of \Cref{thm:approximation}}

\subsection{Auxiliary lemmas in the proof of \Cref{lemma:building-block-in-policy-class-construction}}
In this section, we provide the proofs of \Cref{lemma:spatial-regularity-truncation}, \Cref{lemma:exact-isometry}, and \Cref{lem:rho-time-derivative}.

\subsubsection{Proof of \Cref{lemma:spatial-regularity-truncation}}\label{subsubsec:proof-spatial-regularity-truncation}
Recall from \Cref{eq:rhohat-fhat-clean-1d} in the proof of \Cref{lemma:free-stream-decay-general} that the spatial Fourier coefficients of the density correspond to the mixed Fourier transform of the initial condition:
\begin{equation}\label{eq:rhohat-fhat-lemma15}
    \widehat{\rho}_t(k) = \widehat{f}_0(k, kt).
\end{equation}
By \Cref{assume:initial-regularity}, the analytic norm of the initial condition is bounded by $B_0$, which implies that for every $m \in \mathbb{N}$,
\begin{equation}\label{eq:Cm-bound-spatial}
    \|f_0\|_{\mathcal{C}^m_q} \le B_0 \, m! \, \lambda_0^{-m}.
\end{equation}
Fix $k \in \mathbb{Z} \setminus \{0\}$ and $\eta \in \mathbb{R}$. By the definition of the mixed Fourier transform,
\begin{equation*}
    \widehat{f}_0(k, \eta) = \int_{\mathbb{T}} \int_{\mathbb{R}} f_0(x, v) e^{-2\pi i (kx + \eta v)} \, dv \, dx.
\end{equation*}
Integrating by parts $m$ times with respect to the spatial variable $x$, and noting that the boundary terms vanish due to the periodic boundary conditions on $\mathbb{T}$, we obtain
\begin{equation*}
    \widehat{f}_0(k, \eta) = \frac{1}{(2\pi i k)^m} \int_{\mathbb{T}} \int_{\mathbb{R}} \partial_x^m f_0(x, v) e^{-2\pi i (kx + \eta v)} \, dv \, dx.
\end{equation*}
Taking the absolute value, we can bound the integral as
\begin{equation*}
    |\widehat{f}_0(k, \eta)| \le \frac{1}{(2\pi |k|)^m} \int_{\mathbb{T}} \int_{\mathbb{R}} |\partial_x^m f_0(x, v)| \, dv \, dx.
\end{equation*}
Using the polynomial velocity weight built into the definition of the $\|\cdot\|_{\mathcal{C}^m_q}$ norm, we know that $|\partial_x^m f_0(x, v)| \le \|f_0\|_{\mathcal{C}^m_q} (1 + |v|)^{-q}$. Substituting this into the integral yields
\begin{equation*}
    \int_{\mathbb{T}} \int_{\mathbb{R}} |\partial_x^m f_0(x, v)| \, dv \, dx 
    \le \|f_0\|_{\mathcal{C}^m_q} \int_{\mathbb{T}} \int_{\mathbb{R}} (1 + |v|)^{-q} \, dv \, dx 
    = C_* \|f_0\|_{\mathcal{C}^m_q},
\end{equation*}
where $C_* > 0$ is a universal finite constant. Combining this with the bound in \eqref{eq:Cm-bound-spatial}, we get
\begin{equation}\label{eq:fhat-ibp-bound-m}
    |\widehat{f}_0(k, \eta)| \le C_* \, B_0 \, m! \, (2\pi \lambda_0 |k|)^{-m}, \qquad \forall m \in \mathbb{N}.
\end{equation}
To reconstruct the exponential decay, we multiply both sides of \eqref{eq:fhat-ibp-bound-m} by $(\pi \lambda_0 |k|)^m / m!$ and sum over all integers $m \ge 0$:
\begin{equation*}
    |\widehat{f}_0(k, \eta)| \sum_{m=0}^\infty \frac{(\pi \lambda_0 |k|)^m}{m!} 
    \le C_* B_0 \sum_{m=0}^\infty \frac{(\pi \lambda_0 |k|)^m}{(2\pi \lambda_0 |k|)^m}.
\end{equation*}
Recognizing the Taylor series for the exponential function on the left side, and simplifying the geometric series on the right side, we find
\begin{equation*}
    |\widehat{f}_0(k, \eta)| \, e^{\pi \lambda_0 |k|} 
    \le C_* B_0 \sum_{m=0}^\infty \left(\frac{1}{2}\right)^m 
    = 2 C_* B_0.
\end{equation*}
Letting $C_0 := 2 C_* B_0$ and $c_0 := \pi \lambda_0$, we arrive at the uniform exponential bound
\begin{equation}\label{eq:fhat-exp-spatial-rigorous}
    |\widehat{f}_0(k, \eta)| \le C_0 e^{-c_0 |k|}, \qquad \forall k \in \mathbb{Z}, \, \eta \in \mathbb{R}.
\end{equation}
which implies $|\widehat{\rho}_t(k)| = |\widehat{f}_0(k, kt)| \le C_0 e^{-c_0 |k|}$ uniformly for all $t \in [0, T]$. 
By Parseval's identity, the truncation error is bounded by the tail of this geometric series:
\begin{equation*}
    \big\|\rho_t - \rho_t^{(M)}\big\|_{\mathbb{L}^2(\mathbb{T})}^2 
    = \sum_{|k| > M} |\widehat{\rho}_t(k)|^2 
    \le 2 \sum_{k = M+1}^\infty C_0^2 e^{-2c_0 k} =  2 C_0^2 \frac{e^{-2c_0 (M+1)}}{1 - e^{-2c_0}} 
    \le C_1 e^{-2c_0 M},
\end{equation*}
where $C_1 > 0$ is a constant depending exclusively on $\lambda_0$ and $B_0$. For any arbitrary tolerance $\varepsilon_{\mathrm{tr}} > 0$, choosing an integer $M \ge \frac{1}{2c_0} \ln(C_1 / \varepsilon_{\mathrm{tr}}^2)$ guarantees that $\big\|\rho_t - \rho_t^{(M)}\big\|_{\mathbb{L}^2(\mathbb{T})} \le \varepsilon_{\mathrm{tr}}$ uniformly over $[0, T]$, which completes the proof.
\subsubsection{Proof of \Cref{lemma:exact-isometry}}\label{subsubsec:proof-exact-isometry}
  Let $G = \frac{1}{N}\Phi^\ast \Phi \in \mathbb{C}^{J \times J}$ be the normalized Gram matrix. For any two frequency indices $p, q \in \mathcal{K}$, the $(p, q)$-th entry of $G$ is given by
  \begin{equation*}
    G_{p,q} = \frac{1}{N} \sum_{j=1}^N \overline{\Phi_{jp}} \Phi_{jq} = \frac{1}{N} \sum_{j=1}^N e^{2\pi i (q - p) \frac{j}{N}}.
  \end{equation*}
  If $p = q$, the exponent is zero for all $j$, yielding $G_{p,p} = \frac{1}{N} \sum_{j=1}^N 1 = 1$.\\
  If $p \neq q$, we define the frequency difference $\Delta k = q - p$. Since $p, q \in \{-M, \dots, M\}$, the absolute difference is strictly bounded by $|\Delta k| \le 2M$. Since $N > 2M$, it follows that $0 < |\Delta k| < N$. \\
  Let $\omega = e^{2\pi i \Delta k / N}$. Since $\Delta k / N$ is strictly between $-1$ and $1$ and not zero, $\omega \neq 1$. Noting that $\omega^N = e^{2\pi i \Delta k} = 1$, we conclude that the off-diagonal entry is a finite geometric series of roots of unity:
  \begin{equation*}
    G_{p,q} = \frac{1}{N} \sum_{j=1}^N \omega^j = \frac{1}{N} \omega \frac{1 - \omega^N}{1 - \omega} = 0.
  \end{equation*}
  Thus, $G = I_J$, which immediately establishes the exact isometry $\frac{1}{N}\|\Phi u\|_2^2 = u^\ast G u = \|u\|_2^2$.\\
  Since $G$ is the identity matrix, its eigenvalues are all exactly $1$, implying that $\Phi$ has full column rank. The pseudoinverse simplifies to $\Phi^\dagger = (\Phi^\ast \Phi)^{-1}\Phi^\ast = \frac{1}{N}\Phi^\ast$. The operator norm of $\Phi^\dagger$ is the square root of the maximum eigenvalue of $\Phi^\dagger (\Phi^\dagger)^\ast = \frac{1}{N^2}\Phi^\ast \Phi = \frac{1}{N}I_J$, which is exactly $1/\sqrt{N}$, which completes the proof.

\subsubsection{Proof of Lemma~\ref{lem:rho-time-derivative}}\label{subsec:proof-rho-time-derivative}Since the expert policy perfectly cancels the internal electric field for $t \ge t_0$, the system is governed by the free-stream equation $\partial_t f_t + v \partial_x f_t = 0$. The explicit solution is $f_s(x,v) = f_0(x-sv, v)$. \\
For any integer $1 \le \ell \le q-2$, applying the chain rule yields $\partial_s^\ell f_s(x,v) = (-v)^\ell (\partial_x^\ell f_0)(x-sv, v)$. Integrating over the velocity domain gives the temporal derivative of the density:
\[
  \partial_s^\ell \rho_s(x) = \int_{\mathbb{R}} (-v)^\ell (\partial_x^\ell f_0)(x-sv, v) \, dv.
\]
Taking absolute values and using the polynomial velocity weight $(1+|v|)^q$ implicit in the analytic norm $\|f_0\|_{\mathcal{E}_q(\lambda_0)}$ from Assumption~\ref{assume:initial-regularity}, we obtain
\[
  |\partial_s^\ell \rho_s(x)| \le \int_{\mathbb{R}} |v|^\ell \|f_0\|_{\mathcal{C}^\ell_q} (1+|v|)^{-q} \, dv \le \|f_0\|_{\mathcal{C}^\ell_q} \int_{\mathbb{R}} (1+|v|)^{\ell-q} \, dv.
\]
Because we restrict $\ell \le q - 2$, the exponent satisfies $\ell - q \le -2 < -1$. This strictly guarantees that the integral $\int_{\mathbb{R}} (1+|v|)^{\ell-q} \, dv$ converges to a finite constant $C_{q,\ell}$ depending only on $q$ and $\ell$. Thus, we have
\[
  |\partial_s^\ell \rho_s(x)| \le C_{q,\ell} \|f_0\|_{\mathcal{C}^\ell_q} \le C_\ell,
\]
uniformly for all $s \in [t-K\Delta t, t]$ and $x \in \torus$, which completes the proof.
\subsubsection{Proof of Lemma~\ref{lem:local-extrap}}\label{subsec:proof-local-extrap}
Fix a sensor index $i$ and a base time step $k \in \mathcal{K}_p$. Let $\delta_k \defn k\Delta t$. The $p$ observation times used by the estimator are $t_{jk} = t - j\delta_k$ for $j = 1, \dots, p$. \\
By Taylor's theorem with the Lagrange remainder, we expand the macroscopic density $\rho$ around the current time $t$ up to order $p-1$:
\begin{equation}
  \rho_{t_{jk}}(x_i) = \sum_{m=0}^{p-1} \frac{\partial_t^m \rho_t(x_i)}{m!} (-j\delta_k)^m + \frac{\partial_t^p \rho_{s_j}(x_i)}{p!} (-j\delta_k)^p, \label{eq:taylor-p}
\end{equation}
where $s_j \in [t_{jk}, t]$ is the intermediate time for the remainder term. \\
Taking the expectation of the estimator $Z_{i,k}$, the zero-mean noise terms $\varepsilon_{i,jk}$ vanish. Substituting the Taylor expansion \eqref{eq:taylor-p} into the expectation yields:
\begin{align*}
  \E[Z_{i,k}] 
  &= \sum_{j=1}^p (-1)^{j-1} \binom{p}{j} \rho_{t_{jk}}(x_i) \\
  &= \sum_{m=0}^{p-1} \frac{\partial_t^m \rho_t(x_i)}{m!} (-\delta_k)^m \underbrace{ \left[ \sum_{j=1}^p (-1)^{j-1} \binom{p}{j} j^m \right] }_{\defn S_m} + R_p,
\end{align*}
where $R_p \defn \sum_{j=1}^p (-1)^{j-1} \binom{p}{j} \frac{\partial_t^p \rho_{s_j}(x_i)}{p!} (-j\delta_k)^p$ is the aggregated remainder.\\
By the standard properties of the finite difference operator applied to polynomials, the combinatorial sums $S_m$ evaluate to exactly:
\begin{equation*}
  S_m = \sum_{j=1}^p (-1)^{j-1} \binom{p}{j} j^m = 
  \begin{cases} 
    1 - (1-1)^p = 1, & \text{if } m = 0, \\
    0, & \text{if } 1 \le m \le p-1.
  \end{cases}
\end{equation*}
Consequently, all time derivatives of order $1$ through $p-1$ perfectly cancel out, and the $m=0$ term recovers exactly $\rho_t(x_i)$. The estimation bias is purely governed by the $p$-th order remainder $R_p$:
\begin{equation*}
  \big|\E[Z_{i,k}] - \rho_t(x_i)\big| = |R_p| \le \frac{\delta_k^p}{p!} \sum_{j=1}^p \binom{p}{j} j^p \big|\partial_t^p \rho_{s_j}(x_i)\big|.
\end{equation*}
By Lemma~\ref{lem:rho-time-derivative}, there exists a uniform constant $C_p' > 0$ such that $|\partial_t^p \rho_s(x_i)| \le C_p'$ for all $s \in [t-K\Delta t, t]$. Absorbing all $p$-dependent combinatorial constants into a single constant $C_p$, we obtain the rigorous bias bound:
\begin{equation*}
  \big|\E[Z_{i,k}] - \rho_t(x_i)\big| \le C_p \delta_k^p = C_p (k\Delta t)^p,
\end{equation*}
which establishes \eqref{eq:local-bias}.\\
For the variance, the estimator decomposes as $Z_{i,k} = \E[Z_{i,k}] + \sum_{j=1}^p (-1)^{j-1} \binom{p}{j} \varepsilon_{i,jk}$. Since the observation noises $\varepsilon_{i,jk}$ are sampled at distinct historical time steps $t_{jk}$, they are mutually independent. The variance is thus the sum of the individual variances scaled by the squared weights:
\begin{equation*}
  \Var(Z_{i,k}) = \sum_{j=1}^p \left( (-1)^{j-1} \binom{p}{j} \right)^2 \Var(\varepsilon_{i,jk}) = \sigma^2 \sum_{j=1}^p \binom{p}{j}^2.
\end{equation*}
Using the Vandermonde convolution identity $\sum_{j=0}^p \binom{p}{j}^2 = \binom{2p}{p}$, we evaluate the sum starting from $j=1$ as:
\begin{equation*}
  \Var(Z_{i,k}) = \left[ \binom{2p}{p} - \binom{p}{0}^2 \right] \sigma^2 = \left[ \binom{2p}{p} - 1 \right] \sigma^2 \defn V_p \sigma^2,
\end{equation*}
which establishes \eqref{eq:local-var}, and completes the proof.

\subsubsection{Proof of Lemma~\ref{lem:sensor-bias-variance}}\label{subsec:proof-sensor-bias-variance}
By the linearity of expectation and the bias bound \eqref{eq:local-bias} from Lemma~\ref{lem:local-extrap}, the expectation of the averaged estimator $\widetilde{Y}_i$ satisfies:
\begin{align*}
  \big|\E[\widetilde{Y}_i] - \rho_t(x_i)\big| &= \Big| \frac{1}{J_p} \sum_{k \in \mathcal{K}_p} \big( \E[Z_{i,k}] - \rho_t(x_i) \big) \Big| \\
  &\le \frac{1}{J_p} \sum_{k \in \mathcal{K}_p} C_p (k \Delta t)^p.
\end{align*}
Since all indices $k \in \mathcal{K}_p$ are bounded by $K/p$, we have $(k \Delta t)^p \le \frac{1}{p^p} (K\Delta t)^p$. Therefore, the uniform bound on the overall bias is:
\begin{equation*}
  \big|\E[\widetilde{Y}_i] - \rho_t(x_i)\big| \le \frac{C_p}{p^p} (K \Delta t)^p \defn C_B (K \Delta t)^p,
\end{equation*}
which establishes \eqref{eq:sensor-bias}.\\
For the variance, the construction of the index set $\mathcal{K}_p$ guarantees that for any two distinct $k_1, k_2 \in \mathcal{K}_p$, their corresponding sets of observation indices $\{j k_1\}_{j=1}^p$ and $\{j k_2\}_{j=1}^p$ are mutually disjoint. Consequently, the random variables $\{Z_{i,k}\}_{k \in \mathcal{K}_p}$ are mutually independent because they are constructed from non-overlapping noise samples. Using this independence and \eqref{eq:local-var}, we have:
\begin{equation*}
  \Var(\widetilde{Y}_i) = \frac{1}{J_p^2} \sum_{k \in \mathcal{K}_p} \Var(Z_{i,k}) = \frac{1}{J_p^2} \big( J_p \cdot V_p\sigma^2 \big) = \frac{V_p\sigma^2}{J_p}.
\end{equation*}
Under the condition $K \ge 2p^2$, we have established that the number of estimators satisfies $J_p \ge K / (2p^2)$. Substituting this lower bound, we obtain:
\begin{equation*}
  \Var(\widetilde{Y}_i) \le \frac{2p^2 V_p \sigma^2}{K} \defn C_V \frac{\sigma^2}{K}.
\end{equation*}
This establishes \eqref{eq:sensor-var}, and completes the proof.

\subsection{Proof of \Cref{lemma:entropy-properties}}\label{subsec:app-proof-lemma-entropy-properties}
For the first property, we note that for any $x$ in the support of $\Prob$, we have
\begin{align*}
  (\phi_\sharp \Prob) \big( \ball_{\ltwospace} (\phi(x), \varepsilon) \big) = \Prob \Big( \ltwonorm{\phi (X) - \phi (x)} \leq \varepsilon \Big) \geq \Prob \Big( \ltwonorm{X - x} \leq \varepsilon / L \Big) = \Prob \big( \ball_{\ltwospace} (x, \varepsilon / L) \big).
\end{align*}
Taking the logarithm and expectation, we have
\begin{align*}
  H_\varepsilon (\phi_\sharp \Prob) &= - \Exs_{X \sim \Prob} \big[ \log (\phi_\sharp \Prob) \big(\ball_{\ltwospace} (\phi(X), \varepsilon) \big) \big] \\
  &\leq - \Exs_{X \sim \Prob}big[ \log \Prob \big( \ball_{\ltwospace} (X, \varepsilon / L) \big) \big] = H_{\varepsilon / L} (\Prob).
\end{align*}
Now we prove the second property. Denote the mixture measure by $\Prob_\mu \mydefn \int_{\mathcal{A}} \Prob_a d\mu(a)$ and define the posterior distribution $\mu_x$ over the index set $\mathcal{A}$ given an observation $x$ as
\begin{align*}
  \mu_x (a) \mydefn \frac{d\Prob_a}{d\Prob_\mu} (x).
\end{align*}
By Donsker--Varadhan variational formula, we have
\begin{align*}
  \kull{\mu_x}{\mu} \geq \int \log \Prob_a \big( \ball_{\ltwospace} (x, \varepsilon) \big) d\mu_x(a) - \log \Prob_\mu \big( \ball_{\ltwospace} (x, \varepsilon) \big),
\end{align*}
for any $x$ in the support of $\Prob_\mu$. Rearranging the terms and taking expectation over $X \sim \Prob_\mu$, we have
\begin{align*}
  H_\varepsilon (\Prob_\mu) &= - \Exs_{X \sim \Prob_\mu} \big[ \log \Prob_\mu \big( \ball_{\ltwospace} (X, \varepsilon) \big) \big] \\
  &\leq - \Exs_{X \sim \Prob_\mu} \big[ \int \log \Prob_a \big( \ball_{\ltwospace} (X, \varepsilon) \big) d\mu_X(a) \big] + \int \kull{\mu_x}{\mu} d \mu (a) \\
  &= \int H_\varepsilon (\Prob_a) d\mu(a) + I
  \big( A; X \big),
\end{align*}
which completes the proof.

\subsection{Proof of \Cref{lemma:expert-policy-regularity}}\label{subsec:app-proof-lemma-expert-policy-regularity}
We analyze the difference between the two solutions, denoted as $\Delta f_t \defn f_t^{(1)} - f_t^{(2)}$ and the corresponding internal electric field difference $\Delta E_t \defn E_t^{(1)} - E_t^{(2)}$. The evolution naturally divides into two phases: the uncontrolled nonlinear burn-in phase and the expert-controlled free-streaming phase.\\
We firstly analyze the nonlinear burn-in period, both solutions satisfy the uncontrolled Vlasov--Poisson equation. Subtracting the two equations yields the corresponding equation:
\begin{equation*}
    \partial_t \Delta f_t + v \partial_x \Delta f_t - E_t^{(1)} \partial_v \Delta f_t - \Delta E_t \partial_v f_t^{(2)} = 0.
\end{equation*}
Multiplying by $\Delta f_t$ and integrating over the phase space $\torus \times \real$, the transport term and the $E_t^{(1)}$ term vanish due to integration by parts. This yields the standard $L^2$ energy identity:
\begin{equation*}
    \frac{1}{2} \frac{d}{dt} \| \Delta f_t \|_{L^2_{x,v}}^2 = \iint_{\torus \times \real} \Delta E_t(x) \, \partial_v f_t^{(2)}(x,v) \, \Delta f_t(x,v) \, dx dv.
\end{equation*}
Applying the Cauchy--Schwarz inequality sequentially in velocity and space, the right-hand side is bounded by:
\begin{align*}
    \frac{1}{2} \frac{d}{dt} \| \Delta f_t \|_{L^2_{x,v}}^2 &\le \int_{\torus} |\Delta E_t(x)| \Big( \int_{\real} |\partial_v f_t^{(2)}| |\Delta f_t| \, dv \Big) dx \\
    &\le \int_{\torus} |\Delta E_t(x)| \| \partial_v f_t^{(2)}(x, \cdot) \|_{L^2_v} \| \Delta f_t(x, \cdot) \|_{L^2_v} dx \\
    &\le \| \Delta E_t \|_{L^2_x} \sup_{x \in \torus} \| \partial_v f_t^{(2)}(x, \cdot) \|_{L^2_v} \| \Delta f_t \|_{L^2_{x,v}}.
\end{align*}
By Lemma~\ref{lemma:short-time-vp}, the solution $f_t^{(2)}$ is uniformly bounded in the analytic space $\mathcal{E}_q(\lambda_t)$ up to $\tburn$. Because $\lambda_{\tburn} > 0$, the velocity derivative is uniformly bounded: there exists a constant $C_B > 0$, depending only on $B_0, \lambda_{\tburn},$ and $q$, such that $\sup_{x} \|\partial_v f_t^{(2)}\|_{L^2_v} \le C_B$. 
Furthermore, the Poisson equation $\partial_x \Delta E_t = \int \Delta f_t dv$ implies $\| \Delta E_t \|_{L^2_x} \le C_P \| \Delta f_t \|_{L^2_{x,v}}$. Substituting these into the energy inequality, we obtain:
\begin{equation*}
    \frac{d}{dt} \| \Delta f_t \|_{L^2_{x,v}}^2 \le 2 C_P C_B \| \Delta f_t \|_{L^2_{x,v}}^2.
\end{equation*}
Applying Gr\"onwall's inequality over $t \in [0, \tburn]$, we secure the local Lipschitz stability:
\begin{equation}
    \| \Delta f_{\tburn} \|_{L^2_{x,v}} \le \exp(C_P C_B \tburn) \| \Delta f_0 \|_{L^2_{x,v}}. \label{eq:gronwall-bound}
\end{equation}
For $t \ge \tburn$, the expert policy applies the exact canceling field $\ctrlE_t^{\mathrm{expert}} = -E_t$, reducing the Vlasov equation to the pure free-streaming equation $\partial_t f_t + v \partial_x f_t = 0$. Consequently, the perturbation evolves via $\Delta f_t(x,v) = \Delta f_{\tburn}(x - v(t-\tburn), v)$. Since this spatial shift is a measure-preserving transformation, the $L^2$ norm is perfectly conserved:
\begin{equation*}
    \| \Delta f_t \|_{L^2_{x,v}} = \| \Delta f_{\tburn} \|_{L^2_{x,v}}, \qquad \forall t \ge \tburn.
\end{equation*}
Finally, by the definition of the expert policy and the Poisson equation bound, we translate the phase-space bound back to the macroscopic control field:
\begin{align*}
    \| \ctrlE_t^{\mathrm{expert}, (1)} - \ctrlE_t^{\mathrm{expert}, (2)} \|_{L^2_x} &= \| \Delta E_t \|_{L^2_x} \\
    &\le C_P \| \Delta f_t \|_{L^2_{x,v}} \\
    &= C_P \| \Delta f_{\tburn} \|_{L^2_{x,v}}.
\end{align*}
Combining this with the Gr\"onwall bound \eqref{eq:gronwall-bound}, we obtain:
\begin{equation*}
    \| \ctrlE_t^{\mathrm{expert}, (1)} - \ctrlE_t^{\mathrm{expert}, (2)} \|_{L^2_x} \le \underbrace{C_P \exp(C_P C_B \tburn)}_{ \defn C } \| f_0^{(1)} - f_0^{(2)} \|_{L^2_{x,v}},
\end{equation*}
which establishes the desired uniform bound and completes the proof.
\end{document}